%% file: arxiv_main.tex
\documentclass{article}
\usepackage{style,times}
\usepackage{amsopn}
\usepackage[english]{babel}
\usepackage{booktabs,graphicx,xcolor,placeins,microtype}
\input{preambule.tex}

\input{commands.tex}
\usepackage{mathrsfs,xurl}
\hypersetup{hidelinks}
\providecommand{\doi}[1]{\href{https://doi.org/#1}{doi:\nolinkurl{#1}}}

\newcommand{\pclean}{p_{\mathrm{clean}}}
\newcommand{\Lobs}{L_{\mathrm{obs}}}
\newcommand{\Pois}{\operatorname{Pois}}
\newcommand{\Binomial}{\operatorname{Binomial}}
\newcommand{\GammaDist}{\operatorname{Gamma}}

\title{Learning Clean Generative Processes from Noisy Measurements}
\author{Fabian Kruse, Lorenzo Baldassari \& Ivan Dokmani\'c \\
Department of Mathematics and Computer Science \\
University of Basel \\
\texttt{fabian.kruse@unibas.ch} \\
\texttt{lorenzo.baldassari@unibas.ch} \\
\texttt{ivan.dokmanic@unibas.ch}}
\iclrfinalcopy

\begin{document}
\maketitle

\begin{abstract}
We study how to learn a generative model of clean images using only noisy observations. Diffusion processes and their generalizations to non-Gaussian noise require posterior predictions at every noise level. We first show that it is possible to learn these predictions from independent noisy views using Noise2Noise regression, but only at or above the noise level of the training data. To learn the predictions below that level, needed to generate clean images, we impose consistency: a prediction should equal its expectation after continuing the sampling process. We derive conditions under which consistency uniquely determines the continuation to the clean image distribution and verify them for Poisson, binomial, Gamma, and negative-binomial measurements. Experiments on CIFAR-10, photon-counting microscopy, and synthetic aperture radar show that the required posterior predictions can be learned from noisy views alone and that the consistency relation repairs the continuation beyond the noise level of the data, which severely deteriorates without consistency.
\end{abstract}

\section{Introduction}\label{sec:intro}
Can one learn to generate clean images from a dataset of noisy observations? This question arises naturally in scientific imaging. For example, photon-counting microscopy often allows only short exposures with a few photons per pixel, because more light would damage the sample \citep{krull2024image}. In synthetic aperture radar (SAR), images are corrupted by speckle \citep{dalsasso2021if,dalsasso2021sar2sar}. In cryo-electron tomography the electron beam damages the sample, so the dose is kept low and projections are very noisy \citep{buchholz2019cryo,liu2022isotropic,kishore2026icecream}.
In each case the noise model is known, but clean images are unavailable. This presents a challenge for diffusion models, which are typically trained
by regressing onto clean images.

\cite{pmlr-v80-lehtinen18a} showed that denoisers can be trained using only noisy images. Their method, Noise2Noise (N2N), trains a network to predict a second noisy measurement of an image from the first. If the two measurements are independent conditional on the clean image, and averaging the second measurement over its noise gives the clean image, the minimizer of the expected squared error is the posterior mean of the clean image given the first measurement. The network therefore learns a denoiser without ever seeing a clean image. Later work obtained the two noisy views from a single measurement, either by adding further noise \citep{moran2020noisier2noise,pang2021recorrupted} or by splitting the measurement into two parts with independent noise. Variants of splitting exist for Gaussian, Poisson, Gamma and negative-binomial measurements \citep{neufeld2024data,leiner2025data,monroy2025generalized}. 
Since Gaussian diffusion models generate images with denoisers at a sequence of noise levels
\citep{song2020score}, N2N allows one to train them from noisy data. \citet{pmlr-v235-daras24a} take this route: noisy
images at one noise level supervise the denoisers at all noisier levels, and
the consistency loss of \citet{daras2023consistent} extends them to cleaner levels.

But this only solves the problem when the measurements are corrupted by Gaussian
noise. For photon counts, radar speckle and other measurement laws, a
Gaussian diffusion is no longer matched to the data: it would add Gaussian
noise on top of the different noise  in the measurements, and reversing
this process would lead back to the noisy measurements, not to the
clean images. 
 
To address this, we build the generative process directly from the specific noise model that produced the data. We index the process by an \emph{information level} $a$ that increases as the noise decreases, such as the precision (inverse variance) of a Gaussian measurement or the exposure of a photon count. Its state at level $a$ is distributed like a measurement at that level, and as $a$ grows, the normalized state converges to the clean image. Stochastic localization describes Gaussian diffusion in this way: after a change of time and scale, the process can be viewed as
accumulating Gaussian measurements of a single image. As more measurements are accumulated, their average converges to
that image \citep{montanari2023sampling}. Photon counting works the same way. The process accumulates photons as the exposure increases,  and the counts divided by the exposure converge to the clean image
\citep{krull2024image}. We simulate both processes using the posterior mean of the clean image given the measurements accumulated so far. This mean is the drift in the Gaussian process and the photon arrival rate at each pixel in the Poisson process. In Section~\ref{sec:observation-processes} and Appendix~\ref{app:models} we unify these formulations in a framework that, alongside Gaussian and Poisson noise, also covers binomial, Gamma and negative-binomial noise, and we identify the posterior predictions that drive each process. These noise models have a range of applications: binomial noise arises when a detector registers each particle with a fixed probability; Gamma noise describes speckle in radar and other coherent imaging systems; negative-binomial noise describes counts whose variance exceeds their mean. 

Once the posterior predictions that drive these processes have been identified, the question is how to learn them from noisy data. We describe how to do this by N2N regression in Section~\ref{sec:formulation}. Let $a_{\rm in}$ denote the information level of the available noisy  measurements that we further degrade to train the network. For each sample $X$ from the clean image distribution, we assume access to two noisy measurements $Y_{\rm in}$ and $Y_{\rm tar}$, at information levels $a_{\rm in}$ and $a_{\rm tar}$, respectively, that are independent conditional on $X$.  To train the network at a chosen level $a\leq a_{\rm in}$, we degrade $Y_{\rm in}$ to obtain a training input $Z_a$ at that level. The network takes $Z_a$ as an input and is trained to predict a target formed from $Y_{\rm tar}$.  For photon counts, this target is $Y_{\rm tar}/a_{\rm tar}$, whose expectation conditional on $X$ is the clean image itself. Since $Y_{\rm tar}$ and $Z_a$ are independent conditional on $X$, N2N regression against $Y_{\rm tar}/a_{\rm tar}$ learns $\mathbb E[X\mid Z_a]$. For other noise models, simulating the process may require posterior expectations of different functions $\psi (X)$. We denote these posterior predictions by $h_{a}(z)=\mathbb E[\psi (X)\mid Z_a=z]$. We derive transformations $T (Y_{\rm tar})$ whose conditional means given $X$ equal $\psi (X)$, so that N2N regression learns $h_{a}$.

By training on degraded measurements at different information levels $a\leq a_{\rm in}$, we learn the posterior predictions needed to simulate each generative process up to $a_{\rm in}$. To generate clean images, however, we also need predictions at $a>a_{\rm in}$ (Figure~\ref{fig:diff-overview}). Measurements at these higher levels cannot be obtained by degrading $Y_{\rm in}$, so N2N regression alone cannot be used to learn these predictions. To learn them, we use the tower property of conditional expectation: the posterior prediction at one state must equal the expected posterior prediction after continuing the process to a higher information level. We enforce this consistency relation along paths simulated with the model itself, and keep the N2N loss at and below $a_{\rm in}$. This extends the consistency approach of \citet{daras2023consistent,pmlr-v235-daras24a} from Gaussian noise to the other noise models of the paper.

The theoretical question is why the consistency relation should recover the correct predictions above $a_{\rm in}$. 
\citet{daras2023consistent} provide an answer for Gaussian
diffusion, assuming that the score computed from the learned
denoiser is a gradient field, using tools specific to the Gaussian case such as the heat equation. In Section~\ref{sec:continuation}, we give two arguments that do not rely on these tools and cover a more general class of noise models. The first applies to all noise models considered in this paper, provided that the learned  predictions are derived
from a common density ratio. The second does not require a common density ratio and gives uniqueness
conditions for the non-Gaussian noise models considered.
We then connect these results to the training objective that combines the N2N and
consistency losses, giving conditions
under which exact minimization recovers the correct predictions above $a_{\rm in}$.

Finally, we complement these theoretical results with experiments evaluating the proposed approach on synthetic and real data. On corrupted CIFAR-10 images and on photon-counting microscopy data, the quality of sampled images deteriorates when the process continues above $a_{\rm in}$. Training with the consistency loss restores most of the lost quality. On synthetic aperture radar data, we find that the learned sampler reproduces dependence between pixels in conditional samples, whereas sampling each pixel independently fails to reproduce this dependence. We discuss related work in
Appendix~\ref{sec:related}.
\input{process_overview}

\section{From noisy observations to a generative model}\label{sec:setup}

Let $X\sim\pclean$ denote a clean image. We observe noisy measurements of $X$ under a known measurement law, but do not know the clean distribution $\pclean$. In the following, we define a process that can generate images from $\pclean$ using predictions learned from the measurements.

\subsection{Observation processes and posterior predictions}
\label{sec:observation-processes}
Consider the Gaussian noising process
\[
 Y_t=X+ W_t,
\]
where $ W$ is standard Brownian motion independent of $X$. The noise variance is $t$, and the usual stochastic sampler runs towards $t=0$~\citep{song2020score}. We instead run sampling forward in the precision $a\eqdef 1/t$. We rescale each state by setting 
\begin{equation*}
    Z_a = a Y_{1/a} = aX + V_a, \qquad V_a = aW_{1/a}, \quad a>0,
\end{equation*}
where time inversion makes $V$ a standard Brownian motion with $V_0 = 0$. Therefore 
\[
Z_a/a\mid X=x\sim\Nc(x,a^{-1}\Id).
\]
For $b>a$, the additional observation
\[
\frac{Z_b-Z_a}{b-a}= X+\frac{V_b-V_a}{b-a}
\]
has precision $b-a$ and is independent of $Z_a$ conditional on $X$. Thus $Z_a$ accumulates measurements and the normalized state $Z_a/a$ converges almost surely to the clean image $X$~\citep{montanari2023sampling}. 

For photon counts, let $N_a$ record the photons collected by exposure $a$. Conditional on $X$, pixel $i$ receives photons at rate $X_i$, independently of other pixels: 
\[
N_a \mid X = x \sim \Pois(ax).
\]
Longer exposure adds independent counts, and $N_a/a\to X$ almost surely. As before, increasing the information level reduces the noise in the normalized image.

We write $Z_a$ for the observation state in general, and $N_a$ or $S_a$ when it holds counts or Gamma variables. We define the stochastic process $(Z_a)$, indexed by the information level $a$. We write 
\[
Q_a(x,\dd z) = \operatorname{Law} (Z_a\in \dd z \mid X= x).
\]
For example, $Q_a(x,\cdot) = \Nc(ax, a\Id)$ in the Gaussian case and $\operatorname{Pois}(ax)$ in the Poisson case. The process starts at the empty state $Z_0=0$, which carries no information about $X$, and is defined on $0 \leq a < a_{\rm max}$, where $a_{\rm max}= \infty$ except for binomial observations. In the binomial model, $X$ is integer-valued, $a$ is a detection probability, and $N_a\mid X\sim\Binomial(X,a)$. The clean endpoint is $N_1=X$, reached at $a_{\rm max}=1$.

Table~\ref{tab:instances} lists these models together with Gamma and negative-binomial observations, where $a$ is the Gamma shape or the negative-binomial shape.
Similarly, observations of these processes become more informative as $a$ increases and, after normalization, converge almost surely to $X$ or to a known invertible encoding of $X$ for negative-binomial observations. Appendix~\ref{app:models} gives the details for all five measurement models.

\paragraph{Degradation.}
Consider first the process in the direction from more information to less. An observation at information level $b$ can be degraded to level $a<b$ without knowing $X$. For photon counts this is binomial thinning: each recorded photon is kept independently with probability $a/b$, so $N_a\mid N_b =n\sim \Binomial(n, a/b)$, a rule that does not involve $X$. Let $\overleftarrow{K}_{b,a}(z,\dd z')$ denote this degradation
kernel. It satisfies
\begin{equation}
Q_b(x,\cdot)\overleftarrow{K}_{b,a}=Q_a(x,\cdot),
\qquad
\overleftarrow{K}_{c,b}\overleftarrow{K}_{b,a}=\overleftarrow{K}_{c,a},
\qquad a<b<c,
\label{eq:main-kernels}
\end{equation}
so degradation produces the correct law at level $a$, directly or through an intermediate level. Appendix~\ref{app:degradation} gives the analogous kernels for the other noise models. We use the kernels to define the process at all levels: given $Z_b$, the states below $b$ are obtained from $Z_b$ by degradation, without further reference to $X$. Consequently, once $Z_a$ is known, the states at lower levels provide no additional information about $X$:
\begin{equation*}
\operatorname{Law}\!\left(X\mid (Z_u)_{u\leq a}\right)
=
\operatorname{Law}\!\left(X\mid Z_a\right),
\end{equation*}
and $(Z_a)$ is a Markov process (Lemma~\ref{lem:app-posterior}). We can therefore  predict its next increment from the current state alone.

\paragraph{Generation.}

The generative process moves in the opposite direction, from less information to more. Starting from a state
$Z_a$, we simulate states $Z_b$ at information levels $b>a$.
If $X$ were known, the noise model would specify how the state evolves as information is added. In the Gaussian case, $\dd Z_a =X\, \dd a+\dd V_a$. In the Poisson case, each pixel $i$ gains a count in a short interval of length $\delta$ with probability proportional to its intensity,
\[
\operatorname{Pr}(N_{a+\delta, i} - N_{a,i}=1 \mid X,N_a) = X_i \delta + o(\delta), \quad \delta \downarrow 0.
\]
In both cases the clean image enters the dynamics through the drift or the rates, which equal $X$. For the other noise models the rates involve other functions of $X$, such as $e^{-w/X_i}$ for a Gamma jump size $w$ in pixel $i$ (Table~\ref{tab:instances}). We index these functions by $\xi$ and write them as $\psi_\xi(X)$. Since the clean image $X$ is unknown, we estimate each $\psi_\xi(X)$ by its posterior mean given the current state, which we call a posterior prediction:
\begin{equation}
h_{\xi,a}(z)
=
\E\!\left[\psi_\xi(X)\mid Z_a=z\right].
\label{eq:main-posterior-prediction}
\end{equation}
In the Gaussian case, $\psi_\xi(X) = X$ and the posterior mean $m_a(z)=\E[X\mid Z_a=z]$ takes the place of $X$:
\[
\dd Z_a=m_a(Z_a)\,\dd a+\dd \overline V_a,
\]
where $\overline V$ is a Brownian motion with respect to the history of the observation process.
For the Poisson process, $\psi_i(X)=X_i$, and pixel $i$ gains counts at rate $h_{i,a}(N_a)=\E[X_i\mid N_a]$. The Gamma process has $\psi_{i,w}(X)=e^{-w/X_i}$. In each case, conditioning on the observation replaces $\psi_\xi(X)$ by $h_{\xi,a}(Z_a)$, as shown in Table~\ref{tab:instances}. Learning these predictions lets us simulate the process without knowing the clean image. 

\begin{table}[t]
\centering
\small
\setlength{\tabcolsep}{3.5pt}
\begin{tabular}{@{}llll@{}}
\toprule
Measurement & $Z_a$ given $X$ & Dynamics given $X$ & Dynamics given $Z_a$\\
\midrule
Gaussian
& $\Nc(aX,a\Id)$
& drift $X$
& $\E[X\mid Z_a]$
\\
Poisson
& $\Pois(aX)$
& pixel $i$ gains a count at rate $X_{i}$
& $\E[X_{i}\mid N_a]$
\\
Binomial ($a<1$)
& $\Binomial(X,a)$
& pixel $i$ gains a count at rate $\frac{X_{i}-N_{a,i}}{1-a}$
& $\frac{\E[X_{i}\mid N_a]-N_{a,i}}{1-a}$
\\[1mm]
Gamma
& $\GammaDist(a,1/X)$
& pixel $i$ jumps by $w>0$ at rate $e^{-w/X_{i}}\,\dd w/w$
& $\E[e^{-w/X_{i}}\mid S_a]\,\dd w/w$
\\[1mm]
Negative binomial
& $\mathrm{NB}(a,q(X))$
& pixel $i$ jumps by $k\ge1$ at rate $q_i(X)^k/k$
& $\E[q_i(X)^k\mid N_a]/k$
\\
\bottomrule
\end{tabular}

\caption{Measurement laws and the dynamics they induce. Gamma distributions are parameterized by shape and rate, and $\mathrm{NB}(a,q)$ has shape $a$ and mean $aq/(1-q)$. The noise acts
independently across pixels given $X$, and the level $a$ is written
$\lambda$, $p$, $L$, $r$ in the appendices. The last column replaces the
function of the clean image in the dynamics by its posterior prediction. The binomial image is integer-valued, and
$q(X)$ is a known pixelwise encoding with values in $[0,1)$.}
\label{tab:instances}
\end{table}

\subsection{Learning posterior predictions $h_{\xi,a}$ from noisy measurements}
\label{sec:formulation}

\paragraph{Learning at and below the training information level.}
Let $Y_{\rm in}$ and $Y_{\rm tar}$ be two noisy measurements of $X$, independent conditional on $X$, with $Y_{\rm in}$ at information level $a_{\rm in}$. For Gaussian, Poisson, Gamma and negative-binomial observations, such a pair can also be obtained by splitting a single acquisition. Each resulting view has a lower information level than the original measurement (Appendix~\ref{app:splitting}). Degrading $Y_{\rm in}$ with the known kernels $\overleftarrow{K}_{a_{\rm in},a}$ gives an input $Z_a$ at any level $a\leq a_{\rm in}$.
If the clean image were available, the natural
target associated with this input would be $\psi_\xi(X)$, since
$h_{\xi,a}(Z_a)
=
\E[\psi_\xi(X)\mid Z_a]$.
Instead, we use $Y_{\rm tar}$ to construct a noisy target $T_\xi(Y_{\rm tar})$ with the same conditional mean:
\[ 
\E[T_\xi(Y_{\rm tar})\mid X]=\psi_\xi(X).
\] 
For Poisson measurements at exposure $a_{\rm tar}$, the target is $T_\xi(Y_{\rm tar}) =Y_{\rm tar}/a_{\rm tar}$, and Appendix~\ref{app:targets} derives the targets for the other posterior predictions in Table~\ref{tab:instances}.

Since $Y_{\rm tar}$ and $Z_a$ are independent conditional on $X$, the tower property gives
\begin{equation}
\E[T_\xi(Y_{\rm tar})\mid Z_a]
=
\E[\psi_\xi(X)\mid Z_a]
=
h_{\xi,a}(Z_a).
\label{eq:noisy-target}
\end{equation}
Regressing $T_\xi(Y_{\rm tar})$ on $Z_a$ therefore recovers the needed posterior prediction. We train a network $\widehat h=(\widehat h_{\xi,a})$ to minimize
\begin{equation}
\mathcal L_{\rm N2N}(\widehat h)
=
\E_{(a,\xi)\sim \rho}\left[
\left\|
\widehat h_{\xi,a}(Z_a)-T_\xi(Y_{\rm tar})
\right\|^2
\right],
\label{eq:pf-loss}
\end{equation}
where $\rho$ samples information levels $a \le a_{\rm in}$ and indices $\xi$. If $\E_{(a,\xi)\sim \rho}\, \E\|T_\xi(Y_{\rm tar})\|^2< \infty$, the excess loss is
\[
\mathcal L_{\rm N2N}(\widehat h)
-
\mathcal L_{\rm N2N}(h) = \E_{(a,\xi)\sim \rho}
\left\|
\widehat h_{\xi,a}(Z_a)-h_{\xi,a}(Z_a)
\right\|^2,
\]
so the true predictions $h=(h_{\xi,a})$ minimize the loss and are its only minimizer up to events of probability zero (Lemma~\ref{lem:app-n2n}). For Gaussian measurements, \eqref{eq:pf-loss} is the loss of \citet{pmlr-v235-daras24a} up to a level-dependent weight (Appendix~\ref{app:gaussian-n2n}).

\paragraph{Consistency above the training information level.}
Degradation provides training inputs only at levels $a\leq a_{\rm in}$, so the N2N loss constrains $\widehat h_{\xi,a}$ only there. To approach the clean limit, the sampler also needs the predictions above $a_{\rm in}$. For $a<b$, let $\overrightarrow{K}_{a,b}$ denote the true transition from
$Z_a$ to $Z_b$, so that
\[
(\overrightarrow{K}_{a,b}f)(z)=\E[f(Z_b)\mid Z_a=z].
\]
Since $Z_a$ adds no information about $X$ once $Z_b$ is known, the tower property gives
\begin{equation}
h_{\xi,a}(z)
=
\E\!\left[h_{\xi,b}(Z_b)\mid Z_a=z\right]
=
(\overrightarrow{K}_{a,b}h_{\xi,b})(z).
\label{eq:main-consistency}
\end{equation}
Thus, the prediction at state $z$ and level $a$ equals the expected prediction at any higher information level $b$, under the true process started from $z$. In the Gaussian case this is the consistency relation of \citet{daras2023consistent}. As in that work, the true transition depends on the unknown predictions, so we replace it by the transition $\widehat{\overrightarrow{K}}_{a,b}$ of the process simulated with the current model $\widehat h$~\citep{daras2023consistent,pmlr-v235-daras24a}, and define
\begin{equation}
\mathcal L_{\rm cons}(\widehat h)
=
\E_{(a,b,\xi)\sim \rho_{\rm cons}}\left[
\left\|
\widehat h_{\xi,a}(Z_a)
-
(\widehat{\overrightarrow{K}}_{a,b} \widehat h_{\xi,b})(Z_a)
\right\|^2
\right],
\label{eq:main-consistency-loss}
\end{equation}
where $\rho_{\rm cons}$ is a distribution over levels $a<b$ with
$a\leq a_{\rm in}$ and indices $\xi$, and $Z_a$ is a training input at level
$a$, as in \eqref{eq:pf-loss}.

We first train with $\mathcal L_{\rm N2N}$, and then refine the model with
\begin{equation}
\mathcal L_{\rm N2N}(\widehat h)
+
\lambda_{\rm cons}\mathcal L_{\rm cons}(\widehat h),
\qquad
\lambda_{\rm cons}>0.
\label{eq:combined-training-loss}
\end{equation}
The regression term $\mathcal L_{\rm N2N}$ keeps predictions accurate where noisy targets are available. The consistency term relates those predictions to predictions at higher information levels. 
We estimate $\widehat{\overrightarrow{K}}_{a,b} \widehat h_{\xi,b}(z)$ from simulated paths. 
Every path starts at the training measurement $z$ at level $a=a_{\rm in}$ and is simulated with the current model in coarse steps to a level $b > a_{\rm in}$. Under exact simulation, any global minimizer matches the true predictions at the states and levels sampled during N2N training and is consistent along the simulated paths, assuming that the model class contains the true predictions $h$ (Proposition~\ref{prop:app-population-objective}). However, this alone does not guarantee that the continuation beyond $a_{\rm in}$ coincides with
the true predictions $h$. In Section~\ref{sec:continuation} we study under which conditions it does.
\section{When consistency uniquely determines the continuation}
\label{sec:continuation}\label{sec:theory}

Suppose that a model is consistent and agrees with the true predictions at one positive information level. We give conditions under which it then agrees with the true predictions at every higher level, and therefore generates the true observation process. 

\paragraph{Identification by degradation.}
Assume that at some level $a_* \in (0, a_{\rm in}]$ the N2N loss recovers
the true posterior predictions, $\widehat h_{\xi,a_*}=h_{\xi,a_*}$ for all $\xi$. The training inputs at $a_{*}$ are degraded measurements and have the true law $p_{a_{*}}= \operatorname{Law}(Z_{a_{*}})$. We start the learned process from this law.

To see what is needed to identify its continuation, suppose for the moment
that, in the learned process, the conditional law of the state at a level $a$ given the state at a level $b>a$ is the known degradation kernel $\overleftarrow K_{b,a}$. Write $\widehat p_b$ for its distribution at level $b>a_*$. Under this assumption, degrading $\widehat p_b$ to level $a_*$ recovers its starting distribution $p_{a_*}$. The true law $p_b$ has the same property, so 
\[ 
\widehat p_b\overleftarrow{K}_{b,a_*}=p_{a_*} = p_b\overleftarrow K_{b,a_*}.
\]
For our five measurement laws, the map $\nu\mapsto\nu\overleftarrow{K}_{b,a_*}$ is injective 
(Proposition~\ref{prop:app-degradation}), a property related to the classical identifiability of mixtures \citep{teicher1961identifiability}. Hence $\widehat p_b=p_b$ at every level $b> a_*$, so the distribution at a higher level is
uniquely determined by the distribution obtained after degrading it to $a_*$.  

This identifies the distribution at each level. To identify the whole
process, we also need to know how its states are related across levels.
The backward
transitions provide this information: they give the conditional
distribution of an earlier state given a later state. Thus,
\[
\operatorname{Law}(Z_a\in\dd z,Z_b\in\dd w)
=
p_b(\dd w)\overleftarrow K_{b,a}(w,\dd z),
\qquad a<b, 
\]
and since the learned process has the same marginals and backward transitions as the true process, it has the same joint distribution at any two levels. Both processes are Markov, the learned process because its dynamics depend only on the level and the current state. Applying the
backward transitions successively therefore also gives the same joint
distributions at any finite collection of levels.

Consequently, a learned process that starts from  $p_{a_*}$ and has the known degradation kernels as its backward transitions has the law of the true
observation process (Theorem~\ref{thm:app-backward-identification}). It remains to show that the consistency relation gives the learned process these backward transitions.

\paragraph{The consistency equation.}
The predictions $\widehat h$ determine the dynamics of the learned
process $(\widehat Z_a)$. The consistency relation requires the prediction at a current state to equal the expected prediction after evolving this process to any later level:
\[
\widehat h_{\xi,a}(z)
=
\bigl(\widehat{\overrightarrow K}_{a,b}
      \widehat h_{\xi,b}\bigr)(z)
=
\mathbb E\!\left[
\widehat h_{\xi,b}(\widehat Z_b)
\,\middle|\,\widehat Z_a=z
\right], \qquad a<b.
\]
Thus, along the learned process, the prediction has zero expected change
conditional on the current state.

There are two sources of change in a prediction over a small increment in information level.
First, the level increases, which contributes $\partial_a \widehat h_{\xi,a}$. Second, the state moves under the learned dynamics, which contributes $\mathcal A_a[\widehat h] \widehat h_{\xi, a}$, where $\mathcal A_a[\widehat h]$ is the generator of the learned process (Appendix~\ref{app:generators}).
Taking $b \downarrow a$ in the consistency relation therefore gives
\begin{equation}
\bigl(\partial_a+\mathcal A_a[\widehat h]\bigr)
\widehat h_{\xi,a}=0,
\qquad a_*<a<b_*,
\label{eq:continuation-equation}
\end{equation}
where $b_*>a_*$ is the highest level considered.
This equation is nonlinear because $\widehat h$ appears both in the generator and in the function it acts on. Under our assumption, the N2N loss recovers the true predictions at $a_*$. The question is whether the value of $\widehat h$ at $a_*$ determines a unique solution of \eqref{eq:continuation-equation} at all higher levels. We give two arguments. The first assumes that all learned predictions derive from a common density ratio. The second proves uniqueness directly for the four jump processes, without requiring this representation.

\paragraph{Predictions from a density ratio.}
The first argument covers all five noise models, but it assumes that all predictions come from one function. We fix a reference process that does not depend on $p_\mathrm{clean}$, such as Brownian motion for Gaussian and independent unit-rate Poisson processes for Poisson observations, and let $p^0_a$ denote its distribution at level $a$. The true observation law has a density ratio $F_a$ relative to $p^0_a$  \[
p_a(\dd z)=F_a(z)\,p^0_a(\dd z),
\] 
and the true posterior predictions can be expressed through this function. In the Gaussian and Poisson cases, for example, 
\[
m_a(z)=\nabla_z\log F_a(z),
\qquad
h_{i,a}(n)=\frac{F_a(n+e_i)}{F_a(n)},
\]
where $e_i$ is the image with one count at pixel $i$ and zeros elsewhere.
Our first argument assumes that the learned predictions have these representations, with a single function $\widehat F_a$ in place of $F_a$. Under Assumption~\ref{ass:app-common}, we can rescale $\widehat F_a$ by a factor depending only on $a$ so that substituting the representations into \eqref{eq:continuation-equation} gives a linear equation for the rescaled $\widehat F_a$. Appendix~\ref{app:common} shows that this equation makes the learned process, initialized with $p_{a_*}$, a reweighting of the reference process that preserves its backward transitions, which are the known degradation kernels. The preceding identification argument therefore implies that the learned process is the true observation process.

\paragraph{Direct predictions.} However, our networks, like those of
\citet{pmlr-v235-daras24a} for Gaussian diffusions, predict the required posterior quantities separately, without forcing them to come from a common density ratio $\widehat F$. For Poisson, binomial, Gamma and negative-binomial observations, our second argument does not assume a common density ratio. Theorem~\ref{thm:main-continuation} shows that if
the learned and true predictions satisfy the same consistency equation and
coincide at $a_*$, then they agree at all higher levels, provided that both obey the bounds of Assumption~\ref{ass:app-direct}. The proof is a Gr\"onwall-type estimate on their difference.

For the true predictions, these bounds hold whenever the pixel values are bounded, for Gamma noise also away from zero, and, for negative-binomial noise, $q(X)$ is bounded away from one (Appendix~\ref{app:direct-conclusion}). 

\begin{theorem}
\label{thm:main-continuation}
Fix $0<a_*<b_*<a_{\max}$. Suppose that
$\widehat h_{\xi,a_*}=h_{\xi,a_*}$ for every required prediction and that
$\widehat h$ satisfies~\eqref{eq:continuation-equation} on $(a_*,b_*)$ in the
sense of Appendices~\ref{app:common} and \ref{app:direct}. Assume one of the following. 
\begin{enumerate}
    \item All predictions
are derived from one common density ratio satisfying Assumption~\ref{ass:app-common}.
    \item The observations are Poisson, binomial, Gamma or negative binomial, and $X$ and the learned predictions satisfy Assumption~\ref{ass:app-direct}: bounded pixel values (for Gamma also bounded away from zero, for negative binomial $q(X)$ bounded away from one), bounded predicted rates (geometrically decaying predicted moments for negative binomial) and, for Gamma, predicted Laplace transforms that decay in the jump size and are Lipschitz in the state.
\end{enumerate} 
Then $\widehat h_{\xi,a}=h_{\xi,a}$ for $ a_*\leq a\leq b_*$ and every required prediction. Further, the process generated by
$\widehat h$ and initialized with $p_{a_*}$ has the law of the true
observation process on $[a_*,b_*]$.
\end{theorem}

\paragraph{From the training losses to the theorem.}
An exact global minimizer of \eqref{eq:combined-training-loss} satisfies $\widehat h_{\xi,a_*}=h_{\xi,a_*}$, the first hypothesis of Theorem~\ref{thm:main-continuation}. The second hypothesis, the consistency equation, requires consistency between arbitrary levels $a<b$ above $a_{\rm in}$, whereas our loss compares $a_{\rm in}$ only with later levels, since every path starts at $a_{\rm in}$. 

For the four non-Gaussian laws, whose processes move by jumps, starting simulated paths at $a_{\rm in}$ is enough. For the count models (Poisson, binomial and negative binomial) with bounded rates, the learned transition from $a_{\rm in}$ to any level $a$ is one-to-one on bounded functions, and for Gamma on bounded Lipschitz functions, so it cannot hide an inconsistency: predictions that are consistent from $a_{\rm in}$ to both $a$ and a later level $b$ are also consistent from $a$ to $b$ (Lemmas~\ref{lem:app-count-injective} and~\ref{lem:app-gamma-injective}). For the Gaussian process, we follow \citet{daras2023consistent, pmlr-v235-daras24a} and start paths at every level (Remark~\ref{rem:app-all-starts}).
Under these conditions an exact minimizer has the true predictions at every level in $[a_{\rm in},b_*]$ (Corollaries~\ref{cor:app-count-training} and~\ref{cor:app-gamma-training}).

Our experiments do not meet all of these conditions (Section~\ref{sec:limitations}).

\section{Experiments}\label{sec:experiments}

We generate images by simulating the learned process from $Z_0 = 0$ and evaluate it at two stopping levels. To assess N2N training, we stop at the information level of the training data, \(a_{\mathrm{in}}\), and return the predicted posterior mean of the clean image. To assess continuation beyond the training level, we continue past $a_{\rm in}$, through levels supervised only by the consistency loss, to the sampler’s final level, and return the final state or the posterior mean predicted from it (Table~\ref{tab:app-settings} in Appendix~\ref{app:networks-loss}). We compare the two evaluations on CIFAR-10 and on the \textsc{Conv-PC} photon-counting microscopy data. On radar data, we test whether the learned process, conditioned on the measured image, captures dependence between pixels that independent per-pixel sampling misses (Section~\ref{sec:experiments-radar}).

\subsection{Generation from noisy CIFAR-10 images}

For CIFAR-10, we corrupt each training image with Poisson, binomial, Gamma and negative-binomial noise, drawing two measurements with independent noise, and train a U-Net with 60M parameters to output the posterior predictions of Table~\ref{tab:instances}. We vary $a_{\rm in}$, which changes how much of the trajectory the consistency loss must supervise. Both measurements have the same information level, $a_{\rm tar}=a_{\rm in}$, which we write as $L_{\rm obs}$ for Gamma noise and as $r_{\rm obs}$ for negative-binomial noise (Table~\ref{tab:instances}). Both measurements are redrawn every epoch, so during training the network sees many noisy measurements of each training image, but never a clean one. The microscopy and radar experiments below use fixed measurements. 

Table~\ref{tab:exp-cifar-generation} reports the Fr\'echet inception distance \citep[FID;][]{heusel2017gans} for samples stopped at $a_{\rm in}$ and for continued samples after $200$ further epochs of training, with and without the consistency loss. Without it ($\lambda_{\rm cons}=0$) samples are poor, with FID between $31$ and $182$, and the Gamma sampler diverges. This is expected because the sampler
uses the network at levels where it was never trained. With the consistency loss ($\lambda_{\rm cons}>0$), FID drops by a factor
of $5.3$--$14.7$. The two continuation columns
differ only in the consistency weight $\lambda_{\rm cons}$, so the improvement is due to the consistency term.

When the consistency loss is used, the continued samples also improve on stopping at the training level, the best the Noise2Noise model gives without continuing. They do so in eight of the eleven settings, match it at $100.0$ binomial counts and at $r_{\rm obs}=50$, and are worse only at $180.1$ binomial counts, the binomial setting closest to the clean image.
\begin{table}[t]
    \centering
    \small
    \setlength{\tabcolsep}{6pt}
    \begin{tabular}{llrrr}
        \toprule
        & & & \multicolumn{2}{c}{Continue past $a_{\rm in}$} \\
        \cmidrule(lr){4-5}
        Process & Training Measurement & Stop at $a_{\rm in}$ & \multicolumn{1}{c}{No Consistency} & Consistency \\
        \midrule
        Poisson  & 613.1 counts    & $7.30 \pm 0.60$ & $30.66 \pm 0.03$ & $3.98 \pm 0.13$ \\
        Poisson  & 178.4 counts    & $9.06 \pm 0.06$ & $63.80 \pm 9.24$ & $4.34 \pm 0.03$ \\
        Poisson  & 16.1 counts     & $17.23 \pm 0.44$ & $109.14 \pm 11.19$ & $8.19 \pm 0.30$ \\
        \midrule
        Binomial & 180.1 counts    & ${9.41 \pm 0.06}$ & $61.17 \pm 2.48$ & $11.65 \pm 0.12$ \\
        Binomial & 100.0 counts    & $11.68 \pm 0.51$ & $95.23 \pm 7.47$ & ${11.63 \pm 1.65}$ \\
        Binomial & 25.1 counts     & $16.18 \pm 0.24$ & $152.38 \pm 12.55$ & ${13.23 \pm 0.61}$ \\
        Binomial & 8.0 counts      & $22.70 \pm 0.12$ & $181.55 \pm 8.38$ & ${19.68 \pm 0.15}$ \\
        \midrule
        Neg.\ binomial & $r_{\rm obs}=5$  & $19.99 \pm 0.17$ & $95.63 \pm 1.29$ & $14.22 \pm 0.96$ \\
        Neg.\ binomial & $r_{\rm obs}=16$ & $13.54 \pm 0.18$ & $79.53 \pm 4.85$ & $10.99 \pm 0.42$ \\
        Neg.\ binomial & $r_{\rm obs}=50$ & $7.78 \pm 0.15$ & $65.58 \pm 2.55$ & $7.64 \pm 0.03$ \\
        Gamma    & $L_{\rm obs}=64$ & $9.32 \pm 0.06$ & \multicolumn{1}{r}{---\hspace*{6mm}} & $7.01 \pm 0.27$ \\
        \bottomrule
    \end{tabular}
    \caption{CIFAR-10 FID from 50,000 EMA samples against the clean training set, mean $\pm$ standard deviation over two training seeds. All models train for 2000 epochs, where runs with the consistency loss use positive weight ($\lambda_{\rm cons}>0$) for the last 200. The values for stopping at $a_{\rm in}$ are the predicted posterior means of the $\lambda_{\rm cons}=0$ model, listed for reference. Counts are expected counts at a white pixel. Rescoring one checkpoint with a fresh set of 50,000 samples moves FID by $0.143$. Without the consistency loss the Gamma trajectory diverges~(---).}
    \label{tab:exp-cifar-generation}
\end{table}

\paragraph{Misspecifying the measurement law.}
To test whether the process must follow the noise model, we train two Gaussian diffusion baselines on the same Poisson measurements, with the same architecture and exposure. One assumes a single noise variance for all pixels, and the other uses an Anscombe transform. Both stop at the noise level of the training measurements and return the predicted posterior mean. At the two higher photon counts, their FIDs exceed that of the Poisson model by $11.0$--$17.5$ when all three are trained for 1800 epochs. At the lowest count, their FIDs are more than six times as large (Table~\ref{tab:exp-cifar-misspec} in Appendix~\ref{app:cifar-counts}). 

The negative-binomial jump rates require predictions of all posterior moments $\E[q_i(X)^k\mid N_a]$, instead of only the posterior mean. We encode the image intensity by $q_i(X)=0.9X_{i}$ and replace the predicted moments by powers of the predicted mean. Sampling from this misspecified process instead of the correct one raises FID from $9.61$ to $18.50$, with both processes stopped at $r_{\rm obs}=50$ (Appendix~\ref{app:cifar-counts}).

\subsection{Photon-counting microscopy}

The \textsc{Conv-PC} dataset consists of photon-counting frames of a \textit{Convallaria} tissue section, 512 for each of four fields of view, each frame with 0.04 to 0.07 photons per pixel on average \citep{krull2024image}.
No clean image is available, and the only reference is the sum of the 512 frames of a field of view, which is itself a
count image with 512 times the exposure of one frame. 
Generative Accumulation of Photons (GAP), introduced together with this dataset, trains a network to predict the complement of a binomially sampled subset of each frame. To generate an image, GAP repeatedly draws photons from the network's predictions, adding at each step a fraction $\beta$ of the photons already present. 

We train with the same splitting procedure, so supervision ends at the exposure of one frame. We simulate the Poisson observation process using predicted rates and supervise the continuation beyond one frame with the consistency loss. Following the GAP protocol, training includes frames from the field of view used for evaluation. 
 
\begin{figure}[t]
    \centering
    \includegraphics[width=\linewidth]{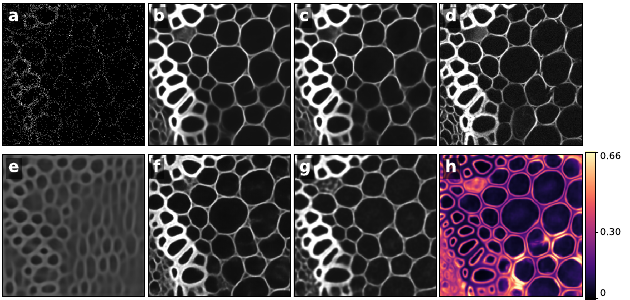}
    \caption{Reconstructions and samples from one measured microscopy frame.
    (a) The input: 5{,}020 photons in $256\!\times\!256$ pixels.
    (b, c) Predicted images from N2N and GAP.
    (d) The 512-frame reference.
    (e) N2N continuation to the reference exposure without the consistency loss.
    (f) GAP continuation under its own accumulation rule, which stops at 0.32 photons per pixel.
    (g) N2N continuation to the reference exposure after training with the consistency loss.
    (h) Pixelwise standard deviation divided by mean over 500 continuations sampled after training with the consistency loss, on a color scale from 0 to 0.66.
    Panels (b)--(g) are divided by their means and clipped at four.}
    \label{fig:exp-convpc-main}
    \label{fig:exp-convpc-continuation}
\end{figure}

Table~\ref{tab:exp-convpc-generation} compares stopping at one frame with continuing to the reference exposure of 512 frames. We retrain GAP for the comparison. Stopped at the exposure of one frame, our model matches GAP at $256\!\times\!256$ but gives a higher FID at $64\!\times\!64$ ($114.1$ against $97.2$ and $98.1$).
Continuing to the reference exposure without the consistency loss produces unusable images with FID $267.4$ and $339.4$: the example in Figure~\ref{fig:exp-convpc-main}(e) loses the tissue's sharp cell boundaries. In contrast, after twenty epochs of training with the consistency loss, samples such as (g) are sharper and agree better with the measurement displayed in (a). 
Three training runs with the consistency loss, all started from the same checkpoint, give FID $44.7\pm2.5$ at $256\!\times\!256$. Every run outperforms GAP for both values of $\beta$. At $64\!\times\!64$, our method gives lower FID than GAP under two of the three intensity renderings tested (Appendix~\ref{app:microscopy}).

\begin{table}[t]
    \centering
    \small
    \begin{tabular}{lrr}
        \toprule
        Method & FID $64^2$ & FID $256^2$ \\
        \midrule
        GAP ($\beta=0.1$) & 98.1 & 58.4 \\
        GAP ($\beta=0.05$) & 97.2 & 55.8 \\
        N2N, stop at one frame & 114.1 & 55.7 \\
        N2N, continue to 512 frames & 267.4 & 339.4 \\
        N2N $+$ consistency loss, continue to 512 frames & ${93.3\pm2.5}$ & ${44.7\pm2.5}$ \\
        \bottomrule
    \end{tabular}
    \caption{Generation on \textsc{Conv-PC}: FID from 10,000 images of $256\!\times\!256$ pixels per method, and from one $64\!\times\!64$ crop of each, against crops of the same sizes from the 512-frame sums. Every method returns a predicted image. N2N sampling starts from zero counts and stops at the indicated exposure. GAP follows its own accumulation rule. The consistency row gives mean $\pm$ standard deviation over three trainings started from one checkpoint and supervised by the consistency loss. Other rows are single runs.}
    \label{tab:exp-convpc-generation}
\end{table}
\subsection{Synthetic aperture radar}
\label{sec:experiments-radar}

A radar intensity image contains speckle from interference between scattered waves. Averaging independent intensity measurements, called \emph{looks}, reduces this noise. Under the Gamma model, the sum of $L$ looks satisfies
\[
S_L\mid X\sim\operatorname{Gamma}(L,1/X),
\qquad
\E[S_L/L\mid X]=X,
\qquad
\operatorname{Var}(S_L/L\mid X)=X^2/L,
\]
with shape $L$ and rate $1/X$ for each pixel, where the clean intensity $X$ is called the \emph{reflectivity}.  We use three satellite radar acquisitions from Sentinel-1. Following \citet{dalsasso2021if}, we obtain two conditionally independent measurements with shape $\Lobs=1/2$ from the squared real and imaginary components, each being a \emph{half-look}. To reduce correlation in the speckle between neighboring pixels, we separate each image into four interlaced sublattices (Appendix~\ref{app:radar}). N2N can supervise the predictions up to half a look, and the consistency loss supervises their continuation to larger $L$.

The network predicts a four-component inverse-Gamma mixture for the posterior of each pixel. For each jump size $w$, we use the predicted expectation of $e^{-w/X_i}$ to compute the jump rate. These predictions give the marginal posterior of each pixel. Because neighboring pixels of the clean image are dependent, they are also dependent in the joint posterior, and sampling each pixel from its marginal ignores this dependence. When we simulate the learned process, the prediction at each pixel depends on the whole current image, so the simulated process can reproduce it.

We compare two samplers that start from the same observed half-look. The \emph{independent sampler} draws a reflectivity at each pixel from its predicted marginal posterior, then draws an independent half-look conditional on these reflectivities. The \emph{joint sampler} simulates the learned process from $\Lobs$ to $2\Lobs$, where its increment is a new half-look. We evaluate both samplers against the second half-look of held-out patches, without a clean reference.

The two samplers give similar pixelwise coverage for the held-out view. To test spatial dependence, we sum the intensities over $16\!\times\!16$ blocks and compare predictive intervals for these sums with the held-out block sums. The variance of a sum includes the covariances between its pixels, which the independent sampler sets to zero. The nominal $90\%$ intervals cover $88.6$--$89.2\%$ of the held-out block sums for the joint sampler, compared with $82.5$--$83.2\%$ for the independent sampler across two seeds (Table~\ref{tab:app-radar-blocks} in Appendix~\ref{app:radar}). Only the joint sampler is close to nominal coverage at this scale.

\begin{figure}[t]
    \centering
    \includegraphics[width=\linewidth]{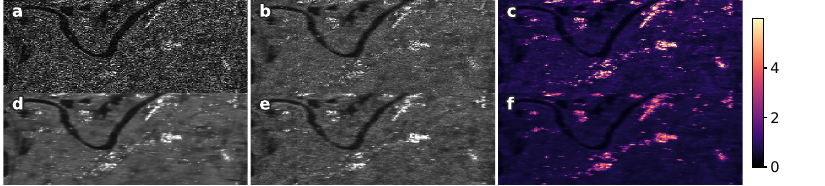}
    \caption{Conditional radar samples at $L=16$ ($32\Lobs$).
    (a) Observed half-look. (d) Posterior mean.
    (b, e) An independent and a joint sample.
    (c, f) Standard deviations over 50 draws, with a shared color bar.
    The lower speckle level makes the sampled spatial structure visible. Independent draws vary from pixel to pixel, whereas the joint samples vary in patches. The joint sampler underestimates the absolute spread beyond $16\Lobs$ (Appendix~\ref{app:radar}). Images are shown at ground aspect, with range along the short axis.}
    \label{fig:exp-sar-marginal-vs-joint}
\end{figure}

The difference is also visible after continuing to $32\Lobs$ (Figure~\ref{fig:exp-sar-marginal-vs-joint}). Independent sampling breaks bright structures into isolated pixels, whereas joint sampling produces coherent patches. At these longer continuations, however, the learned process underestimates the spread of the intensities (Appendix~\ref{app:radar}).

To stabilize predictions at the training level, we use an additional regression term. Starting from an observed half-look, we simulate to a higher information level and regress the prediction there onto a target from the second half-look. This term does not supervise the process above $L_{\rm obs}$, because its optimum is the posterior given the starting half-look. An ablation in Appendix~\ref{app:radar} (Table~\ref{tab:app-radar-ablation}) separates the effect of the additional regression term from that of the consistency loss and shows that using the consistency loss is needed for the simulated trajectory to realize the network's predicted posterior spread beyond the information level of the training data.

\FloatBarrier

\section{Conclusion}\label{sec:limitations}
We learn generative models of clean images from noisy measurements by building the
generative process from the measurement law. Noise2Noise regression supervises it up to the information level covered by the data, and the consistency relation determines its continuation toward clean images. In the CIFAR-10 and photon-counting microscopy experiments, training with the consistency loss improves sample quality when sampling continues above the information level of the training data. On radar data, the learned sampler captures posterior dependence between pixels, which independent per-pixel sampling misses. 

Our theory gives mild conditions under which this continuation is uniquely determined. Under the conditions of Section~\ref{sec:continuation}, an exact minimizer of the training losses averaged over the true distribution
of noisy measurements has the true predictions and therefore generates the true observation process.
\newpage
\paragraph{Limitations.}
The theory concerns exact minimizers and exact simulation, and our experiments depart from it in four ways. To reduce the cost per training step, the paths for the consistency loss are simulated with coarse steps. The networks enforce only some of the bounds of Assumption~\ref{ass:app-direct}, the negative-binomial networks predict finitely many moments, and the radar model adds a regression on simulated states to stabilize its predictions. The resulting errors propagate along the continuation, and the stability estimate in Remark~\ref{rem:app-stability}, which bounds this propagation, grows exponentially with the image size and the length of the continuation. In practice, continuing does not always help. At $180.1$ binomial counts on CIFAR-10, the continued samples have a higher FID than samples stopped at the information level of the training data, and on radar data the learned process underestimates the spread of the intensities beyond $16L_{\rm obs}$. We have no criterion for how far a learned continuation should be run.

Because we build the generative process from the measurement law, the method needs that law, including the information level of the data. Using the wrong law is costly: on the same Poisson data, Gaussian diffusion models give FIDs higher by $11.0$--$17.5$ at the two higher photon counts. We also assume that the noise is independent across pixels given the clean image. Raw radar images do not satisfy this, since their speckle is correlated between neighboring pixels, and we split each image into four sublattices to reduce the correlation.

\newpage
\begingroup
\interlinepenalty=10000
\bibliographystyle{bibstyle}
\bibliography{refs}
\endgroup

\newpage

\appendix

\section{Guide to the appendix}
\label{app:guide}
Appendix~\ref{sec:related} relates the method to denoising from noisy data,
diffusion models trained on corrupted data, generative processes beyond
Gaussian noise, and the consistency loss of \citet{daras2023consistent}.

Appendix~\ref{app:models} defines the five observation processes considered
in the paper. For each model, we specify how observations at different
information levels are related, which posterior quantities determine the generative process, and which noisy targets can be used to learn
these quantities. We also prove an identification result: once the
distribution of the observation is known at any positive information level,
there is at most one compatible distribution at every higher information
level.

Appendices~\ref{app:common} and \ref{app:direct} then address the main
continuation question. They give conditions under which posterior predictions
that are correct at a positive level $a_*$ and consistent above it equal
the true predictions at every higher level. Appendix~\ref{app:common} treats the case where
all posterior predictions are obtained from one common density ratio.
Appendix~\ref{app:direct} treats the setting used in our experiments, where
the network predicts the required quantities directly.

Appendix~\ref{app:training} connects these continuation results to the
training objective. It states conditions under which an exact minimizer of
the expected regression and consistency losses satisfies the hypotheses
required by the continuation theorems. Finally,
Appendix~\ref{app:sampling} describes the approximations introduced by
numerical sampling, and Appendix~\ref{app:experiments} gives the experimental
details and additional results.

We use the notation of the main text throughout. The kernel
$\overleftarrow K_{b,a}$ takes an observation at a more informative level
$b$ and degrades it to a less informative level $a<b$. In the opposite
direction, $\overrightarrow K_{a,b}$ denotes the true forward transition of
the generative process. A hat denotes a learned quantity. A superscript $0$
denotes a reference process introduced below. In every model, an observation at a
lower level can be generated from one at a higher level with
$\overleftarrow K$, which does not depend on the clean image, so higher
levels are more informative.

\begin{table}[htbp]
\centering
\small
\begin{tabular}{@{}p{0.49\linewidth}p{0.45\linewidth}@{}}
\toprule
Result & Location \\
\midrule
Theorem~\ref{thm:main-continuation} & Item 1: Corollary~\ref{cor:app-common-continuation}.
Item 2: Corollary~\ref{cor:app-direct-continuation} on $[a_*,b_*]$ \\
Noisy-target regression & Lemma~\ref{lem:app-n2n} and
Proposition~\ref{prop:app-gamma-idiv} \\
Injectivity of degradation & Proposition~\ref{prop:app-degradation} \\
Identification from a positive-level marginal &
Theorem~\ref{thm:app-backward-identification} \\
Continuation from a common density ratio &
Corollary~\ref{cor:app-common-continuation} \\
Uniqueness for direct predictions &
Theorem~\ref{thm:app-direct-uniqueness} and
Corollary~\ref{cor:app-direct-continuation} \\
Poisson and binomial rates with linear growth &
Theorem~\ref{thm:app-linear-growth} \\
Exact minimizers of the training objective &
Proposition~\ref{prop:app-population-objective} \\
Training from one positive starting level &
Corollaries~\ref{cor:app-count-training} and \ref{cor:app-gamma-training} \\
Poisson generation from zero & Corollary~\ref{cor:app-poisson-zero} \\
\bottomrule
\end{tabular}
\caption{Main statements proved in the appendix and where they are located.}
\label{tab:app-proof-guide}
\end{table}

\begin{table}[htbp]
\centering
\small
\begin{tabular}{@{}p{0.33\linewidth}p{0.61\linewidth}@{}}
\toprule
Symbol & Meaning \\
\midrule
$X\sim p_{\mathrm{clean}}$ & Clean image and its distribution \\
$d$, $e_i$ & Number of pixels, and the image that is one at pixel $i$ and zero elsewhere \\
$a$; $\lambda,p,L,r$ & Generic information level; Poisson exposure,
binomial detection probability, Gamma shape, and negative-binomial shape \\
$Z_a$, $Q_a(x,\mathrm dz)$, $p_a$ & Observation at level $a$, its
conditional law given $X=x$, and its marginal law \\
$N_a$, $S_a$ & The observation $Z_a$ for count and for Gamma observations \\
$\overleftarrow K_{b,a}$ & Known degradation kernel from $b$ to $a<b$ \\
$\overrightarrow K_{a,b}$,
$\widehat{\overrightarrow K}_{a,b}$,
$\overrightarrow K^0_{a,b}$ & True, learned, and reference forward transitions \\
$p^0_a$, $F_a$, $\widehat F_a$, $G_a$ & Reference marginal, true and candidate
density ratios, and a rescaled candidate used in the continuation proof \\
$\xi\in\Xi$, $\psi_\xi(X)$, $h_{\xi,a}$ & Index of a required prediction and
the set of these indices, the function of the clean image that the dynamics
need, and its posterior prediction at level $a$ \\
$m_a,h_{i,\lambda},D_{i,p},H_{i,L}(\cdot,w),M_{i,k,r}$ & Gaussian posterior mean,
Poisson posterior mean intensity, binomial posterior mean, Gamma posterior Laplace
transform, and negative-binomial posterior moment \\
$u_{\xi,a}$, $u_{i,a}$ & Jump rates in the consistency equation
\eqref{eq:app-direct-equation} shared by the four jump models, and the birth rate of coordinate $i$ for
Poisson and binomial observations \\
$\mathcal A_a[h]$, $\mathcal A^0$ & Generator of the process driven by
predictions $h$, and generator of the reference process \\
$a_*$, $a_{\mathrm{in}}$, $I=[a_*,b_*]$ & Positive level where the predictions
are already identified, information level of the training data, and a finite continuation interval \\
$Y_{\mathrm{in}}$, $Y_{\mathrm{tar}}$, $a_{\mathrm{tar}}$, $T_\xi$ & Input and target measurements,
the level of the target, and a target transformation with
$\mathbb E[T_\xi(Y_{\mathrm{tar}})\mid X]=\psi_\xi(X)$ \\
$\mathcal S$ & Count states that can occur with positive probability \\
$\mathcal L_{\mathrm{N2N}}$, $\mathcal L_{\mathrm{cons}}$,
$\lambda_{\mathrm{cons}}$ & Regression loss, consistency loss, and its
positive weight \\
$\mathcal R_\xi(a,b,z)$ & Consistency residual: the prediction at level $a$
minus its expectation after continuing the learned process to level $b$ \\
$v_\eta$, $\Lambda$, $\mathcal V$, $\|\cdot\|_\mu$ & Jump increment,
measure over possible jumps, right-hand side of the consistency equation,
and weighted supremum norm \\
$\|\cdot\|$, $\|\cdot\|_1$, $\|\cdot\|_\infty$ & Euclidean, $\ell^1$, and
supremum norms; $|\cdot|$ denotes an absolute value \\
\bottomrule
\end{tabular}
\caption{Notation used in the appendix. The negative-binomial probability parameter is $q(X)$ for the observations and $q_0$ for the reference process.}
\label{tab:app-notation}
\end{table}

\begin{table}[htbp]
\centering
\footnotesize
\setlength{\tabcolsep}{4pt}
\begin{tabular}{@{}p{0.13\linewidth}p{0.25\linewidth}p{0.27\linewidth}p{0.27\linewidth}@{}}
\toprule
Model & Common density ratio & Direct predictions & Training from one level \\
\midrule
Gaussian & Any image law, under
Assumption~\ref{ass:app-common} & Not treated here & Not treated here;
see Remark~\ref{rem:app-all-starts} \\
Poisson & Any image law & Bounded rates and bounded images;
alternatively, linearly growing rates & Bounded rates,
Corollary~\ref{cor:app-count-training} \\
Binomial & Any image law, with $b_*<1$ & Bounded counts;
alternatively, linearly growing rates & Bounded rates on $b_*<1$,
Corollary~\ref{cor:app-count-training} \\
Gamma & Joint Laplace representation and
Assumption~\ref{ass:app-common} & Bounded positive image values,
Lipschitz dependence on the state, and a bound on large jumps &
In addition, the integrability and Lipschitz conditions in
Corollary~\ref{cor:app-gamma-training} \\
Negative binomial & Any image law & $q_i(X)\leq\bar q<1$ and
geometrically bounded posterior moments & Every moment index is covered,
Corollary~\ref{cor:app-count-training} \\
\bottomrule
\end{tabular}
\caption{Scope of the continuation results. These are exact statements
about the true posterior predictions and expected losses. Starting the Gamma
process from zero additionally requires the initialization assumption in
Appendix~\ref{app:model-gamma}.}
\label{tab:app-scope}
\end{table}

\FloatBarrier
\section{Related work}\label{sec:related}

\paragraph{Denoising from noisy data.}
Noise2Noise \citep{pmlr-v80-lehtinen18a} regresses one noisy measurement
of an image on another and learns the posterior mean of the image given the
input. Noisier2Noise and recorrupted-to-recorrupted training
\citep{moran2020noisier2noise,pang2021recorrupted} obtain the two measurements
from a single Gaussian one, data thinning and fission
\citep{neufeld2024data,leiner2025data} split a single Poisson, Gamma or
negative-binomial measurement into independent parts, and generalized
recorrupted-to-recorrupted training \citep{monroy2025generalized} gives the
split for exponential families together with its guarantee.
Noise2Score \citep{kim2021noise2score} reaches the same posterior mean through
Tweedie's formula. Blind-spot methods such as Noise2Void and Noise2Self
\citep{krull2019noise2void,batson2019noise2self} learn from single noisy images
by predicting each pixel from the others, which requires noise that is
independent across pixels. For radar speckle, MERLIN \citep{dalsasso2021if} uses
the real and imaginary parts of one acquisition as two views, and SAR2SAR
\citep{dalsasso2021sar2sar} applies Noise2Noise to acquisitions of the same
scene at different dates. All of these methods train a denoiser at the noise
level of the data. We use their splitting rules to build the N2N loss at every level the
data can be degraded to, and for predictions other than the posterior mean.

\paragraph{Diffusion models from corrupted data.}
Ambient diffusion \citep{daras2023ambient}, GSURE training
\citep{kawar2024gsurebased}, SURE-Score \citep{aali2023solving} and
expectation-maximization schemes
\citep{rozet2024learning,bai2024expectation,hosseintabar2025diffem} train
Gaussian diffusion models from linearly corrupted or noisy data, and Ambient
Diffusion Omni \citep{daras2026ambient} uses low-quality images to supervise
the noisier levels of a Gaussian model. \citet{pmlr-v235-daras24a} showed
that data at one Gaussian noise level supervise the denoisers at all noisier
levels, and that the consistency loss of \citet{daras2023consistent} extends them to
the cleaner ones. \citet{levac2026normalizationequivariant} obtain the extension to lower
noise levels from the scale equivariance of Gaussian denoisers. Inverse flows
\citep{pmlr-v267-zhang25dm} and consistent stochastic interpolants
\citep{modi2026generative} learn a transport from corrupted to clean data for
continuous noise and for black-box corruptions. In all of these the sampling
process is a Gaussian diffusion or a continuous interpolant. Our sampling
process is the measurement process itself. Its states are photon counts or
speckled intensities, and a run started from a measurement samples from the
posterior given that measurement.

\paragraph{Generative processes beyond Gaussian noise.}
Binomial diffusion \citep{sohl2015deep}, Blackout Diffusion
\citep{santos2023blackout}, thinning and thickening of count images
\citep{chen2023learning}, star-shaped and Beta diffusion
\citep{okhotin2023star,zhou2023beta} and count bridges \citep{fishman2026count}
define generative processes through non-Gaussian laws, Generator Matching \citep{holderrieth2025generator} does so for arbitrary Markov processes, and stochastic localization \citep{montanari2023sampling} gives the accumulation
view of the Gaussian and Poisson processes that we use.
\citet{fishman2026count} also train count bridges from aggregated counts by expectation maximization. The other methods are trained on clean data. The generative accumulation of photons
\citep{krull2024image} is trained on noisy microscopy frames by photon
splitting and generates from an empty canvas, but it has no rule for the
predictions at photon counts beyond those of the training frames. We use the
Poisson and binomial processes of \citet{chen2023learning} and
\citet{santos2023blackout} and construct the Gamma and negative-binomial ones
in the same way. Beyond these constructions, we train from measurements at an interior level of the process, construct noisy targets for predictions other than the mean, and give conditions under which the consistency relation determines the continuation.

\paragraph{Consistency.}
\citet{daras2023consistent} introduced the consistency loss to reduce sampling drift
in Gaussian diffusion models trained on clean data, and proved that a
consistent denoiser whose score is a gradient field is the denoiser of some
image distribution. We keep the loss of \citet{daras2023consistent} and
similarly estimate it from two simulated paths (Appendix~\ref{app:networks-loss}). When all
predictions derive from one density ratio, the analog of their gradient-field condition, we prove that the consistency relation and the correct predictions at one level identify the process for all five measurement laws
(Appendix~\ref{app:common}). For the four jump laws we prove that the consistency relation determines the continuation for rates predicted directly, with paths that start at the level of the data, which is how the models are trained.
\section{The five measurement laws}
\label{app:models}

Let $X\sim p_{\mathrm{clean}}$ denote the clean image. Conditional on
$X$, the measurement noise is independent across coordinates.

We write $e_i$ for the $i$th coordinate vector,
$\|n\|_1=\sum_i n_i$, $n!=\prod_i n_i!$, and
$\mathbb N=\{0,1,\ldots\}$. Gamma distributions are parameterized by
shape and rate. Thus
$\operatorname{Gamma}(L,1/x)$ has shape $L$, rate $1/x$, and mean $Lx$.

We use $a$ for the information level in statements that cover several
measurement models and in the continuation proofs. Where an argument
concerns a single model, we use its customary parameter and write
$\lambda=a$ for Poisson exposure, $p=a$ for binomial detection probability,
$L=a$ for Gamma shape, and $r=a$ for negative-binomial shape. The information
level can grow without bound in the Gaussian, Poisson, Gamma, and
negative-binomial models. In the binomial model it instead approaches the
maximum value $p=1$. We therefore write $a_{\max}=1$ for binomial
observations and $a_{\max}=\infty$ for the other four models.

The degradation kernels introduced in the main text allow observations at
different levels to be placed on one joint probability space. For example,
if $a'<a<b$, we define the joint law of
$(X,Z_b,Z_a,Z_{a'})$ by
\begin{equation}
\begin{split}
&p_{\mathrm{clean}}(\mathrm dx)Q_b(x,\mathrm dz_b)
\overleftarrow K_{b,a}(z_b,\mathrm dz_a)
\overleftarrow K_{a,a'}(z_a,\mathrm dz_{a'}).
\end{split}
\label{eq:app-joint-law}
\end{equation}
The identities
\[
Q_b\overleftarrow K_{b,a}=Q_a, \qquad \overleftarrow K_{c,b}\overleftarrow K_{b,a} =\overleftarrow K_{c,a}, \qquad a<b<c
\]
say that this construction is consistent: degrading from $b$ to $a$
produces the correct measurement law at $a$, and degrading in one step or
through an intermediate level gives the same result. We write
\[
p_a=p_{\mathrm{clean}}Q_a
\]
for the marginal distribution of $Z_a$.

\subsection{Density ratios and generators}
\label{app:generators}

The proofs compare the observation process with a process that depends only
on the measurement law and not on the unknown clean-image distribution. We call this the \emph{reference
process}. Let $p^0_a$ denote its distribution at level $a$ and let
$\overrightarrow K^0_{a,b}$ denote its forward transition from $a$ to $b$.

For a fixed clean image $x$, let
\[
L_a^x(z)=\frac{\mathrm dQ_a(x,\cdot)}{\mathrm dp^0_a}(z)
\]
be the likelihood ratio of the measurement law relative to this reference.
Averaging this likelihood ratio over the unknown clean-image distribution
gives
\[
F_a(z)=\int L_a^x(z)\,p_{\mathrm{clean}}(\mathrm dx).
\]
The basic identities are
\begin{equation}
\begin{aligned}
L_a^x(z)&=\frac{\mathrm dQ_a(x,\cdot)}{\mathrm dp^0_a}(z),
&F_a(z)&=\int L_a^x(z)\,p_{\mathrm{clean}}(\mathrm dx),\\
p_a(\mathrm dz)&=F_a(z)p^0_a(\mathrm dz),
&\pi_{a,z}(\mathrm dx)&=
\frac{L_a^x(z)p_{\mathrm{clean}}(\mathrm dx)}{F_a(z)}.
\end{aligned}
\label{eq:app-density-ratio}
\end{equation}
Thus $F_a$ is the density ratio of the actual
observation distribution with respect to the reference distribution. The measure
$\pi_{a,z}$ is the posterior distribution of the clean image given
$Z_a=z$. Whenever a ratio involving $F_a$ appears below, it is understood
only at states for which $F_a>0$.

\begin{lemma}
\label{lem:app-posterior}
Under the joint law in \eqref{eq:app-joint-law}, once $Z_a$ is known,
the observations at lower information levels contain no additional
information about $X$. The observation process is Markov, and its
forward transition satisfies
\begin{equation}
\overrightarrow K_{a,b}(z,\mathrm dz')
=\overrightarrow K^0_{a,b}(z,\mathrm dz')
\frac{F_b(z')}{F_a(z)},
\qquad
\overrightarrow K^0_{a,b}F_b=F_a.
\label{eq:app-true-transform}
\end{equation}
For every test function for which the following expressions are defined,
the generator of the observation process is
\begin{equation}
\mathcal A_a[h]f
=\frac{\mathcal A^0(F_af)-f\mathcal A^0F_a}{F_a}.
\label{eq:app-transform-generator}
\end{equation}
\end{lemma}

\begin{proof}
By construction, all states below level $a$ are obtained from $Z_a$ through
degradation kernels that do not depend on $X$. Therefore, once $Z_a$ is
known, observing the lower-level states cannot change the posterior
distribution of $X$. For the same reason, the lower-level states are
conditionally independent of $X$ and of all higher-level states given
$Z_a$, so past and future are conditionally independent given the present,
which is the Markov property. Bayes' rule then gives
\eqref{eq:app-density-ratio}.

Proposition~\ref{prop:app-degradation} below shows that the actual
observation process and the reference process have the same transitions
when run backward from a higher information level to a lower one. Hence
their two-level joint laws can be written as
\[
p_b(\mathrm dz')\overleftarrow K_{b,a}(z',\mathrm dz)
\]
and
\[
p^0_b(\mathrm dz')\overleftarrow K_{b,a}(z',\mathrm dz),
\]
respectively. Dividing the first joint law by the second and then
conditioning on the state at level $a$ gives the first identity in
\eqref{eq:app-true-transform}. Integrating that identity over $z'$ gives
$\overrightarrow K^0_{a,b}F_b=F_a$.

Finally,
\[
\overrightarrow K_{a,b}f = \frac{\overrightarrow K^0_{a,b}(F_bf)}{F_a}.
\]
Differentiating with respect to $b$ at $b=a$ and using
$\partial_aF_a=-\mathcal A^0F_a$ gives
\eqref{eq:app-transform-generator}.
\end{proof}

\subsubsection{Gaussian observations}
\label{app:model-gaussian}

For $X\in\mathbb R^d$, define
\[
Z_a=aX+V_a, \qquad Q_a(x,\cdot)=\mathcal N(ax,a\Id),
\]
where $V$ is Brownian motion independent of $X$.

We take Brownian motion as the reference process. Thus
\[
p^0_a=\mathcal N(0,a\Id), \qquad \mathcal A^0=\tfrac12\Delta.
\]
The density ratio is
\[
F_a(z)=\mathbb E\exp\!\left( z\cdot X-\frac a2\|X\|^2 \right).
\]
Differentiating its logarithm gives
\[
m_a(z) = \nabla\log F_a(z) = \mathbb E[X\mid Z_a=z].
\]
Therefore the posterior quantity required by the Gaussian dynamics is
the posterior mean of the clean image. The generator is
\[
\mathcal A_a[m]f = m_a\cdot\nabla f+\tfrac12\Delta f.
\]
For every $a>0$, the factor
$\exp(-a\|X\|^2/2)$ controls the large values of $X$. As a result,
the displayed expectation and the derivatives needed below are locally
finite in $(a,z)$ for any probability distribution
$p_{\mathrm{clean}}$.

The posterior mean also determines the Gaussian score. Since
\[
\nabla\log p_a(z)=m_a(z)-z/a,
\]
the normalized observation $Y_{1/a}=Z_a/a$ satisfies Tweedie's identity~\citep{robbins1954Empirical, efron2011tweedie}
\[
\nabla_y\log p_{Y_{1/a}}(y) = a\bigl(\mathbb E[X\mid Y_{1/a}=y]-y\bigr).
\]

\subsubsection{Poisson observations}
\label{app:model-poisson}

For $X\in[0,\infty)^d$, let $N_{\lambda,i}$ be conditionally
independent Poisson processes with rates $X_{i}$. Conditional on
$X=x$,
\[
Q_\lambda(x,\{n\}) = \prod_i e^{-\lambda x_i} \frac{(\lambda x_i)^{n_i}}{n_i!}.
\]
We take independent unit-rate Poisson processes as the reference process.
Its distribution is
\[
p^0_\lambda(n) = e^{-d\lambda}\frac{\lambda^{\|n\|_1}}{n!}.
\]
The corresponding density ratio is
\[
F_\lambda(n) = e^{d\lambda} \mathbb E\left[ \prod_i X_{i}^{n_i}e^{-\lambda X_{i}} \right].
\]
The reference generator is
\[
\mathcal A^0f(n) = \sum_i[f(n+e_i)-f(n)].
\]
Using the ratio of $F_\lambda$ at two neighboring count states gives \begin{equation}
\begin{aligned}
h_{i,\lambda}(n)
&= \frac{F_\lambda(n+e_i)}{F_\lambda(n)}
= \mathbb E[X_{i}\mid N_\lambda=n] = \frac{n_i+1}{\lambda}
\frac{p_\lambda(n+e_i)}{p_\lambda(n)},\\
\mathcal A_\lambda[h]f(n)
&= \sum_i h_{i,\lambda}(n)[f(n+e_i)-f(n)].
\end{aligned}
\label{eq:app-poisson-rates}
\end{equation}
Thus the jump rate in coordinate $i$ is the posterior mean of the clean
Poisson intensity in that coordinate.

At any positive exposure, the factor $e^{-\lambda X_{i}}$ makes these
posterior rates finite on every state with positive probability, without
requiring moment assumptions on the clean-image distribution. This is the
count-thickening process used by~\citet{chen2023learning}.

\subsubsection{Binomial observations}
\label{app:model-binomial}

Here the clean image is integer valued:
$X\in\mathbb N^d$. Think of
$X_{i}$ as the number of particles in coordinate $i$. Give each particle
an independent arrival time uniformly distributed on $[0,1]$, and let
$N_{p,i}$ be the number that have arrived by time $p$. Then
\[
N_p\mid X \sim \operatorname{Binomial}(X,p)
\]
coordinatewise.

Write
\[
(x)_n=\frac{x!}{(x-n)!}
\]
for $x\geq n$, and set $(x)_n=0$ for $x<n$.
Relative to the unit-rate Poisson reference,
\[
F_p(n) = e^{dp}\mathbb E\left[ \prod_i (X_{i})_{n_i}(1-p)^{X_{i}-n_i} \right], \qquad 0<p<1.
\]
There are two closely related posterior quantities. The first is the
posterior mean of the total clean count,
\[
D_{i,p}(n) = \mathbb E[X_{i}\mid N_p=n].
\]
The second is the rate at which new particles arrive. Since $n_i$ particles have already arrived, this rate is \begin{equation}
u_{i,p}(n)
= \frac{D_{i,p}(n)-n_i}{1-p}
= \frac{F_p(n+e_i)}{F_p(n)}.
\label{eq:app-binomial-rate}
\end{equation}
The denominator $1-p$ appears because each particle that has not yet
arrived has instantaneous arrival rate $1/(1-p)$. Averaging the
image-dependent rate
\[
\frac{X_{i}-n_i}{1-p}
\]
over the posterior distribution of $X$ gives $u_{i,p}(n)$.

The generator is therefore
\[
\mathcal A_p[D]f(n) = \sum_i u_{i,p}(n)[f(n+e_i)-f(n)].
\]
The network predicts $D_i$ and consistency is imposed on $D_i$, whereas the
generator and the uniqueness argument use $u_i$. For an arbitrary family of birth rates $u_{i,p}$, define
\[
D_{i,p}(n)=n_i+(1-p)u_{i,p}(n)
\]
and
\[
\mathcal A_p[u]f = \sum_i u_{i,p}[f(\cdot+e_i)-f].
\]
Since $\mathcal A_p[u]n_i=u_{i,p}$, \begin{equation}
(\partial_p+\mathcal A_p[u])D_{i,p}
= (1-p)(\partial_p+\mathcal A_p[u])u_{i,p}.
\label{eq:app-binomial-consistency}
\end{equation}
Hence, away from $p=1$, imposing consistency on the posterior counts
is equivalent to imposing it on the birth rates. The reverse, thinning
direction of this process is the one used by Blackout Diffusion~\citep{santos2023blackout}.

\subsubsection{Gamma observations}
\label{app:model-gamma}

For $X\in(0,\infty)^d$, let $S^0_{L,i}$ be independent rate-one Gamma
processes, independent of $X$, and define
\[
S_{L,i}=X_{i}S^0_{L,i}.
\]
Here a rate-one Gamma process is an increasing process whose increment
from $L$ to $L'>L$ has distribution
$\operatorname{Gamma}(L'-L,1)$. Conditional on $X=x$,
\[
Q_L(x,\mathrm ds) = \prod_i \frac{s_i^{L-1}e^{-s_i/x_i}} {\Gamma(L)x_i^L} \,\mathrm ds_i, \qquad s\in(0,\infty)^d.
\]
The reference process has independent rate-one Gamma coordinates. Its generator is \begin{equation}
\mathcal A^0f(s)
= \sum_i\int_0^\infty
[f(s+we_i)-f(s)]e^{-w}\frac{\mathrm dw}{w}.
\label{eq:app-gamma-reference-generator}
\end{equation}
It is convenient to write the density ratio as
\begin{equation}
\begin{aligned}
F_L(s) =e^{\sum_i s_i}U_L(s),\qquad
U_L(s)=
\mathbb E\left[
\prod_i X_{i}^{-L}e^{-s_i/X_{i}}
\right].
\end{aligned}
\label{eq:app-gamma-potential}
\end{equation}
For every $L>0$ and every state with $s_i>0$, $U_L(s)$ is finite and
positive for any distribution supported on finite positive image values.
Indeed, for fixed $s_i>0$, the function
$x^{-L}e^{-s_i/x}$ is bounded on $(0,\infty)$.

Unlike the Gaussian and Poisson cases, the Gamma process requires more than a posterior mean. A jump of size $w$ in coordinate $i$ occurs at a rate determined by
\begin{equation}
\begin{aligned}
H_{i,L}(s,w)
= \mathbb E[e^{-w/X_{i}}\mid S_L=s] = e^{-w}\frac{F_L(s+we_i)}{F_L(s)}.
\end{aligned}
\label{eq:app-gamma-prediction}
\end{equation}
The generator is
\begin{equation}
\mathcal A_L[H]f(s)
= \sum_i\int_0^\infty
[f(s+we_i)-f(s)]H_{i,L}(s,w)
\frac{\mathrm dw}{w}.
\label{eq:app-gamma-generator}
\end{equation}
These same posterior Laplace transforms appear in the Gamma L\'evy channel
of~\citet{jiao2017relations}.
The normalized measurement has
\[
\mathbb E[S_L/L\mid X]=X, \qquad \operatorname{Var}(S_{L,i}/L\mid X) = \frac{X_{i}^2}{L}.
\]
Thus $S_L/L$ becomes increasingly concentrated around $X$ as $L$
increases.

The Gamma process has infinitely many small jumps on every interval, but
their total size is finite. At every positive state,
\begin{equation}
\begin{aligned}
\int_0^1H_{i,L}(s,w)\,\mathrm dw&\leq1,\\
\int_1^\infty H_{i,L}(s,w)\frac{\mathrm dw}{w}
&\leq
1+\mathbb E[\log(1+X_{i})\mid S_L=s]
<\infty.
\end{aligned}
\label{eq:app-gamma-integrability}
\end{equation}
For the second inequality we use
\[
\int_1^\infty e^{-w/x}\frac{\mathrm dw}{w} \leq1+\log(1+x),
\]
together with the fact that
$x^{-L}e^{-s_i/x}\log(1+x)$ is bounded when $L>0$ and $s_i>0$.

The formulas above are written for states in $(0,\infty)^d$, but the
completely empty state $S_0=0$ lies on the boundary of this state space. For every
$L>0$, however, $S_L$ lies in $(0,\infty)^d$ almost surely.

Conditional on $X=x$, the true process is
\[
S_{L,i}=x_iS^0_{L,i}.
\]
The unconditional process is obtained by averaging over
$x\sim p_{\mathrm{clean}}$. A theorem that identifies the process only from
some positive level $a_*>0$ says nothing by itself about how a
learned process enters the interior from $L=0$.

\begin{assumption}\label{ass:app-gamma-entrance}
When generation is started from $S_0=0$, the learned Gamma process is
initialized with the correct family of small-positive-level distributions:
for every $\varepsilon>0$, its state at level $\varepsilon$ has law
$p_\varepsilon = p_{\mathrm{clean}}Q_\varepsilon$,
and these initializations are consistent as $\varepsilon\downarrow0$.
\end{assumption}

Appendix~\ref{app:cifar-counts} describes the initialization used in the
experiments and how it departs from this assumption.
For the count models, zero belongs to the state space, so their
initialization can be handled directly in Appendix~\ref{app:zero-start}.

\subsubsection{Negative-binomial observations}
\label{app:model-nb}

Let $q=q(X)\in[0,1)^d$
be a known encoding of the clean image. Conditional on $X$, coordinate
$i$ has jumps of size $k\geq1$ at rate
$\frac{q_i^k}{k}.$
The total jump rate in that coordinate is
\[
\sum_{k\geq1}\frac{q_i^k}{k} = -\log(1-q_i).
\]
At shape $r$, the conditional measurement law is
\[
Q_r(x,\{n\}) = \prod_i \frac{\Gamma(n_i+r)}{n_i!\,\Gamma(r)} q_i(x)^{n_i}(1-q_i(x))^r.
\]
Choose a fixed $q_0\in(0,1)$ for the reference process. Its marginal is
\[
p^0_r(n) = \prod_i \frac{\Gamma(n_i+r)}{n_i!\,\Gamma(r)} q_0^{n_i}(1-q_0)^r,
\]
and the density ratio is
\[
F_r(n) = \frac{ \mathbb E[ \prod_iq_i(X)^{n_i}(1-q_i(X))^r ] }{ q_0^{\|n\|_1}(1-q_0)^{dr} }.
\]
The reference generator, posterior moments, and true generator are \begin{equation}
\begin{aligned}
\mathcal A^0 f(n)
&= \sum_{i=1}^d\sum_{k\geq1}
\frac{q_0^k}{k}[f(n+ke_i)-f(n)],\\
M_{i,k,r}(n)
&= \mathbb E[q_i(X)^k\mid N_r=n]
= q_0^k\frac{F_r(n+ke_i)}{F_r(n)},\\
\mathcal A_r[M]f(n)
&= \sum_{i=1}^d\sum_{k\geq1}
\frac{M_{i,k,r}(n)}{k}[f(n+ke_i)-f(n)].
\end{aligned}
\label{eq:app-nb-generator}
\end{equation}
Thus an exact sampler needs the full sequence of posterior moments
$M_{i,k,r}$, not only the first one. These moments also appear in the
negative-binomial L\'evy channel of~\citet{jiao2017relations}.

At every positive shape $r$, the total posterior jump rate is finite. To
see this, note that
\[
(1-q)^r|\log(1-q)|
\]
is bounded on $[0,1)$. Conditional on a fixed image, the process has
finitely many jumps on every finite interval, and this remains true after
averaging over the clean-image distribution.

\subsection{Degradation kernels and identification of the process}
\label{app:degradation}

We now describe what happens when an observation is moved in the direction
of \emph{less} information. These degradation rules depend only on the
measurement law, not on the clean image.

Moreover, degradation from a positive level loses no information about the
distribution at the higher level: if two distributions at level $b$ become the
same after degradation to a positive level $a$, they were already the same.
This is the injectivity property used later.

\begin{proposition}
\label{prop:app-degradation}
Let $0<a<b<a_{\max}$. The actual observation process and the reference
process have the same backward transition
$\overleftarrow K_{b,a}$. Coordinatewise,
\begin{equation}
\begin{array}{ll}
\text{Gaussian:}&
Z_a\mid Z_b=z\sim
\mathcal N((a/b)z,a(1-a/b)),\\[1mm]
\text{Poisson and binomial:}&
N_a\mid N_b=n\sim
\operatorname{Binomial}(n,a/b),\\[1mm]
\text{Gamma:}&
S_a\mid S_b=s\ \stackrel{\mathrm d}=s\Theta,
\quad
\Theta\sim\operatorname{Beta}(a,b-a),\\[1mm]
\text{Negative binomial:}&
N_a\mid N_b=n\sim
\operatorname{BetaBinomial}(n;a,b-a).
\end{array}
\label{eq:app-degradation-kernels}
\end{equation}
The coordinates are conditionally independent in each of these kernels.
Moreover, degradation is injective on probability distributions:
\[
\nu_1\overleftarrow K_{b,a} = \nu_2\overleftarrow K_{b,a} \quad\Longrightarrow\quad \nu_1=\nu_2.
\]
Equivalently, the map
$\nu\mapsto\nu\overleftarrow K_{b,a}$ is one-to-one.
\end{proposition}
\begin{proof}
We first derive the backward transitions.

For Gaussian observations,
\[
Z_a-\frac abZ_b = V_a-\frac abV_b.
\]
The quantity on the right is independent of $(X,V_b)$ and is Gaussian
with covariance
$a(1-a/b)\Id$. This gives the first line of
\eqref{eq:app-degradation-kernels}.

For Poisson and binomial observations, condition on the fact that $n$
particles have been observed by level $b$. Their arrival times are
independent and uniform on $[0,b]$. Each particle was therefore already
present by level $a$ with probability $a/b$, which gives binomial thinning.

For Gamma observations, write the level-$b$ state as the sum of the
independent increments from $0$ to $a$ and from $a$ to $b$. The ratio of
a Gamma variable with shape $a$ to the sum of independent Gamma variables
with shapes $a$ and $b-a$ has distribution
$\operatorname{Beta}(a,b-a)$ and is independent of the sum. This gives
$S_a=S_b\Theta$ conditionally on $S_b$.

For negative-binomial observations, the increment from $a$ to $b$ is
independent of the state at $a$ conditional on the image. Dividing the
product of the masses of independent
$\operatorname{NB}(a,q)$ and $\operatorname{NB}(b-a,q)$ variables by
the mass of their sum gives
\[
\mathbb P(N_a=j\mid N_b=n) = \binom nj \frac{\mathrm B(a+j,b-a+n-j)} {\mathrm B(a,b-a)}.
\]
None of these conditional distributions contains the clean image or its
parameters. They therefore apply equally to the actual observation process
and to the reference process. Applying the same conditioning at three
levels also gives the composition property of the degradation kernels.

We next prove injectivity. For each of the four kernels in
\eqref{eq:app-degradation-kernels}, we recover from the degraded distribution
a transform that uniquely determines the original distribution.

For Gaussian observations, let $\chi(\theta)$ denote the characteristic
function of the distribution at level $b$. Gaussian degradation evaluates
this characteristic function at $(a/b)\theta$ and multiplies it by a Gaussian
factor that is never zero. Hence the original characteristic function can
be recovered.

For Poisson and binomial observations, let $\phi$ be the probability
generating function of the law at level $b$, and set
$\alpha=a/b$. After binomial thinning, the generating function becomes
\[
\phi(1-\alpha+\alpha z).
\]
Knowing this function determines $\phi$ on an open interval inside
$(0,1)$. Since a probability generating function is analytic in its
domain, this determines $\phi$ everywhere.

For Gamma observations, take logarithms:
\[
\log S_a=\log S_b+\log\Theta.
\]
The characteristic function of $\log\Theta$ is
\[
\mathbb E[\Theta^{it}] = \frac{\Gamma(a+it)\Gamma(b)} {\Gamma(a)\Gamma(b+it)}.
\]
The Gamma function has no zeros, so this factor is never zero. Therefore
the characteristic function, and hence the distribution, of $\log S_b$
can be recovered from that of $\log S_a$.

For negative-binomial observations, put $c=b-a$ and denote by
$\phi_b$ and $\phi_a$ the generating functions before and after
degradation. For $0<t<1$,
\[
\phi_a(1-t) = \frac{t^{1-b}\Gamma(c)}{\mathrm B(a,c)} \mathcal I^c\!\left[ v^{a-1}\phi_b(1-v) \right](t),
\]
where
\[
\mathcal I^cf(t) = \frac1{\Gamma(c)} \int_0^t(t-v)^{c-1}f(v)\,\mathrm dv.
\]
The fractional integral $\mathcal I^c$ is one-to-one on locally integrable
functions: choose an
integer $m>c$, apply $\mathcal I^{m-c}$, and then differentiate $m$
times. The function to which it is applied is locally integrable because
$a>0$ and
$|\phi_b(1-v)|\leq1$. Hence $\phi_b$ is uniquely recovered.

For multivariate observations, the kernels act independently on the
coordinates. For the Gaussian and Gamma kernels, the joint transform
factorizes, and the univariate argument applies to each factor. For
thinning, the joint generating function is analytic on the open unit
polydisc and known on a product of real intervals, which determines it by
the identity theorem in several variables. For the negative-binomial
kernel, the fractional-integral representation is applied in each
coordinate in turn with the others held fixed. None of these arguments
requires moment assumptions on the clean-image distribution.
\end{proof}
Injectivity fails at $a=0$. At level zero, the observation is
empty, so every possible higher-level distribution degrades to the same
state.
\begin{theorem}
\label{thm:app-backward-identification}
Fix
$0<a_*<b_*<a_{\max}$. 
Let $(\widehat Z_a)_{a\in[a_*,b_*]}$ be a Markov process whose law at
$a_*$ is the true marginal $p_{a_*}$. Suppose that, for every
$a_*\leq a<b\leq b_*$, its backward transition from $b$ to $a$ is the
known degradation kernel $\overleftarrow K_{b,a}$. Then
$(\widehat Z_a)$ has the same finite-dimensional distributions as the
true observation process.
\end{theorem}
\begin{proof}
Let $\widehat p_b$ denote the marginal law of $\widehat Z_b$.
Since the process has the prescribed backward transition,
\[
\widehat p_b\overleftarrow K_{b,a_*} = p_{a_*}.
\]
The true process satisfies the same relation:
\[
p_b\overleftarrow K_{b,a_*} = p_{a_*}.
\]
By the injectivity in Proposition~\ref{prop:app-degradation},
\[
\widehat p_b=p_b.
\]
Thus the two processes have the same marginal distribution at every level.
Now consider any finite collection of levels. Starting from the marginal at
the largest level and repeatedly applying the common backward transitions
determines the entire joint law. Since the candidate and true processes
have the same final marginal and the same backward transitions, their joint
laws agree at every finite collection of levels.
\end{proof}
\subsection{Regression targets}
\label{app:targets}
We now explain how the posterior predictions required by the dynamics can
be learned without observing $X$.
Let $Y_{\mathrm{in}}$ and $Y_{\mathrm{tar}}$ be two measurements of the same clean image,
independent conditional on $X$. Suppose their information levels are
$a_{\mathrm{in}}$ and $a_{\mathrm{tar}}$, respectively.
By degrading $Y_{\mathrm{in}}$, we can construct an input $Z_a$ at any level
$0<a\leq a_{\mathrm{in}}$.
Let $\xi$ specify which posterior quantity is being learned. The level
$a$ and the index $\xi$ are sampled independently of the clean image and of
the noise in the target view.
Suppose that a transformation of the second measurement,
\[
T=T_\xi(Y_{\mathrm{tar}}),
\]
has the property
\[
\mathbb E[T\mid X,a,Z_a,\xi] = \psi_\xi(X).
\]
In words, $T$ is noisy, but its conditional mean is the clean
quantity whose posterior expectation we need. Conditional independence and
the tower property then give
\[
\mathbb E[T\mid a,Z_a,\xi] = \mathbb E[\psi_\xi(X)\mid a,Z_a,\xi] = h_{\xi,a}(Z_a).
\]
Thus regression onto $T$ learns the required posterior prediction.
\begin{lemma}\label{lem:app-n2n}
Let $C\subseteq\mathbb R^k$ be convex, and let $\phi:C\to\mathbb R$ be
strictly convex and differentiable in the interior of $C$. Its Bregman
divergence is
\begin{equation}
D_\phi(t,u)=\phi(t)-\phi(u)-\nabla\phi(u)\cdot(t-u),
\qquad t\in C,\ u\in\operatorname{int}C.
\label{eq:app-bregman}
\end{equation}
Let $T$ take values in $C$, with $\mathbb E\|T\|<\infty$ and
$\mathbb E|\phi(T)|<\infty$, and suppose that
$h=h_{\xi,a}(Z_a)=\mathbb E[T\mid a,Z_a,\xi]$ lies in $\operatorname{int}C$. Then, for every
predictor $\widehat h=\widehat h_{\xi,a}(Z_a)$ with values in $\operatorname{int}C$,
\begin{equation}
\mathbb E D_\phi(T,\widehat h)
= \mathbb E D_\phi(T,h)
+ \mathbb E D_\phi(h,\widehat h),
\label{eq:app-n2n-decomposition}
\end{equation}
and the first term on the right is finite. Consequently, $h$ is the unique
minimizer of the expected loss, up to events of probability zero. The same
holds for the loss $D_\phi(T,\widehat h)+g(T)$ whenever
$\mathbb E|g(T)|<\infty$.
\end{lemma}
\begin{proof}
Condition on $(a,Z_a,\xi)$. Then $h$ and $\widehat h$ are fixed and
$\mathbb E[T-h\mid a,Z_a,\xi]=0$, so
\[
\mathbb E[D_\phi(T,h)\mid a,Z_a,\xi]=\mathbb E[\phi(T)\mid a,Z_a,\xi]-\phi(h)
\]
and
\[
\mathbb E[D_\phi(T,\widehat h)\mid a,Z_a,\xi]-\mathbb E[D_\phi(T,h)\mid a,Z_a,\xi]
=\phi(h)-\phi(\widehat h)-\nabla\phi(\widehat h)\cdot(h-\widehat h)
=D_\phi(h,\widehat h).
\]
All three conditional terms are nonnegative by convexity, so taking
expectations gives \eqref{eq:app-n2n-decomposition}. Jensen's inequality and
an affine lower bound for $\phi$ show that $\phi(h)$ is integrable, so
$\mathbb E D_\phi(T,h)=\mathbb E\phi(T)-\mathbb E\phi(h)$ is finite. Strict
convexity gives $D_\phi(h,\widehat h)=0$ only when $\widehat h=h$. Adding
$g(T)$ changes the expected loss of every predictor by the same finite
constant.
\end{proof}
The squared error is the case $\phi(t)=\|t\|^2$ on $C=\mathbb R^k$, for which
$D_\phi(t,u)=\|t-u\|^2$ and the conditions reduce to $\mathbb E\|T\|^2<\infty$.
The I-divergence
\[
D_{\mathrm I}(t,u)=t\log(t/u)-t+u,
\]
with the convention $0\log0=0$, is the case $\phi(t)=t\log t$ on
$C=[0,\infty)$. Its conditions reduce to $\mathbb E|T\log T|<\infty$ and
$h>0$. For targets in $[0,1]$, the binary cross-entropy
$-t\log u-(1-t)\log(1-u)$ equals $D_\phi(t,u)-\phi(t)$ for
$\phi(t)=t\log t+(1-t)\log(1-t)$ on $C=[0,1]$. Since $|\phi|\leq\log2$, it
requires only $0<h<1$. Multiplying the loss by a positive weight $c(a,Z_a,\xi)$ does
not change the minimizer, provided the weighted loss at $h$ is finite.
For Gaussian, Poisson, and binomial observations, unbiased targets for the
posterior mean are, respectively,
\[
\frac{Y_{\mathrm{tar}}}{a_{\mathrm{tar}}}, \qquad \frac{Y_{\mathrm{tar}}}{\lambda_{\mathrm{tar}}}, \qquad \frac{Y_{\mathrm{tar}}}{p_{\mathrm{tar}}}.
\]
Their coordinatewise conditional variances are
\[
\frac1{a_{\mathrm{tar}}}, \qquad \frac{X_{i}}{\lambda_{\mathrm{tar}}}, \qquad \frac{X_{i}(1-p_{\mathrm{tar}})}{p_{\mathrm{tar}}}.
\]
To ensure that the total expected regression loss is finite, these
quantities must also be integrable over the distribution of images and over
the information levels sampled during training.
\subsubsection{Gaussian regression and double Tweedie}
\label{app:gaussian-n2n}
In the Gaussian case, the noisy target can also be constructed from a
single noisy image.
Let $t_{\mathrm{in}}=1/a_{\mathrm{in}}$. The normalized measurement
\[
Y_{t_{\mathrm{in}}}=Y_{\mathrm{in}}/a_{\mathrm{in}}=X+W_{t_{\mathrm{in}}}
\]
is the state at time $t_{\mathrm{in}}$ of the noising process $Y_t=X+W_t$
of Section~\ref{sec:observation-processes}. For a larger noise variance
$t>t_{\mathrm{in}}$, form
\[
Y_t = Y_{t_{\mathrm{in}}}+\sqrt{t-t_{\mathrm{in}}}\,\varepsilon,
\]
where $\varepsilon$ is a standard Gaussian vector independent of $X$ and
$Y_{t_{\mathrm{in}}}$, so that the pair has the joint law of the noising
process at these two times. Set
\[
\alpha=\frac{t_{\mathrm{in}}}{t}.
\]
A direct Gaussian-conditioning calculation gives
\[
\mathbb E[Y_{t_{\mathrm{in}}}\mid X,Y_t] = (1-\alpha)X+\alpha Y_t.
\]
Therefore
\[
T = \frac{Y_{t_{\mathrm{in}}}-\alpha Y_t}{1-\alpha}
\]
satisfies
\[
\mathbb E[T\mid X,Y_t]=X.
\]
It is therefore a valid noisy target for the clean posterior mean.
For a predictor $f$, \begin{equation}
\mathbb E\bigl\|(1-\alpha)f(Y_t)+\alpha Y_t-Y_{t_{\mathrm{in}}}\bigr\|^2
= (1-\alpha)^2\mathbb E\|f(Y_t)-T\|^2.
\label{eq:app-double-tweedie}
\end{equation}
This is the noisy-target form of the double-Tweedie objective of
\citet{pmlr-v235-daras24a}. At a fixed $t$, the two losses differ only by
the positive factor $(1-\alpha)^2$ and therefore have the same
minimizer of the expected loss. If different noise levels are mixed in one objective, this
factor changes how much each level is weighted, but it does not change the
optimal predictor at any individual level when that predictor can be
represented by the model.
In the information coordinate of our observation process, the network
input is
\[
Z_a=aY_{1/a}, \qquad a=1/t<a_{\mathrm{in}}.
\]
The formula above requires $t>t_{\mathrm{in}}$. At $t=t_{\mathrm{in}}$,
one either takes a limit or uses an independent target measurement.
\subsubsection{Gamma targets and their loss}
\label{app:gamma-targets}
For the Gamma process, the required posterior quantity is
\[
H_{i,L}(s,w) = \mathbb E[e^{-w/X_{i}}\mid S_L=s].
\]
We therefore need a noisy target whose conditional expectation is
$e^{-w/X_{i}}$.
For a target measurement whose coordinates have shape $L_{\mathrm{tar}}>0$, and for a
queried jump size $w\geq0$, define
\[
T_{w,L_{\mathrm{tar}}}(v) = \begin{cases} (1-w/v)^{L_{\mathrm{tar}}-1},&v>w,\\ 0,&0<v\leq w. \end{cases}
\]
Notice that $T_{0,L_{\mathrm{tar}}}=1$, while for $L_{\mathrm{tar}}=1$,
\[
T_{w,1}(v)=\mathbf1_{\{v>w\}}.
\]
Below, $Y_{\mathrm{tar}}$ denotes one coordinate of the target measurement.
\begin{proposition}
\label{prop:app-gamma-target}
If $Y_{\mathrm{tar}}\mid X_i=x \sim \operatorname{Gamma}(L_{\mathrm{tar}},1/x)$,
then
\[
\mathbb E[T_{w,L_{\mathrm{tar}}}(Y_{\mathrm{tar}})\mid x] = e^{-w/x}
\]
for every $L_{\mathrm{tar}}>0$.
For a fixed $w>0$, the conditional second moment of $T_{w,L_{\mathrm{tar}}}(Y_{\mathrm{tar}})$ is finite if and only if $L_{\mathrm{tar}}>1/2$. When $L_{\mathrm{tar}}>1/2$, writing $\vartheta=w/x$, \begin{equation}
\operatorname{Var}(T_{w,L_{\mathrm{tar}}}(Y_{\mathrm{tar}})\mid x)
= \frac{e^{-\vartheta}}{\Gamma(L_{\mathrm{tar}})}
\int_0^\infty
u^{2L_{\mathrm{tar}}-2}(u+\vartheta)^{1-L_{\mathrm{tar}}}e^{-u}\,\mathrm du
-e^{-2\vartheta}.
\label{eq:app-gamma-variance}
\end{equation}
For every fixed $L_{\mathrm{tar}}>1/2$, the second moment is bounded uniformly over all
$x>0$ and $w\geq0$.
\end{proposition}
\begin{proof}
Multiplying $T_{w,L_{\mathrm{tar}}}$ by the Gamma density simplifies the power of $v$:
\[
\mathbb E[T_{w,L_{\mathrm{tar}}}(Y_{\mathrm{tar}})\mid x] = \frac1{\Gamma(L_{\mathrm{tar}})x^{L_{\mathrm{tar}}}} \int_w^\infty (v-w)^{L_{\mathrm{tar}}-1}e^{-v/x}\,\mathrm dv.
\]
With the change of variable $u=v-w$, this becomes
$e^{-w/x}$,
which proves unbiasedness.
For the second moment, set
$v=w+xu.
$
After this substitution, the integrand in
\eqref{eq:app-gamma-variance} behaves like $u^{2L_{\mathrm{tar}}-2}$ as
$u\downarrow0$ whenever $w>0$. This is integrable at zero exactly when
\[
2L_{\mathrm{tar}}-2>-1,
\]
or equivalently $L_{\mathrm{tar}}>1/2$.
If $L_{\mathrm{tar}}\geq1$, then $0\leq T_{w,L_{\mathrm{tar}}}\leq1$, so the second moment is
bounded. If $1/2<L_{\mathrm{tar}}<1$, use
$(u+\vartheta)^{1-L_{\mathrm{tar}}} \leq u^{1-L_{\mathrm{tar}}}+\vartheta^{1-L_{\mathrm{tar}}}$
to obtain
\[
\mathbb E[T_{w,L_{\mathrm{tar}}}(Y_{\mathrm{tar}})^2\mid x] \leq e^{-\vartheta}\left[ 1+ \frac{\Gamma(2L_{\mathrm{tar}}-1)}{\Gamma(L_{\mathrm{tar}})} \vartheta^{1-L_{\mathrm{tar}}} \right].
\]
The right-hand side is uniformly bounded in $\vartheta\geq0$.
At $w=0$, the target is identically equal to one.
\end{proof}
When $L_{\mathrm{tar}}\leq1/2$, squared-error regression is not appropriate because the
target has infinite second moment for $w>0$. In this case, we instead replace
the squared loss with the I-divergence $D_{\mathrm I}$ defined after
Lemma~\ref{lem:app-n2n}, whose risk is finite for every target shape.
\begin{proposition}
\label{prop:app-gamma-idiv}
Let
\[
T=T_{w,L_{\mathrm{tar}}}(Y_{\mathrm{tar}}), \qquad m=\mathbb E[T\mid a,Z_a,\xi]\in(0,1].
\]
For every fixed $L_{\mathrm{tar}}>0$, $\mathbb E[|T\log T|\mid X_i=x]$ is bounded uniformly
over $x>0$ and $w\geq0$. Consequently, the optimal expected I-divergence
$\mathbb E D_{\mathrm I}(T,m)$ is finite, and $m$ is its unique minimizer
among finite positive predictors $\widehat h_{\xi,a}(Z_a)$, up to events of
probability zero.
\end{proposition}
\begin{proof}
When $L_{\mathrm{tar}}\geq1$, $T\in[0,1]$, so
\[
|T\log T|\leq1/e.
\]
When $0<L_{\mathrm{tar}}<1$, we have $T\geq1$ on $\{Y_{\mathrm{tar}}>w\}$ and $T=0$ otherwise, so
$T\log T\geq0$. The same change of variable as above gives, with
$G\sim\operatorname{Gamma}(L_{\mathrm{tar}},1)$,
\[
\mathbb E[T\log T\mid x] = (1-L_{\mathrm{tar}})e^{-\vartheta} \mathbb E\log(1+\vartheta/G).
\]
Using
\[
\log(1+\vartheta/G) \leq \log(1+\vartheta)+\log(1+1/G),
\]
we obtain the bound
\[
(1-L_{\mathrm{tar}})(e^{-1}+C_{L_{\mathrm{tar}}}), \qquad C_{L_{\mathrm{tar}}}=\mathbb E\log(1+1/G)<\infty.
\]
Hence $\mathbb E[|T\log T|\mid x]$ is bounded uniformly over $x$ and $w$.
Since $T\geq0$, $\mathbb E[T\mid x]=e^{-w/x}\leq1$, and $m>0$ lies in the
interior of $[0,\infty)$, the remaining claims follow from
Lemma~\ref{lem:app-n2n} with $\phi(t)=t\log t$.
\end{proof}
\subsubsection{Negative-binomial moment targets}
\label{app:nb-targets}
For the negative-binomial process, the exact jump rates require the
posterior moments
\[
\mathbb E[q_i(X)^k\mid N_r].
\]
Regression onto the following target learns these moments directly.
For target shape $r_{\mathrm{tar}}>0$ and integer $k\geq1$, define \begin{equation}
T_{k,r_{\mathrm{tar}}}(v)
= \mathbf1_{\{v\geq k\}}
\prod_{j=0}^{k-1}
\frac{v-j}{v+r_{\mathrm{tar}}-1-j}.
\label{eq:app-nb-target}
\end{equation}
If
$Y_{\mathrm{tar}}\mid q\sim\operatorname{NB}(r_{\mathrm{tar}},q)$ for one coordinate $Y_{\mathrm{tar}}$ of the target,
then
$\mathbb E[T_{k,r_{\mathrm{tar}}}(Y_{\mathrm{tar}})\mid q] = q^k.
$ Moreover,
\[
0\leq T_{k,r_{\mathrm{tar}}}(v)\leq C_{k,r_{\mathrm{tar}}} := \max\left\{ 1, \frac{k!\,\Gamma(r_{\mathrm{tar}})}{\Gamma(k+r_{\mathrm{tar}})} \right\}.
\]
To verify unbiasedness, multiply the target by the negative-binomial mass.
The factors in the product cancel, leaving
\[
\sum_{v\geq k} \frac{\Gamma(v-k+r_{\mathrm{tar}})} {(v-k)!\,\Gamma(r_{\mathrm{tar}})} q^v(1-q)^{r_{\mathrm{tar}}} = q^k.
\]
When $r_{\mathrm{tar}}\geq1$, every factor in
\eqref{eq:app-nb-target} is at most one. When $r_{\mathrm{tar}}<1$, the product
decreases as $v$ increases beyond $k$, so its maximum occurs at $v=k$.
This proves the stated bound and, in particular, shows that squared loss
is finite for every fixed $k$ and $r_{\mathrm{tar}}>0$.
If the moment index itself is sampled from an unbounded set, say
$K\in\{1,2,\ldots\}$, a sufficient condition for finite total regression
risk is
\[
\mathbb E[C_{K,r_{\mathrm{tar}}}^2]<\infty.
\]
For $k=1$, regression onto this target learns
\[
\mathbb E[q_i(X)\mid N_r].
\]
This should not be confused with the normalized count $Y_{\mathrm{tar}}/r_{\mathrm{tar}}$, whose
conditional mean is
\[
\frac{q}{1-q}.
\]
If the encoding is linear,
$q_i(X)=cX_{i}$, then $T_{1,r_{\mathrm{tar}}}/c$ is an unbiased target for the clean
image coordinate. Applying a nonlinear transformation to $Y_{\mathrm{tar}}/r_{\mathrm{tar}}$ does not
in general produce an unbiased image target.
\subsection{Splitting a single measurement}
\label{app:splitting}
The previous discussion assumed that two conditionally independent noisy
views $Y_{\mathrm{in}}$ and $Y_{\mathrm{tar}}$ are available. For Gaussian, Poisson, Gamma, and
negative-binomial observations, two such views can also be created by
randomly splitting a single acquisition. These are instances of Gaussian
recorruption and data thinning \citep{pang2021recorrupted, neufeld2024data, leiner2025data}.
\begin{proposition}
\label{prop:app-splitting}
Let
\[
a_{\mathrm{in}},a_{\mathrm{tar}}>0, \qquad a_{\mathrm{in}}+a_{\mathrm{tar}}=a_{\mathrm{acq}},
\]
and suppose
\[
Y_{\mathrm{acq}}\mid X \sim Q_{a_{\mathrm{acq}}}(X,\cdot).
\]
The following coordinatewise constructions produce two observations
$Y_{\mathrm{in}}$ and $Y_{\mathrm{tar}}$ that are independent conditional on $X$, with
\[
Y_{\mathrm{in}}\sim Q_{a_{\mathrm{in}}}(X,\cdot), \qquad Y_{\mathrm{tar}}\sim Q_{a_{\mathrm{tar}}}(X,\cdot):
\]
\begin{align*}
\text{Gaussian:}\quad
Y_{\mathrm{in}}&=
\frac{a_{\mathrm{in}}}{a_{\mathrm{acq}}}Y_{\mathrm{acq}}
+
\sqrt{\frac{a_{\mathrm{in}}a_{\mathrm{tar}}}
{a_{\mathrm{acq}}}}\,\varepsilon,
&
Y_{\mathrm{tar}}&=Y_{\mathrm{acq}}-Y_{\mathrm{in}},\\
\text{Poisson:}\quad
Y_{\mathrm{in}}\mid Y_{\mathrm{acq}}
&\sim
\operatorname{Binomial}
(Y_{\mathrm{acq}},
a_{\mathrm{in}}/a_{\mathrm{acq}}),
&
Y_{\mathrm{tar}}&=Y_{\mathrm{acq}}-Y_{\mathrm{in}},\\
\text{Gamma:}\quad
Y_{\mathrm{in}}&=\Theta Y_{\mathrm{acq}},
\quad
\Theta\sim\operatorname{Beta}
(a_{\mathrm{in}},a_{\mathrm{tar}}),
&
Y_{\mathrm{tar}}&=(1-\Theta)Y_{\mathrm{acq}},\\
\text{Negative binomial:}\quad
Y_{\mathrm{in}}\mid Y_{\mathrm{acq}}
&\sim
\operatorname{BetaBinomial}
(Y_{\mathrm{acq}};
a_{\mathrm{in}},a_{\mathrm{tar}}),
&
Y_{\mathrm{tar}}&=Y_{\mathrm{acq}}-Y_{\mathrm{in}}.
\end{align*}
Here $\varepsilon\sim\mathcal N(0,\Id)$. All auxiliary random variables
are independent across coordinates and independent of $(X,Y_{\mathrm{acq}})$.
\end{proposition}
\begin{proof}
For Gaussian observations, a direct calculation shows that, conditional
on $X$, $Y_{\mathrm{in}}$ and $Y_{\mathrm{tar}}$ have the required means and variances and have
zero cross-covariance. Since they are jointly Gaussian, zero
cross-covariance implies independence.
For the other three models, conditional on $X$ the observation at level
$a_{\mathrm{acq}}$ can be written as a sum of two independent observations
at levels $a_{\mathrm{in}}$ and $a_{\mathrm{tar}}$:
\[
Q_{a_{\mathrm{acq}}} = Q_{a_{\mathrm{in}}}*Q_{a_{\mathrm{tar}}}.
\]
The splitting rule displayed above is the conditional distribution
of the first summand given their sum, as derived in
Proposition~\ref{prop:app-degradation}. Reversing this conditioning
recovers the independent pair.
\end{proof}
The split level trades the noise of the input against the noise of the target. If a Poisson
acquisition at level $a_{\mathrm{acq}}$ is split so that the input has
level $a<a_{\mathrm{acq}}$, then the target has level
$a_{\mathrm{acq}}-a$. Its normalized variance is
\[
\frac{X}{a_{\mathrm{acq}}-a},
\]
which diverges as $a$ approaches the acquisition level. Thus inputs very
close to the original acquisition leave very little information for the
independent target view.
For binomial observations, an analogous split does not produce
conditionally independent views of the clean count. We therefore use
independent acquisitions for that model.
\subsection{Reachable states and clean limits}
\label{app:support-limits}
For the count models, some count vectors may have zero probability at every
level, depending on $p_{\mathrm{clean}}$. The next proposition identifies the
states with positive probability.
\begin{proposition}
\label{prop:app-reachable}
For Poisson, binomial, and negative-binomial observations, define
\[
\mathcal S = \{n\in\mathbb N^d:p_a(n)>0\}.
\]
This set does not depend on the positive information level
$a\in(0,a_{\max})$.
For binomial observations,
\[
n\in \mathcal S \quad\Longleftrightarrow\quad p_{\mathrm{clean}}\{X\geq n\}>0,
\]
where the inequality is coordinatewise.
For Poisson observations, $n\in \mathcal S$ exactly when the image distribution
assigns positive probability to images satisfying
$X_{i}>0$ for every coordinate with $n_i>0$.
For negative-binomial observations, the same statement holds with
$q_i(X)>0$ in place of $X_{i}>0$.
The set $\mathcal S$ contains the zero vector, and if $n\in \mathcal S$ and $0\leq m\leq n$
coordinatewise, then $m\in \mathcal S$.
The true density ratio $F_a$ is positive on $\mathcal S$ and zero outside $\mathcal S$.
If the reference process is killed when it attempts to leave $\mathcal S$, that is,
sent to an absorbing cemetery state on which all functions vanish, its
transform by $F$ and its backward transitions inside $\mathcal S$ are unchanged.
\end{proposition}
\begin{proof}
The characterization of $\mathcal S$ follows directly from the conditional
likelihoods. At every positive level, those likelihoods are positive for
exactly the states described in the proposition. Hence the support does not
depend on the information level.
Now let $n\in \mathcal S$ and $0\leq m\leq n$. For binomial observations, $X\geq n$
implies $X\geq m$. For Poisson and negative-binomial observations, every
coordinate with $m_i>0$ also has $n_i>0$, so the condition for $m$ is part of
the condition for $n$. In each case $m\in \mathcal S$. The zero state has positive
conditional probability for every admissible clean image.
The reference mass $p^0_a(n)$ is strictly positive on all of
$\mathbb N^d$. Therefore
\[
F_a(n)=\frac{p_a(n)}{p^0_a(n)}
\]
is positive exactly on $\mathcal S$.
The counts of a reference process never decrease, so every state $m$ on a
reference path satisfies $m\leq n$, where $n$ is the state at which the path
ends. Hence a reference path that ends in $\mathcal S$ never leaves $\mathcal S$. Killing the reference process when it attempts to leave $\mathcal S$
therefore does not change paths that remain in $\mathcal S$, or the transition
probabilities between states in $\mathcal S$. Formula~\eqref{eq:app-true-transform}
then gives the same transformed process on $\mathcal S$.
\end{proof}
We finally record the limit in which the observation becomes clean.
\begin{proposition}
\label{prop:app-clean-limits}
For every admissible clean-image distribution, the observation processes
satisfy, almost surely,
\begin{equation}
\begin{gathered}
\frac{Z_a}{a}\longrightarrow X,\qquad
\frac{N_\lambda}{\lambda}\longrightarrow X,\qquad
N_p\longrightarrow X,\\
\frac{S_L}{L}\longrightarrow X,\qquad
\frac{N_r}{r+N_r}\longrightarrow q(X).
\end{gathered}
\label{eq:app-clean-limits}
\end{equation}
The limits are taken as
$a,\lambda,L,r\to\infty$ and $p\uparrow1$. The last ratio is understood
coordinatewise.
\end{proposition}
\begin{proof}
For the Gaussian process,
\[
\frac{Z_a}{a} = X+\frac{V_a}{a},
\]
and Brownian motion satisfies $V_a/a\to0$ almost surely.
Conditional on a fixed clean image, the usual strong law for a Poisson
process gives
\[
N_\lambda/\lambda\to X.
\]
The same argument for the Gamma process gives
\[
S_L/L\to X.
\]
In the binomial construction, each coordinate contains only finitely many
particles. Every particle has an arrival time strictly below one almost
surely, so
\[
N_p\to X
\]
as $p\uparrow1$.
For the negative-binomial process, the expected count added per unit shape
in coordinate $i$ is
\[
\sum_{k\geq1}k\frac{q_i^k}{k} = \sum_{k\geq1}q_i^k = \frac{q_i}{1-q_i}.
\]
Therefore
\[
\frac{N_{r,i}}r \longrightarrow \frac{q_i}{1-q_i}.
\]
Applying the continuous map $u\mapsto u/(1+u)$ gives
\[
\frac{N_{r,i}}{r+N_{r,i}} \longrightarrow q_i.
\]
These statements hold with conditional probability one for every
admissible clean image. Averaging over $X$ therefore gives the claimed
almost-sure limits without requiring moments of $p_{\mathrm{clean}}$.
\end{proof}
For the negative-binomial model, the limiting quantity is the encoded image
$q(X)$ rather than $X$ itself. Recovering the clean image therefore
requires the known encoding $q$ to be invertible on the support of the data.
The experimental choice
\[
q(X)=0.9X
\]
has this property.
\section{Continuation from a common density ratio}
\label{app:common}
This appendix gives the first of two continuation arguments.
Fix a finite interval
\[
I=[a_*,b_*], \qquad 0<a_*<b_*<a_{\max}.
\]
Suppose that, instead of predicting each required posterior quantity
independently, all predictions are obtained from one common density ratio
$\widehat F_a$, a candidate for $\mathrm dp_a/\mathrm dp^0_a$ in
\eqref{eq:app-density-ratio}.
For Gaussian observations, the corresponding prediction is
\[
\nabla\log \widehat F_a.
\]
For the jump models, the required predictions are obtained from ratios such
as $\widehat F_a(z+v)/\widehat F_a(z)$, together with the fixed model-dependent factors given
in Appendix~\ref{app:generators}. We denote all these predictions by
$h^{\widehat F}$. We also define the associated generator through
\eqref{eq:app-transform-generator}, even when $\widehat F$ is only a candidate
and not yet known to be the true density ratio.
The infinitesimal form of consistency is \begin{equation}
\bigl(\partial_a+\mathcal A_a[\widehat h]\bigr)
\widehat h_{\xi,a}=0.
\label{eq:app-infinitesimal-consistency}
\end{equation}
This equation says that, along the process generated by the predictions,
the expected value of each posterior prediction does not change as the
information level increases.
Written directly in terms of $\widehat h$, this is a nonlinear equation.
For a common $\widehat F$, a scalar rescaling of $\widehat F$
satisfies a linear equation for the reference process. We then use that equation to show that the learned process has the
degradation kernels as its backward transitions.
\subsection{Consistency gives a linear equation}
\label{app:local-harmonicity}
Define
\[
\gamma_a(z) = \frac{ (\partial_a+\mathcal A^0)\widehat F_a(z) }{ \widehat F_a(z) }.
\]
If $\widehat F$ already satisfied the linear reference equation
\[
(\partial_a+\mathcal A^0)\widehat F_a=0,
\]
then $\gamma$ would be identically zero. Thus $\gamma$ measures how far a
candidate $\widehat F$ is from satisfying that equation.
\begin{lemma}
\label{lem:app-consistency-identity}
Let
\[
R_v(a,z) = \frac{\widehat F_a(z+v)}{\widehat F_a(z)}.
\]
Assume that the derivatives below exist and that the generator integrals below converge absolutely. Then \begin{equation}
\bigl(\partial_a+\mathcal A_a[h^{\widehat F}]\bigr)R_v(a,z)
= R_v(a,z)\bigl[\gamma_a(z+v)-\gamma_a(z)\bigr].
\label{eq:app-ratio-identity}
\end{equation}
For the Brownian reference, if
\[
m=\nabla\log \widehat F,
\]
then \begin{equation}
\partial_a m+(m\cdot\nabla)m+\tfrac12\Delta m
= \nabla\gamma.
\label{eq:app-gradient-identity}
\end{equation}
Consequently, if the predictions generated by $\widehat F$ satisfy the consistency
equation \eqref{eq:app-infinitesimal-consistency}, then for each fixed
information level $a$,
\[
\gamma_a(z)=c(a)
\]
is independent of the state $z$.
If
\[
c\in C((a_*,b_*))\cap L^1(I),
\]
define \begin{equation}
G_a(z)
= \exp\!\left(-\int_{a_*}^a c(t)\,\mathrm dt\right)\widehat F_a(z).
\label{eq:app-rescaling}
\end{equation}
Then $G$ produces the same posterior predictions as $\widehat F$, and
\[
(\partial_a+\mathcal A^0)G_a=0
\]
on $(a_*,b_*)$.
\end{lemma}
\begin{proof}
We first consider a jump-model ratio
\[
R_v(a,z)=\widehat F_a(z+v)/\widehat F_a(z).
\]
Multiply the left-hand side of
\eqref{eq:app-ratio-identity} by $\widehat F_a(z)$. Using
\[
\widehat F_a(z)R_v(a,z)=\widehat F_a(z+v)
\]
and the translation structure of the reference generator gives
\[
(\partial_a+\mathcal A^0)\widehat F_a(z+v) - R_v(a,z) (\partial_a+\mathcal A^0)\widehat F_a(z).
\]
Dividing by $\widehat F_a(z)$ yields
\eqref{eq:app-ratio-identity}.
For Gaussian observations, write
\[
m=\nabla\log \widehat F.
\]
A direct calculation gives
\[
\gamma = \partial_a\log \widehat F + \tfrac12\Delta\log \widehat F + \tfrac12\|\nabla\log \widehat F\|^2.
\]
Taking the spatial gradient of this identity gives
\eqref{eq:app-gradient-identity}.
We now impose consistency.
For Gaussian observations,
\eqref{eq:app-infinitesimal-consistency} and
\eqref{eq:app-gradient-identity} imply
\[
\nabla\gamma_a=0.
\]
Hence $\gamma_a$ is constant in the state.
For a jump model,
\eqref{eq:app-ratio-identity} shows that consistency forces
\[
\gamma_a(z+v)=\gamma_a(z)
\]
whenever $v$ is an allowed jump. In the count models, every reachable state
can be connected to zero by a sequence of unit coordinate steps. Therefore
$\gamma_a$ has the same value at every reachable state. For Gamma observations,
given any two positive states, one can find a third state lying
coordinatewise above both. Each original state can be connected to that
common state by positive coordinate increments. Thus $\gamma_a$ is again
constant.
We may therefore write
\[
\gamma_a(z)=c(a).
\]
Multiplying $\widehat F_a$ by a scalar depending only on $a$ does not change
$\nabla\log \widehat F_a$ and does not change any ratio
$\widehat F_a(z+v)/\widehat F_a(z)$. Hence $G$ in
\eqref{eq:app-rescaling} gives the same posterior predictions.
The derivative of the scalar factor cancels $c(a)$, leaving
\[
(\partial_a+\mathcal A^0)G_a=0.
\]
\end{proof}
For binomial observations, we apply this argument to the birth rates rather
than directly to the posterior counts. Equation
\eqref{eq:app-binomial-consistency} shows that the two consistency equations are equivalent for $p<1$.
For negative-binomial observations, the ratios corresponding to unit jumps
already connect the reachable count states and are therefore enough to show
that $\gamma_a$ is state independent.
For the count models, $\widehat F$ is extended by zero outside the reachable set
$\mathcal S$, and the reference process is killed when it attempts to leave $\mathcal S$.
Equation~\eqref{eq:app-ratio-identity} is used only on edges contained in
$\mathcal S$. This zero extension is consistent because the reference paths move
only upward.
Finally, the requirement
$c\in L^1(I)$ constrains the endpoints: continuity of
$\widehat F$ inside the interval does not by itself imply that the integral of
$c(a)$ remains finite near $a_*$ or $b_*$.
\subsection{From the differential equation to reference transitions}
\label{app:reference-harmonicity}
The previous subsection gives the differential equation
\[
(\partial_a+\mathcal A^0)G_a=0.
\]
The continuation argument needs the corresponding identity over finite intervals:
\begin{equation}
\overrightarrow K^0_{a,b}G_b
= G_a,
\qquad a_*\leq a\leq b\leq b_*.
\label{eq:app-harmonicity}
\end{equation}
In words, if the reference process starts from a state at level $a$, moves
to level $b$, and we evaluate $G_b$ at the resulting state, then the
conditional expectation of this value is $G_a$ at the starting state.
Equation~\eqref{eq:app-harmonicity} is sometimes called
space--time harmonicity. We prove it separately for the count, Brownian,
and Gamma reference processes because the regularity issues are different.
\subsubsection{Count references}
\begin{proposition}
\label{prop:app-count-harmonicity}
Let $\Pi$ be a nonnegative kernel on a countable state space $\mathcal S$ with
$\Pi(z,\mathcal S)\leq1,$
and let
\[
\mathcal A^0=\kappa(\Pi-\Id), \qquad 0<\kappa<\infty.
\]
Suppose
\[
G:I\times \mathcal S\to[0,\infty)
\]
is continuous in the information level at every state, is $C^1$ in the
level on $(a_*,b_*)$, and satisfies
\[
\Pi G_a(z)<\infty, \qquad (\partial_a+\mathcal A^0)G_a(z)=0.
\]
Then
$\overrightarrow K^0_{a,b}G_b=G_a,
$ where \begin{equation}
\overrightarrow K^0_{a,b}
= e^{-\kappa(b-a)}
\sum_{j=0}^\infty
\frac{[\kappa(b-a)]^j}{j!}\Pi^j.
\label{eq:app-uniformization}
\end{equation}
No uniform bound on $G$ over the state space is required.
\end{proposition}
\begin{proof}
Set
\[
T=b_*-a_*, \qquad g(t,z)=e^{\kappa t}G_{b_*-t}(z).
\]
The differential equation for $G$ becomes
\[
g'(t)=\kappa\Pi g(t).
\]
Since $g\geq0$ and $\Pi$ is nonnegative, $g'(t)\geq0$.
We want to prove the power-series representation \begin{equation}
g(t,\cdot)
= \sum_{j=0}^\infty
\frac{[\kappa(t-s)]^j}{j!}\Pi^j g(s,\cdot),
\qquad 0\leq s\leq t\leq T.
\label{eq:app-count-series}
\end{equation}
Once this is established, undoing the change of variables gives
\eqref{eq:app-harmonicity}.
Because $G$ is not assumed uniformly bounded over the countable state space,
we justify the Taylor expansion using positivity.
For a step $h>0$, define the $k$th forward difference
\[
d_k(t) = \sum_{j=0}^k (-1)^{k-j} \binom kj g(t+jh).
\]
Each finite sum satisfies
\[
d_k'=\kappa\Pi d_k.
\]
If $d_k\geq0$, then $d_k'\geq0$, so
\[
d_{k+1}(t) = d_k(t+h)-d_k(t) \geq0.
\]
Since $d_0=g\geq0$, induction gives nonnegative forward differences of
every order.
A classical result of \citet{boas1940functions} then implies that, in the
interior of the interval, $g$ is smooth and all of its derivatives are
nonnegative. Such functions admit their Taylor
representation~\citep[Chapter~IV]{widder1941laplace}.
Repeated differentiation of
$g'=\kappa\Pi g$ gives
\[
g^{(j)}=\kappa^j\Pi^jg.
\]
To justify passing derivatives through $\Pi$, note that the relevant
difference quotients are nonnegative and can be dominated by a fixed
difference quotient at a slightly larger time. Starting from
$\Pi g<\infty$, dominated convergence then proves inductively that every
required integral is finite and that the displayed derivative identity
holds.
Substituting these derivatives into the Taylor expansion gives
\eqref{eq:app-count-series} for interior times.
To include the left endpoint $s=0$, fix a later interior time and let
$s\downarrow0$. Monotonicity gives termwise convergence, while the series
is dominated by its value at a slightly later time, which is finite.
The right endpoint is handled similarly by continuity and monotone
convergence. This proves the finite-interval identity on all of $I$.
\end{proof}
For the Poisson and binomial references, $\kappa=d$; for the
negative-binomial reference, $\kappa=-d\log(1-q_0)$.
If the reference process is killed when it attempts to leave the
reachable set $\mathcal S$, some rows of $\Pi$ have total mass below one. This is
allowed by the proposition.
\subsubsection{The Brownian reference}
\begin{proposition}
\label{prop:app-brownian-harmonicity}
Let
\[
G:I\times\mathbb R^d\to[0,\infty)
\]
be continuous. Assume that, in the interior of $I$, it has one continuous
derivative in the information level and two continuous spatial derivatives,
and satisfies
\[
\partial_aG+\tfrac12\Delta G=0.
\]
Then
\[
\overrightarrow K^0_{a,b}G_b=G_a
\]
for the Brownian reference transitions.
\end{proposition}
\begin{proof}
Fix $b>a_*$ and reverse the level variable:
\[
u(t,z)=G_{b-t}(z).
\]
Then
\[
\partial_tu=\tfrac12\Delta u.
\]
Thus $u$ is a nonnegative solution of the heat equation, with
initial value
\[
u(0,z)=G_b(z).
\]
The representation theorem for nonnegative heat-equation solutions~\citep{widder1944positive}
says that such a solution is obtained by
convolving an initial measure with the Gaussian heat kernel. Because
$G_{b-t}$ converges locally uniformly to $G_b$ as $t\downarrow0$, that
initial measure is
\[
G_b(z)\,\mathrm dz.
\]
Therefore
\[
G_a = \overrightarrow K^0_{a,b}G_b
\]
for every $a_*<a\leq b$.
It remains only to include $a=a_*$. Fatou's lemma gives
\[
\overrightarrow K^0_{a_*,b}G_b \leq G_{a_*} <\infty.
\]
For Gaussian kernels with variances $s$ and $s_0$, where $s_0/2\leq s\leq s_0$, the
variance-$s$ density is bounded by a fixed multiple of the
variance-$s_0$ density. Taking
\[
s=b-a, \qquad s_0=b-a_*
\]
therefore supplies an integrable dominating function. Dominated convergence
allows $a\downarrow a_*$ and extends the equality to the endpoint.
\end{proof}
\subsubsection{The Gamma reference}
The Gamma case is more technical because the reference process has
infinitely many small jumps. The additional assumption below says that,
after removing the factor $e^{\sum_i s_i}$, the candidate function
is a Laplace transform of a positive measure. The true Gamma density ratio
has this form.
\begin{proposition}
\label{prop:app-gamma-harmonicity}
Let $G:I\times(0,\infty)^d\to(0,\infty)$
be continuous in the information level at every state and $C^1$ in the
level in the interior. Suppose that each coordinate integral in
$\mathcal A^0G_a$ converges absolutely and
\[
(\partial_a+\mathcal A^0)G_a=0
\]
on $(a_*,b_*)$.
Assume also that, for every level $a$, there is a positive measure $\rho_a$ on $[0,\infty)^d$ such that \begin{equation}
u_a(s):=e^{-\sum_i s_i}G_a(s)
= \int_{[0,\infty)^d} e^{-s\cdot y}\rho_a(\mathrm dy).
\label{eq:app-gamma-laplace}
\end{equation}
Then
\[
\overrightarrow K^0_{a,b}G_b=G_a
\]
for the rate-one Gamma reference process.
\end{proposition}
\begin{proof}
The proof has three steps.
First, we translate the differential equation for $G$ into an evolution
equation for the measures $\rho_a$. Because the Laplace transform uniquely
determines a positive measure, it is enough to determine how
$\rho_a$ changes with $a$.
Frullani's identity gives
\[
\int_0^\infty (e^{-wy_i}-e^{-w}) \frac{\mathrm dw}{w} = -\log y_i.
\]
For $y_i>1$, the positive logarithmic part is integrable because, for
every fixed positive $s$,
\[
(\log y_i)^+e^{-s\cdot y} \leq C_s e^{-(s/2)\cdot y}.
\]
On each of the two sets $\{y_i>1\}$ and $\{y_i<1\}$, the integrand in
Frullani's identity has one sign. Tonelli's theorem can therefore be
applied separately to its positive and negative parts. The assumed
absolute convergence of the Gamma generator then also implies integrability
of the negative logarithmic part.
In particular, at every interior level the measure $\rho_a$ has no mass
on a coordinate face $\{y_i=0\}$. Define
\[
J(y)=\sum_i\log y_i.
\]
The differential equation for $G$ becomes \begin{equation}
\partial_a u_a(s)
= \int J(y)e^{-s\cdot y}\rho_a(\mathrm dy).
\label{eq:app-gamma-measure-derivative}
\end{equation}
Second, we turn this differential identity into an equation for the
measures themselves. Fix $s_0\in(0,\infty)^d$ and define the finite measure
\[
\sigma_a(\mathrm dy) = e^{-s_0\cdot y}\rho_a(\mathrm dy).
\]
Write
\[
P_a = \int \sum_i(\log y_i)^+ \,\mathrm d\sigma_a, \qquad N_a = \int \sum_i(-\log y_i)^+ \,\mathrm d\sigma_a.
\]
The preceding exponential bound, together with continuity of
$u_a(s_0/2)$ on the compact interval $I$, makes $P_a$ uniformly bounded.
Integrating
\[
\partial_au_a(s_0)=P_a-N_a
\]
and using continuity at the endpoints gives
\begin{equation}
\int_{a_*}^{b_*}
\int
\sum_i|\log y_i|
\,\sigma_a(\mathrm dy)\,\mathrm da
<\infty.
\label{eq:app-gamma-log-integrability}
\end{equation}
Now integrate \eqref{eq:app-gamma-measure-derivative} between levels $a$ and $b$, evaluate it at $s=s_0+q$ with $q\geq0$, and use Fubini's theorem. Uniqueness of Laplace transforms of finite measures gives \begin{equation}
\sigma_b
= \sigma_a+\int_a^b J\sigma_t\,\mathrm dt.
\label{eq:app-gamma-measure-equation}
\end{equation}
The logarithmic integrability bound
\eqref{eq:app-gamma-log-integrability} allows this identity, initially
obtained in the interior, to be extended to the endpoints.
On a compact box $[\varepsilon,R]^d$, the multiplier $J(y)$ is bounded,
so the measure equation is a linear ordinary differential equation.
Its unique solution is
\[
\sigma_b=e^{(b-a)J}\sigma_a.
\]
Letting $\varepsilon\downarrow0$ and $R\uparrow\infty$ yields \begin{equation}
\rho_b(\mathrm dy)
= \left(\prod_i y_i\right)^{b-a}\rho_a(\mathrm dy).
\label{eq:app-gamma-measure-evolution}
\end{equation}
Third, we substitute this relation into the Gamma reference transition.
The increment of the reference process between levels $a$ and $b$ has
independent
$\operatorname{Gamma}(b-a,1)$ coordinates. Its exponential transform at
the required argument contributes the factor
\[
\left(\prod_i y_i\right)^{-(b-a)}.
\]
Therefore, by Tonelli's theorem and
\eqref{eq:app-gamma-measure-evolution},
\[
\begin{aligned} \overrightarrow K^0_{a,b}G_b(s) &= e^{\sum_i s_i} \int e^{-s\cdot y} \left(\prod_i y_i\right)^{-(b-a)} \rho_b(\mathrm dy)\\ &= e^{\sum_i s_i} \int e^{-s\cdot y} \rho_a(\mathrm dy)\\ &= G_a(s). \end{aligned}
\]
This is the desired finite-interval identity.
\end{proof}
\subsection{Assumptions and continuation}
\label{app:common-conclusion}
We collect the hypotheses used above. Item~1, that all predictions come from one common density
ratio, defines the setting of this appendix. The remaining items are regularity conditions that justify the
differentiations, integrals, endpoint limits, and the passage from the
differential equation to the finite-interval relation.
\begin{assumption}
\label{ass:app-common}
On $I=[a_*,b_*]$, assume the following.
\begin{enumerate}
\item All required posterior predictions are obtained from one common density
ratio $\widehat F$ using the formulas of Appendix~\ref{app:generators}. It is
positive, except that for the count models it is set to zero outside the
reachable set $\mathcal S$.
\item In the interior of $I$, all derivatives and integrals used in
Lemma~\ref{lem:app-consistency-identity} exist. For jump processes, the
generator integrals involving $\widehat F$, the shifted values of $\widehat F$, and the
required ratios converge absolutely. For the Brownian process, the spatial
derivatives through order three and the mixed level--space derivatives
appearing in \eqref{eq:app-gradient-identity} are continuous.
\item The scalar function $c$ from
Lemma~\ref{lem:app-consistency-identity} satisfies
\[
c\in C((a_*,b_*))\cap L^1(I).
\]
The rescaled density ratio $G$ from \eqref{eq:app-rescaling} satisfies the
hypotheses of the appropriate finite-interval result:
Proposition~\ref{prop:app-count-harmonicity},
\ref{prop:app-brownian-harmonicity}, or
\ref{prop:app-gamma-harmonicity}.
\item For Gaussian and Gamma observations, $\widehat F$ and the corresponding
posterior predictions are continuous in the state at every level, including
$a_*$ and $b_*$.
\end{enumerate}
\end{assumption}
The true density ratio satisfies these requirements on every finite
positive-level interval. Indeed, its transition identity
\eqref{eq:app-true-transform} implies
\[
(\partial_a+\mathcal A^0)F_a=0,
\]
so the corresponding scalar $c$ is identically zero. The explicit
likelihood formulas above justify the required derivatives and integrals at
positive levels. In the Gamma case,
\[
e^{-\sum_i s_i}F_L(s)=U_L(s)
\]
is a joint Laplace transform, as required by
Proposition~\ref{prop:app-gamma-harmonicity}.
\begin{corollary}
\label{cor:app-common-continuation}
Consider any of the five measurement laws. Suppose that the candidate
predictions are generated from a common density ratio $\widehat F$, agree with the true
posterior predictions at level $a_*$, and satisfy
\eqref{eq:app-infinitesimal-consistency} on $(a_*,b_*)$.
Under Assumption~\ref{ass:app-common}, the candidate predictions equal the
true predictions at every level in $I=[a_*,b_*]$.
Moreover, the process generated by these predictions, when initialized with
the true marginal $p_{a_*}$, has the same law as the true observation
process on $I$.
\end{corollary}
\begin{proof}
By Lemma~\ref{lem:app-consistency-identity}, the consistency relation produces a
positive rescaled density ratio $G$ satisfying the linear reference equation.
The appropriate proposition from the previous subsection then gives
\[
\overrightarrow K^0_{a,b}G_b=G_a.
\]
At the starting level $a_*$, equality of the posterior predictions
determines $G_{a_*}$ up to one positive multiplicative constant. For
Gaussian observations, equality of
$\nabla\log G_{a_*}$ and the true logarithmic gradient determines the
function up to such a constant. For the jump models, equality of the
neighboring-state ratios determines $G_{a_*}$ up to such a constant by
following paths through the state space.
Choose the constant so that
\[
G_{a_*}p^0_{a_*}=p_{a_*}.
\]
Now define candidate forward transitions by \begin{equation}
\widehat{\overrightarrow K}_{a,b}(z,\mathrm dw)
= \overrightarrow K^0_{a,b}(z,\mathrm dw)
\frac{G_b(w)}{G_a(z)}.
\label{eq:app-candidate-transform}
\end{equation}
Because
$\overrightarrow K^0_{a,b}G_b=G_a$, these are probability kernels.
The same identity also gives their composition law.
If the process starts from $G_{a_*}p^0_{a_*}$, its marginal at level $b$
is $G_bp^0_b$. When the corresponding two-level joint law is conditioned
backward, the factors of $G$ cancel, so the backward transition is the known
degradation kernel $\overleftarrow K_{b,a}$.
For the count models with a killed reference process, the same calculation
is performed on the reachable set $\mathcal S$. Since paths move only upward, a path
cannot leave $\mathcal S$ and later return to it.
Theorem~\ref{thm:app-backward-identification} now applies. It gives
\[
G_bp^0_b=p_b
\]
for every $b\in I$. Hence $G_b$ is the true density ratio
$p^0_b$-almost everywhere. For count models, the reference distribution has
positive mass at every reachable state, so equality holds at every state in
$\mathcal S$. For Gaussian and Gamma observations, the reference distribution has
full support and the density ratios are continuous, so equality extends to
every state.
The posterior predictions obtained from $G$ therefore equal the true
ones. The resulting forward transitions are also the true transitions.
Thus the candidate process and the true observation process have the same
law. Since the candidate predictions equal the true ones, this is also the
process generated by the candidate predictions whenever the generator
determines its process, as it does for bounded count rates and for the
Gamma construction of Lemma~\ref{lem:app-gamma-process}.
\end{proof}
Applying the corollary on every finite interval identifies the process at
all higher information levels and therefore gives the clean limits in
Proposition~\ref{prop:app-clean-limits} when the process is initialized at
$p_{a_*}$.
This does not by itself explain how to start from the completely empty
observation. That separate issue is handled in
Appendices~\ref{app:zero-start} and \ref{app:sampler-endpoints}.
\section{Uniqueness for direct predictions}
\label{app:direct}
The common-density-ratio result is mathematically convenient, but our
experimental networks do not use this parameterization. They predict the
required posterior quantities directly.
We therefore need a second continuation argument. It compares two families of
predictions that agree at $a_*$ and satisfy the same consistency equation.
Under bounds on the predictions, the estimate in
Theorem~\ref{thm:app-direct-uniqueness} shows that their difference stays
zero, so the continuation is unique.
The assumptions needed for this argument depend on the observation model.
For the count models, they control the jump rates. For Gamma observations,
they control both small jumps and the tail of the jump-size distribution.
\subsection{A general uniqueness estimate}
\label{app:direct-uniqueness}
Let the outputs be indexed by $\xi$, and choose positive weights
$\mu_\xi$. For a family $u=(u_\xi)$, define
\[
\|u\|_\mu = \sup_{z,\xi} \frac{|u_\xi(z)|}{\mu_\xi}.
\]
All evolution equations in this section are understood in integrated form
at every state and every output index.
The following theorem gives the estimate we will apply to each
observation model.
\begin{theorem}
\label{thm:app-direct-uniqueness}
Let $u$ satisfy
\[
u_a = u_{a_*} + \int_{a_*}^a\mathcal V_t(u_t)\,\mathrm dt
\]
on $I=[a_*,b_*]$, and suppose $v$ satisfies the same equation.
Assume
\[
u_{a_*}=v_{a_*}.
\]
Let
\[
\delta(a)=\|u_a-v_a\|_\mu
\]
be measurable and bounded. Suppose
\[
\|\mathcal V_a(u_a)-\mathcal V_a(v_a)\|_\mu \leq \ell(a)\, \omega(\|u_a-v_a\|_\mu),
\]
where $\ell\geq0$ is integrable and $\omega$ is continuous,
nondecreasing, satisfies
\[
\omega(0)=0, \qquad \omega(r)>0\quad\text{for }r>0,
\]
and obeys
\[
\int_{0^+}\frac{\mathrm dr}{\omega(r)} = \infty.
\]
Then
$u_a=v_a $
for every $a\in I$.
\end{theorem}
\begin{proof}
Subtract the two integrated equations. After dividing by the output weights
and taking the supremum,
\[
\delta(a) \leq \int_{a_*}^a \ell(t)\omega(\delta(t))\,\mathrm dt.
\]
For $\varepsilon>0$, define
\[
w_\varepsilon(a) = \varepsilon + \int_{a_*}^a \ell(t)\omega(\delta(t))\,\mathrm dt.
\]
Then
\[
\delta(a)\leq w_\varepsilon(a)
\]
and, almost everywhere,
\[
w_\varepsilon'(a) \leq \ell(a)\omega(w_\varepsilon(a)).
\]
Dividing by $\omega(w_\varepsilon)$ and integrating gives
\[
\int_\varepsilon^{w_\varepsilon(a)} \frac{\mathrm dr}{\omega(r)} \leq \int_{a_*}^a\ell(t)\,\mathrm dt.
\]
If $\delta(a)$ were positive at some level, then
$w_\varepsilon(a)$ would remain bounded away from zero as
$\varepsilon\downarrow0$. The left-hand side would then diverge because
\[
\int_{0^+}\frac{\mathrm dr}{\omega(r)}=\infty,
\]
while the right-hand side is finite. This contradiction shows that
$\delta(a)=0$ for every $a$.
\end{proof}
The same estimate also gives a stability statement when the starting value
and the consistency equation are only approximately correct.
\begin{remark}
\label{rem:app-stability}
Suppose
\[
\partial_a\widehat u = \mathcal V_a(\widehat u)+\mathrm{res}_a, \qquad \partial_au = \mathcal V_a(u).
\]
Then the argument above begins from
\[
\delta(a) \leq \varepsilon + \int_{a_*}^a \ell(t)\omega(\delta(t))\,\mathrm dt,
\]
where
\[
\varepsilon = \|\widehat u_{a_*}-u_{a_*}\|_\mu + \int_{a_*}^{b_*} \|\mathrm{res}_t\|_\mu\,\mathrm dt.
\]
If $\omega(r)=r$, Gr\"onwall's inequality gives
\[
\sup_I\delta \leq \varepsilon \exp\!\left(\int_I\ell\right).
\]
For the Gamma modulus used below,
\[
\omega(r) = r\bigl(1+\log^+(1/r)\bigr), \qquad \log^+x=\max\{0,\log x\},
\]
and for constant $\ell=K$, one obtains
\[
\sup_I\delta \leq e\, \varepsilon^{\exp[-K(b_*-a_*)]}
\]
whenever the right-hand side is at most one.
These are uniform bounds over states and output indices. A small
\emph{average} training loss does not imply that the quantity
$\varepsilon$ above is small. Moreover, the constants grow with the image size. For the count models,
$\ell=4J$ with $J$ proportional to $d$ (Appendix~\ref{app:modulus-estimates}),
so the factor $\exp(4J(b_*-a_*))$ is exponential in the number of pixels and in
the length of the continuation. For Gamma, the constant $K$ of
Lemma~\ref{lem:app-gamma-modulus} is proportional to $d$, since its proof sums
over the $d$ coordinates. These estimates are therefore qualitative for
images, and they can deteriorate further as the continuation interval
approaches the clean limit.
\end{remark}
\subsection{Admissible predictions}
\label{app:direct-classes}
For each observation model we now state the bounds under which the preceding uniqueness
argument will be applied.
Fix
\[
I=[a_*,b_*].
\]
For the count models, let $\mathcal S$ be the reachable set from
Proposition~\ref{prop:app-reachable}. Any jump leaving $\mathcal S$ is assigned
rate zero. When convenient, we extend such rates by zero to states outside
$\mathcal S$.
\begin{assumption}
\label{ass:app-direct}
On $I$, assume the following model-specific conditions. The constants may
depend on $I$, and they are chosen large enough that both the true
predictions and the candidate predictions satisfy these bounds. Count-model predictions are
measurable in the information level for every fixed state and output index.
\begin{enumerate}
\item \emph{Poisson.}
Assume
\[
X\in[0,M]^d \qquad\text{almost surely},
\]
and
\[
0\leq \widehat h_{i,a}(n) \leq R_I
\]
on $I\times \mathcal S$, where
$M\leq R_I<\infty$. We write
\[
u_{i,a}=\widehat h_{i,a}.
\]
\item \emph{Binomial.}
Assume
\[
X\in\{0,\ldots,M\}^d \qquad\text{almost surely},
\]
and $b_*<1$. Suppose
\[
n_i \leq \widehat D_{i,a}(n) \leq R_I,
\]
where $M\leq R_I<\infty$. Define the corresponding birth rate
\[
u_{i,a}(n) = \frac{\widehat D_{i,a}(n)-n_i}{1-a}.
\]
In this setting the reachable set $\mathcal S$ is finite.
\item \emph{Negative binomial.}
Assume
\[
q(X)\in[0,\bar q]^d \qquad\text{almost surely},
\]
with $\bar q<1$. Choose
\[
\bar q\leq\beta_I<1, \qquad \beta_I>0,
\]
and suppose
\[
0\leq u_{i,k,a}(n) := \widehat M_{i,k,a}(n) \leq \beta_I^k
\]
for every $k\geq1$.
\item \emph{Gamma.}
Assume
\[
X\in[x_{\min},x_{\max}]^d \qquad\text{almost surely},
\]
where
\[
0<x_{\min}\leq x_{\max}<\infty.
\]
Write
\[
u_{i,w,a}(s) = \widehat H_{i,a}(s,w).
\]
Assume the candidate is measurable in $a$, continuous in $(s,w)$, and
satisfies
\begin{equation}
\begin{aligned}
0\leq\widehat H_{i,a}(s,w)
&\leq
\min\{1,C_I(1+w)^{-\alpha_I}\},\\
|\widehat H_{i,a}(s,w)-\widehat H_{i,a}(s',w)|
&\leq
\operatorname{Lip}_I\|s-s'\|_1,
\end{aligned}
\label{eq:app-gamma-class}
\end{equation}
for finite constants
$C_I\geq1$, $\alpha_I>0$, and
$\operatorname{Lip}_I$.
\end{enumerate}
\end{assumption}
With these definitions, the consistency equations for all four jump models take the same form: \begin{equation}
\partial_a u_{\xi,a}(z)
= \mathcal V(u_a)_\xi(z)
:= -\int u_{\eta,a}(z)
\bigl[u_{\xi,a}(z+v_\eta)-u_{\xi,a}(z)\bigr]
\,\Lambda(\mathrm d\eta).
\label{eq:app-direct-equation}
\end{equation}
The notation in this formula specializes as follows.
For Poisson and binomial observations,
\[
\eta=i, \qquad v_i=e_i,
\]
and the integral reduces to a sum over coordinates.
For negative-binomial observations,
\[
\eta=(i,k), \qquad v_{i,k}=ke_i, \qquad \Lambda\{(i,k)\}=\frac1k.
\]
For Gamma observations,
\[
\eta=(i,w), \qquad v_{i,w}=we_i,
\]
and $\Lambda$ is counting measure over $i$ multiplied by
$\mathrm dw/w$.
For binomial observations, equation
\eqref{eq:app-binomial-consistency} converts consistency of the
posterior counts into the common birth-rate equation above.
\subsection{Bounds for the consistency equation}
\label{app:modulus-estimates}
We now verify the estimate required by
Theorem~\ref{thm:app-direct-uniqueness}.
For the count models, choose output weights $\mu_\xi>0$ so that every
candidate in the class satisfies $\|u\|_\mu\leq1$.

Define
\[
J = \int\mu_\eta\,\Lambda(\mathrm d\eta).
\]
When $J<\infty$, compare the two products in
\eqref{eq:app-direct-equation} by adding and subtracting the mixed term
with one factor from $u$ and one from $v$. This gives
\begin{equation}
\|\mathcal V(u)-\mathcal V(v)\|_\mu
\leq 4J\|u-v\|_\mu,
\qquad
\|u\|_\mu,\|v\|_\mu\leq1.
\label{eq:app-count-lipschitz}
\end{equation}
One factor of $2J$ comes from changing the jump rate, and another
$2J$ comes from changing the increment of the predicted quantity.
The useful choices of weights are
\[
\begin{array}{c|c|c} \text{Model}&\mu&J\\ \hline \text{Poisson}&R_I&dR_I\\ \text{Binomial}&R_I/(1-b_*)&dR_I/(1-b_*)\\ \text{Negative binomial}& \mu_{i,k}=\beta_I^k& -d\log(1-\beta_I). \end{array}
\]
If necessary, $R_I$ can be enlarged so that it is strictly positive.
The same bounds imply absolute convergence of the generator sums.
Thus, for all three count models,
Theorem~\ref{thm:app-direct-uniqueness} applies with
\[
\omega(r)=r, \qquad \ell=4J.
\]
For Gamma observations, this argument cannot be used directly because
\[
\int_0^1\frac{\mathrm dw}{w}=\infty.
\]
The Gamma process has infinitely many small jumps, so we must use the fact
that a small jump changes a Lipschitz prediction only by an amount
proportional to its size.
\begin{lemma}
\label{lem:app-gamma-modulus}
Suppose $u$ and $v$ take values in $[0,1]$, are Lipschitz in the state
with common constant $L_0$ for the norm $\|\cdot\|_1$, and satisfy the
common tail bound
\[
u,v\leq C(1+w)^{-\alpha},
\]
with $C\geq1$ and $\alpha>0$. Then
\begin{equation}
\|\mathcal V(u)-\mathcal V(v)\|_\infty
\leq K\delta\bigl(1+\log^+(1/\delta)\bigr),
\qquad
\delta=\|u-v\|_\infty,
\label{eq:app-gamma-modulus}
\end{equation}
where $K$ depends only on $d,C,\alpha,L_0$. The right-hand side is defined
to be zero when $\delta=0$.
\end{lemma}
\begin{proof}
Fix an output $(j,\zeta)$ and a state $s$. For each jump coordinate $i$,
the difference between the two generator integrals is at most $T_1+T_2$,
where
\begin{align*}
T_1
&=
\int_0^\infty
|(u-v)_i(s,w)|
|u_j(s+we_i,\zeta)-u_j(s,\zeta)|
\frac{\mathrm dw}{w},\\
T_2
&=
\int_0^\infty
v_i(s,w)
|(u-v)_j(s+we_i,\zeta)-(u-v)_j(s,\zeta)|
\frac{\mathrm dw}{w}.
\end{align*}
Since $u$ and $v$ take values in $[0,1]$, $\delta\leq1$. The case $\delta=0$ is
trivial, so assume $\delta>0$. We have
\[
|(u-v)_i(s,w)| \leq \min\{\delta,2Cw^{-\alpha}\},
\]
while Lipschitz continuity gives
\[
|u_j(s+we_i,\zeta)-u_j(s,\zeta)| \leq \min\{L_0w,1\}.
\]
Similarly,
\[
v_i(s,w) \leq \min\{1,Cw^{-\alpha}\},
\]
and
\[
|(u-v)_j(s+we_i,\zeta)-(u-v)_j(s,\zeta)| \leq \min\{2L_0w,2\delta\}.
\]
We now separate small and large jumps.
For $T_1$, split first at $w=1$, and in the tail split again where
\[
\delta=2Cw^{-\alpha}, \qquad w=(2C/\delta)^{1/\alpha}.
\]
This gives
\[
T_1 \leq L_0\delta + \frac\delta\alpha \left( \log\frac{2C}{\delta}+1 \right).
\]
For $T_2$, the small-jump contribution is split where
\[
L_0w=\delta
\]
when that point lies below one. The resulting bound is
\[
T_2 \leq 2\delta \left( 1+\log^+\frac{L_0}{\delta} \right) + \frac{2C\delta}{\alpha}.
\]
If $L_0=0$, the small-jump term vanishes.
Summing these estimates over $i=1,\ldots,d$ and absorbing fixed constants
into $K$ gives \eqref{eq:app-gamma-modulus}.
\end{proof}
For Gamma predictions we therefore use the unweighted supremum norm,
corresponding to $\mu\equiv1$. The function
\[
\omega(r) = r\bigl(1+\log^+(1/r)\bigr)
\]
satisfies
\[
\int_{0^+}\frac{\mathrm dr}{\omega(r)}=\infty.
\]
Hence it meets the condition of
Theorem~\ref{thm:app-direct-uniqueness}.
The same estimates also explain why the Gamma generator is well defined:
Lipschitz continuity controls the singular factor $\mathrm dw/w$ near
$w=0$, while the polynomial envelope controls the large-$w$ tail.
Joint continuity in $(s,w)$ allows the supremum over states and jump sizes
to be reduced to a countable dense subset, which ensures measurability as a
function of the information level.
\subsection{The true predictions and the generated process}
\label{app:direct-conclusion}
We next verify that the true posterior predictions satisfy the same
conditions and the same consistency equation.
For the Poisson model, bounded images imply
\[
0\leq h_i\leq M.
\]
For the binomial model,
\[
n_i\leq D_i\leq M.
\]
For the negative-binomial model,
\[
0\leq M_{i,k}\leq\bar q^k.
\]
Differentiating the likelihood formulas from
Appendix~\ref{app:generators} gives
\eqref{eq:app-direct-equation}. The image bounds justify the
differentiation and the infinite negative-binomial sum. Rates of jumps that
leave the reachable set vanish by
Proposition~\ref{prop:app-reachable}.
For Gamma observations, bounded image values give
\[
H_{i,a}(s,w) \leq e^{-w/x_{\max}},
\]
which in particular satisfies a polynomial tail bound of the form required
in \eqref{eq:app-gamma-class}.
Let
\[
\delta_X = x_{\min}^{-1}-x_{\max}^{-1}.
\]
Bayes' rule gives
\[
\partial_{s_i}H_{j,a}(s,\zeta) = -\operatorname{Cov}\!\left( X_{i}^{-1}, e^{-\zeta/X_{j}} \mid S_a=s \right).
\]
A covariance between random variables taking values in intervals of lengths
$c_1$ and $c_2$ has absolute value at most $c_1c_2/4$. Applying this bound here gives
\begin{equation}
|\partial_{s_i}H_{j,a}(s,\zeta)|
\leq
\frac{\delta_X}4
\min\{
1,
\delta_X\zeta e^{-\zeta/x_{\max}}
\}.
\label{eq:app-gamma-derivative}
\end{equation}
Thus the required dependence on the observed state is Lipschitz.
Differentiating with respect to the information level instead gives
\[
\partial_aH_{j,a}(s,\zeta) = -\operatorname{Cov}\!\left( \sum_i\log X_{i}, e^{-\zeta/X_{j}} \,\middle|\, S_a=s \right).
\]
Use
\[
\log x = \int_0^\infty (e^{-w/x}-e^{-w}) \frac{\mathrm dw}{w}
\]
and the posterior identity
\[
H_{i,a}(s,w) H_{j,a}(s+we_i,\zeta) = \mathbb E[ e^{-w/X_{i}} e^{-\zeta/X_{j}} \mid S_a=s ].
\]
Substitution gives the Gamma version of
\eqref{eq:app-direct-equation}. Since
$\int_0^\infty|e^{-w/x}-e^{-w}|\,\mathrm dw/w=|\log x|$ is bounded for
$x\in[x_{\min},x_{\max}]$, Fubini's theorem justifies interchanging the
integral and the conditional expectation.
For the count processes, bounded jump rates are enough to construct a
unique Markov process by uniformization. Gamma requires a separate argument
because there are infinitely many small jumps.
\begin{lemma}
\label{lem:app-gamma-process}
Let
\[
H_{i,a}(s,w)\in[0,1]
\]
be measurable on
$I\times(0,\infty)^d\times(0,\infty)$. Suppose there are envelopes
$\overline H_i(w)\in[0,1]$ and finite constants $M_H,L_H$ such that
\begin{equation}
H_{i,a}(s,w)
\leq
\overline H_i(w),
\qquad
\sum_i\int_0^\infty
\overline H_i(w)\mathrm dw
\leq M_H,
\label{eq:app-gamma-first-moment}
\end{equation}
and
\begin{equation}
\sum_i\int_0^\infty
|H_{i,a}(s,w)-H_{i,a}(s',w)|
\,\mathrm dw
\leq
L_H\|s-s'\|_1.
\label{eq:app-gamma-integrated-lipschitz}
\end{equation}
Then, for every initial distribution on $(0,\infty)^d$, these jump rates
define an increasing c\`adl\`ag Markov process with generator
\eqref{eq:app-gamma-generator}. Its paths have finite total variation on
$I$. Its law is unique among solutions of the Poisson-driven equation
below when all jump sizes $w>0$ are retained.
\end{lemma}
\begin{proof}
The construction is easiest if we first ignore extremely small jumps.
Let $\mathcal N_i$ be independent Poisson random measures with intensity
\[
\mathrm da\,\mathrm dw\,\mathrm du/w,
\]
independent of the initial state. For $\varepsilon>0$, retain only jumps of size at least $\varepsilon$ and define \begin{equation}
S_t^\varepsilon
= S_{a_*}
+ \sum_i e_i
\int_{(a_*,t]\times[\varepsilon,\infty)\times(0,\infty)}
w\mathbf1_{\{u\leq H_{i,a}(S_{a-}^\varepsilon,w)\}}
\,\mathcal N_i(\mathrm da,\mathrm dw,\mathrm du).
\label{eq:app-gamma-driving-equation}
\end{equation}
Only Poisson points with
$u\leq\overline H_i(w)$ can be accepted. For
$w\geq\varepsilon$, their total rate is at most
$M_H/\varepsilon$. Hence there are only finitely many candidate events on
a finite interval, and the cutoff process can be constructed by processing
these events in chronological order.
We next let $\varepsilon\downarrow0$. Construct all cutoff processes with
the same Poisson random measures. If
$0<\varepsilon'<\varepsilon$, the jumps omitted by the larger cutoff have
expected total size at most $d\varepsilon(t-a_*)$.

The integrated Lipschitz condition controls the additional difference
created because the two processes can accept different later jumps. Using
the compensation formula,
\[
\begin{aligned} \mathbb E \sup_{a_*\leq r\leq t} \|S_r^\varepsilon-S_r^{\varepsilon'}\|_1 \leq{}& d\varepsilon(t-a_*)\\ &+ L_H \int_{a_*}^t \mathbb E \sup_{a_*\leq r\leq a} \|S_r^\varepsilon-S_r^{\varepsilon'}\|_1 \,\mathrm da. \end{aligned}
\]
Gr\"onwall's inequality gives
\[
\mathbb E \sup_{a_*\leq r\leq t} \|S_r^\varepsilon-S_r^{\varepsilon'}\|_1 \leq d\varepsilon(t-a_*) e^{L_H(t-a_*)}.
\]
Thus the cutoff paths form a Cauchy family in expected uniform distance.
A subsequence converges uniformly almost surely to an increasing
c\`adl\`ag process. The same estimate allows passage to the limit in the
driving equation, so the limit retains all jump sizes $w>0$.
The limit is a measurable function of the initial condition and of the
driving Poisson measures. If another process is driven by the same measures,
the same estimate compares it with every cutoff approximation, and letting
$\varepsilon\downarrow0$ shows that the two solutions coincide. This gives
pathwise uniqueness for the construction and hence uniqueness in law.
Restarting the same construction from any intermediate level gives the
Markov property. Finally,
\eqref{eq:app-gamma-first-moment} implies that the expected total size of
all jumps added over $I$ is at most
\[
M_H(b_*-a_*),
\]
so the paths have finite variation.
This is a standard Poisson-random-measure construction; see, for example, \citet[Chapters~2 and~6]{applebaum2009levy}.
\end{proof}
\begin{corollary}
\label{cor:app-direct-continuation}
Suppose Assumption~\ref{ass:app-direct} holds on every interval
$[a_*,b_*]$ with $b_*<a_{\max}$. If a candidate prediction agrees with the
true prediction at $a_*$ and satisfies
\eqref{eq:app-direct-equation} in integrated form on every such interval,
then
\[
\widehat h_{\xi,a}=h_{\xi,a}
\]
for every
\[
a\in[a_*,a_{\max}).
\]
The jump process generated from these predictions and initialized with
$p_{a_*}$ has the same path law as the true observation process and
therefore has the clean limits in
Proposition~\ref{prop:app-clean-limits}. For Gamma observations, the
identified process is the one constructed in
Lemma~\ref{lem:app-gamma-process}.
\end{corollary}
\begin{proof}
On every finite interval, the true and candidate predictions belong to the
same admissible class and solve the same integrated consistency equation
with the same value at $a_*$. For Poisson, binomial, and negative-binomial
observations, the Lipschitz estimate
\eqref{eq:app-count-lipschitz} and
Theorem~\ref{thm:app-direct-uniqueness} give equality. For Gamma
observations, use Lemma~\ref{lem:app-gamma-modulus} in the same theorem.
Thus the candidate and true predictions agree on every finite interval.
For the count models, the candidate rates now equal the true rates.
Uniformization therefore gives the true process law.
For Gamma observations, the candidate kernel now equals the true kernel $H$. Its exponential
tail gives
\[
M_H=dx_{\max}.
\]
Integrating \eqref{eq:app-gamma-derivative} over jump size gives
\[
L_H = \frac{d\delta_X^2x_{\max}^2}{4}.
\]
The true Gamma observation process has jump compensator
\[
\sum_i H_{i,a}(S_{a-},w) \, \frac{\mathrm da\,\mathrm dw}{w}
\]
for jumps of size $we_i$. A standard representation theorem for point
processes with absolutely continuous compensators allows these jumps to be
generated, on an enlarged probability space, by the thinning construction
\eqref{eq:app-gamma-driving-equation}~\citep{grigelionis1971representation}.
Lemma~\ref{lem:app-gamma-process} then identifies its law with the process
constructed from the learned kernel.
Equality of process laws gives the clean limits.
\end{proof}
\subsection{Poisson and binomial rates with linear growth}
\label{app:linear-growth}
The bounded-rate assumptions above are sufficient but can be stronger than
necessary. In Poisson and binomial models, the birth rates can grow with
the current count. We therefore give a second argument that allows linear
growth.
The idea is to recover the common density ratio $\widehat F$ from the directly predicted
birth rates. For this to be possible, increasing coordinate $i$ and then
coordinate $j$ must give the same product of rates as increasing
$j$ first and then $i$. We first show that consistency preserves this
path-independence property.
Let $\mathcal S\subseteq\mathbb N^d$ be nonempty, and assume that $n\in \mathcal S$ and $0\leq m\leq n$ imply $m\in \mathcal S$. Let $u_{i,a}(n)\geq0$ denote a birth rate. Set it to zero if $n+e_i\notin \mathcal S$, and extend the rates by zero outside $\mathcal S$. The consistency equation is \begin{equation}
\partial_a u_{i,a}(n)
+ \sum_k u_{k,a}(n)
\bigl[u_{i,a}(n+e_k)-u_{i,a}(n)\bigr]
=0.
\label{eq:app-birth-consistency}
\end{equation}
For Poisson observations, $u_i$ is the predicted posterior intensity.
For binomial observations,
\[
u_i = \frac{\widehat D_i-n_i}{1-a}.
\]
\begin{lemma}
\label{lem:app-compatible}
Suppose $u$ is continuous in the information level at every state,
satisfies \eqref{eq:app-birth-consistency} in integrated form, and obeys
the linear-growth bound
\begin{equation}
0\leq u_{i,a}(n)
\leq
K(1+\|n\|_1)
\qquad
\text{on }I\times \mathcal S
\label{eq:app-linear-bound}
\end{equation}
for some finite $K$.
Define
\[
C_{ij}(n,a) = u_{i,a}(n)u_{j,a}(n+e_i) - u_{j,a}(n)u_{i,a}(n+e_j).
\]
The condition $C_{ij}=0$ says that the product of rates obtained by taking an
$i$-step followed by a $j$-step is the same as the product obtained in
the opposite order.
If
\[
C_{ij}(n,a_*)=0
\]
for every $i,j,n$, then
\[
C_{ij}(n,a)=0
\]
throughout $I$. In addition, every rate that is strictly positive at
$a_*$ remains strictly positive on $I$.
\end{lemma}
\begin{proof}
Because $\partial_a u_{i,a}(n)$, given by
\eqref{eq:app-birth-consistency}, is continuous in $a$, each rate is
$C^1$ in the interior of $I$.
Suppress the level index temporarily and write
\[
\beta(n)=\sum_ku_k(n).
\]
Differentiating the definition of $C_{ij}$ and using \eqref{eq:app-birth-consistency} gives \begin{equation}
\begin{split}
\partial_a C_{ij}(n)
={}&[\beta(n)-\beta(n+e_i+e_j)]C_{ij}(n)\\
&+\sum_k\bigl[u_i(n+e_j)C_{kj}(n)-u_j(n+e_i)C_{ki}(n)\bigr]\\
&-u_i(n)\sum_k C_{kj}(n+e_i)
+u_j(n)\sum_k C_{ki}(n+e_j).
\end{split}
\label{eq:app-compatibility-evolution}
\end{equation}
The identity can be verified by expanding the right-hand side. Because the
rates are extended by zero outside $\mathcal S$, the same expression remains valid
at the boundary.
For $N\geq0$, define
\[
M_N(a) = \max_{\|n\|_1\leq N,i,j} |C_{ij}(n,a)|.
\]
The linear-growth bound implies that the sum of the absolute coefficients
in \eqref{eq:app-compatibility-evolution} is at most
\[
10dK(N+1),
\]
and the terms involving $C$ occur only at states with size at most $N+1$.
Therefore, if $C(\cdot,a_0)=0$, then with $c=10dK$,
\[
M_N(a) \leq c(N+1) \int_{a_0}^aM_{N+1}(t)\,\mathrm dt.
\]
Iterating this estimate $m$ times over an interval of length $\delta$ gives
\[
M_N(a) \leq \frac{(c\delta)^m(N+m)!}{m!N!} \,2K^2(N+m+2)^2.
\]
Using
\[
\frac{(N+m)!}{m!N!}\leq2^{N+m}
\]
gives
\[
M_N(a) \leq 2^N(2c\delta)^m \,2K^2(N+m+2)^2.
\]
If $2c\delta<1$, the right-hand side tends to zero as $m\to\infty$.
Hence $C=0$ on every sufficiently short interval. Covering $I$ with
finitely many such intervals proves $C=0$ everywhere. If $K=0$, the
conclusion is immediate.
Once $C=0$, equation~\eqref{eq:app-birth-consistency} simplifies to
\[
\partial_au_{i,a}(n) = u_{i,a}(n) [ \beta_a(n)-\beta_a(n+e_i) ].
\]
This is a scalar linear equation for each fixed edge. Its
integrating-factor formula shows that a rate that starts strictly positive
cannot become zero at a later level.
\end{proof}
\begin{theorem}
\label{thm:app-linear-growth}
Consider the Poisson or binomial observation process with reachable state
space $\mathcal S$. Suppose candidate birth rates $u_{i,a}(n)$ are continuous in the information level, vanish on edges leaving $\mathcal S$,
satisfy \eqref{eq:app-birth-consistency} in integrated form, obey
\eqref{eq:app-linear-bound}, and agree with the true rates at $a_*$.
Then the candidate rates equal the true rates throughout $I$. The process
generated by these rates and initialized with $p_{a_*}$ has the true path
law.
\end{theorem}
\begin{proof}
At level $a_*$, the candidate rates equal the true birth rates, which are ratios of the true density ratio at neighboring states.
They are positive on every edge inside $\mathcal S$ and satisfy
\[
C_{ij}=0.
\]
Lemma~\ref{lem:app-compatible} shows that both properties remain true at
every later level.
We can therefore reconstruct a common density ratio $\widehat F$ from the rates.
Let
\[
c_* = \frac{p_{a_*}(0)}{p^0_{a_*}(0)}>0
\]
and define \begin{equation}
\widehat F_a(0)
= c_*\exp\!\left(\int_{a_*}^a[d-\beta_t(0)]\,\mathrm dt\right),
\qquad
\widehat F_a(n+e_i)=u_{i,a}(n)\widehat F_a(n).
\label{eq:app-reconstruct-potential}
\end{equation}
Every coordinatewise path from zero to a state $n\in \mathcal S$ stays inside $\mathcal S$.
Any two such paths can be transformed into each other by repeatedly
swapping adjacent coordinate steps. The identity $C_{ij}=0$ says
that each such swap leaves the product of rates unchanged. Hence the value
of $\widehat F_a(n)$ does not depend on the chosen path. The function $\widehat F$ is
therefore well defined and positive on $\mathcal S$. Set it to zero outside $\mathcal S$.
At level $a_*$, its value at zero and all of its neighboring-state ratios
agree with those of the true density ratio. Thus
\[
\widehat F_{a_*}p^0_{a_*}=p_{a_*}.
\]
For an edge inside $\mathcal S$,
\[
\partial_a\log u_{i,a}(n) = \beta_a(n)-\beta_a(n+e_i).
\]
Summing this identity along any path from zero to $n$ gives
\[
\partial_a\log \widehat F_a(n) = d-\beta_a(n).
\]
For the unit-rate Poisson reference killed on leaving $\mathcal S$,
\[
\mathcal A^0\widehat F_a(n) = \widehat F_a(n)[\beta_a(n)-d],
\]
and
\[
\Pi \widehat F_a(n) = d^{-1}\widehat F_a(n)\beta_a(n) <\infty.
\]
Hence
\[
(\partial_a+\mathcal A^0)\widehat F_a=0.
\]
Proposition~\ref{prop:app-count-harmonicity} converts this differential
equation into the finite-interval relation
\[
\overrightarrow K^0_{a,b}\widehat F_b=\widehat F_a.
\]
The transformed process defined by
\eqref{eq:app-candidate-transform} therefore has the known degradation kernels as its backward
transitions. Theorem~\ref{thm:app-backward-identification} gives
\[
\widehat F_ap^0_a=p_a.
\]
Since the reference distribution has positive mass at every reachable
state, the ratios of $\widehat F$ agree pointwise with the true ratios. Hence all
candidate birth rates are the true ones.
Finally, the linear-growth condition prevents explosion. The total jump
rate is bounded by
\[
dK(1+\|n\|_1),
\]
so the number of jumps is dominated by a standard linear birth process,
which is nonexplosive. The directly constructed pure-birth process
therefore has the same law as the transformed process above.
\end{proof}
The same result can be applied on every finite interval if the constant in
the linear-growth bound is allowed to depend locally on the interval. The
linear-growth bound is an assumption on the candidate class. We do not claim that it
holds for every possible clean-image distribution.
\section{From the training objective to continuation}
\label{app:training}
The continuation theorems begin from two properties:
\begin{enumerate}
\item the learned posterior predictions equal the true predictions at some
positive level $a_*$; and
\item the predictions satisfy the consistency relation at the higher levels under consideration.
\end{enumerate}
We now explain when these properties follow from exact minimization of the
expected training objective. This depends on what a minimal expected loss
implies on the states and levels sampled by the loss, and on whether the sampled
set covers enough states and levels to invoke the continuation theorems.
\subsection{Exact minimizers and training support}
\label{app:population-objective}
For a candidate predictor $\widehat h$, let $\widehat{\overrightarrow K}_{a,b}$ denote the transition of the process generated by that predictor.
For a starting state $z$, levels $a<b$, and prediction index $\xi$, define the consistency residual \begin{equation}
\mathcal R_\xi(a,b,z)
= \widehat h_{\xi,a}(z)
-\bigl(\widehat{\overrightarrow K}_{a,b}\widehat h_{\xi,b}\bigr)(z).
\label{eq:app-consistency-residual}
\end{equation}
The ideal consistency loss is
\[
\mathcal L_{\mathrm{cons}}(\widehat h) = \mathbb E \|\mathcal R_\xi(a,b,z)\|^2.
\]
The consistency sampling law of $(a,b,z,\xi)$ is allowed to depend on the
candidate model, because $z$ or the endpoint may come from paths
simulated with that model. We assume that the conditional expectations in
\eqref{eq:app-consistency-residual} are finite. On finite intervals, the
bounded prediction classes used below guarantee this automatically.
\begin{proposition}
\label{prop:app-population-objective}
Suppose that the candidate class contains the true posterior predictions
$h$, that the regression targets satisfy the integrability conditions of
Lemma~\ref{lem:app-n2n}, and that the process generated by $h$ has the
true transitions.
For each output, use a regression loss of the form in
Lemma~\ref{lem:app-n2n}, such as the squared error, the I-divergence or the
binary cross-entropy, and assume that the true predictions lie in the
interior of its domain.
Then every global minimizer of
\begin{equation}
\mathcal L_{\mathrm{N2N}}(\widehat h)
+
\lambda_{\mathrm{cons}}
\mathcal L_{\mathrm{cons}}(\widehat h),
\qquad
\lambda_{\mathrm{cons}}>0,
\label{eq:app-training-objective}
\end{equation}
has the following two properties.
First,
\[
\widehat h=h
\]
almost surely under the regression sampling law.
Second,
\[
\mathcal R_\xi(a,b,z)=0
\]
almost surely under the consistency sampling law generated by that candidate.
If the prediction error and consistency residual are continuous, then these
identities hold at every point in the support of the corresponding sampling
laws.
\end{proposition}
\begin{proof}
A candidate that takes a boundary value of the domain with positive
probability, such as $\widehat H=0$ under the binary cross-entropy, has
infinite expected loss there, because the true prediction is interior. Such a
candidate is not a minimizer, so we may assume that candidate predictions
also lie in the interior.
Lemma~\ref{lem:app-n2n}, applied to each output, gives
\[
\mathcal L_{\mathrm{N2N}}(\widehat h) - \mathcal L_{\mathrm{N2N}}(h) = \mathbb E D_\phi(h,\widehat h) \geq0,
\]
where $\phi$ may depend on the output index. Equality holds exactly when the
prediction is correct almost surely under the regression sampling law.
By the tower property, the true posterior predictions satisfy the consistency relation, so
\[
\mathcal L_{\mathrm{cons}}(h)=0.
\]
Hence the true model reaches the value
$\mathcal L_{\mathrm{N2N}}(h)$ in the combined objective. No candidate can
go below this value, because the excess regression loss and the consistency
loss are both nonnegative.
Therefore every global minimizer must simultaneously attain the minimum
regression loss and zero consistency loss. This remains true even when the
consistency sampling law depends on the candidate
itself.
Finally, if a continuous nonnegative function is zero almost surely under a
measure, it must be zero throughout the support of that measure.
\end{proof}
This is the statement about the expected training objective used in the main text.
For bounded count rates, the required existence and uniqueness of the true
process follow from uniformization. For Gamma observations under the
conditions below, they follow from
Lemma~\ref{lem:app-gamma-process} and
Corollary~\ref{cor:app-direct-continuation}.
For Gaussian observations, a standard sufficient condition is uniqueness
in law for
\[
\mathrm dZ_a = m_a(Z_a)\,\mathrm da+\mathrm d\overline V_a.
\]
For example, local Lipschitz continuity and at most linear growth of $m$
are sufficient.
\paragraph{Coverage at the starting level.}
The proposition gives equality only on the support of the regression
sampling law. To obtain exact agreement at the continuation level
$a_*$, that support must reach $a_*$.
One possibility is to place positive probability directly on level $a_*$,
with full conditional support over the relevant states and prediction
indices. Another is to sample levels approaching $a_*$ and assume
continuity of the predictions, so that equality extends to $a_*$ by taking
a limit. We refer to either situation as \emph{coverage at $a_*$}.
For Gamma observations, training must also cover every queried jump size
$w>0$, and continuity in $w$ is used to extend almost-sure equality to all
jump sizes.
For negative-binomial observations, every coordinate and every moment
index $k\geq1$ must be covered. The total regression risk must remain
finite, as discussed in Appendix~\ref{app:nb-targets}.
\begin{remark}
\label{rem:app-all-starts}
Suppose that
\[
\mathcal R_\xi(a,b,z)=0
\]
for every state and prediction index and for every sufficiently short
interval $a<b$ inside $I$. This follows, for example, if the consistency
sampling law has full support over such intervals and the residual is
continuous.
Assume also that the candidate predictions lie in the domain of the
generator, in the sense that
\[
\lim_{\varepsilon\downarrow0} \frac{ \widehat{\overrightarrow K}_{a,a+\varepsilon} \widehat h_{\xi,a+\varepsilon} - \widehat h_{\xi,a} }{ \varepsilon }(z) = \bigl( \partial_a+\mathcal A_a[\widehat h] \bigr) \widehat h_{\xi,a}(z).
\]
Dividing the consistency relation between $a$ and $a+\varepsilon$ by $\varepsilon$ and
letting $\varepsilon\downarrow0$ gives
\eqref{eq:app-infinitesimal-consistency}.
Chaining short intervals with the semigroup property
$\widehat{\overrightarrow K}_{a,c}\widehat{\overrightarrow K}_{c,b}=\widehat{\overrightarrow K}_{a,b}$
extends the consistency relation to every pair of levels in $I$. For the jump models, the
backward equation then gives the integrated form
\eqref{eq:app-integrated-backward} used in
Theorem~\ref{thm:app-direct-uniqueness}, as in the proofs of
Lemmas~\ref{lem:app-count-injective} and~\ref{lem:app-gamma-injective}.
This argument applies to all five models, but it requires enforcing the consistency relation between paths that start throughout the continuation interval. In the next two subsections we show that, for the jump models under stronger assumptions, it is enough to start every simulated path from one fixed positive level $a_*$.
\end{remark}
\subsection{One starting level for count models}
\label{app:count-training}
Assume now that every simulated path begins at the same level
$a_*$. The consistency loss then directly enforces
\[
\widehat h_{\xi,a_*} = \widehat{\overrightarrow K}_{a_*,b} \widehat h_{\xi,b},
\]
but it does not directly compare the predictions at two later levels
$a$ and $b$ with
$a_*<a<b$.
For bounded count rates, this missing relation follows because the learned
forward transition is one-to-one on bounded functions. Unlike
Proposition~\ref{prop:app-degradation}, which concerns the known
degradation kernel acting on probability distributions, this injectivity
concerns the learned forward transition acting on bounded test functions.
\begin{lemma}
\label{lem:app-count-injective}
Let $\mathcal S$ be countable and let measurable jump rates
$q_a(z,z')\geq0$ satisfy
\[
\sup_{a\in I,z\in \mathcal S} \sum_{z'\ne z} q_a(z,z') \leq q_{\max}<\infty.
\]
Let
$\widehat{\overrightarrow K}_{a,b}$ be the resulting transition operators
and $\mathcal A_a$ the corresponding generator.
Then
$\widehat{\overrightarrow K}_{a,b}$ is injective on
$\ell^\infty(\mathcal S)$.
Moreover, suppose that a uniformly bounded family $u_a$ satisfies
\begin{equation}
\widehat{\overrightarrow K}_{a_*,b}u_b
= u_{a_*},
\qquad b\in I.
\label{eq:app-one-start}
\end{equation}
Then the consistency relation holds between every pair of intermediate levels:
\begin{equation}
\widehat{\overrightarrow K}_{a,b}u_b=u_a,
\qquad
u_b(z)-u_a(z)
= -\int_a^b(\mathcal A_tu_t)(z)\,\mathrm dt.
\label{eq:app-integrated-backward}
\end{equation}
\end{lemma}
\begin{proof}
The uniform rate bound allows the process to be constructed by
uniformization. In particular,
\[
\|\mathcal A_af\|_\infty \leq 2q_{\max}\|f\|_\infty.
\]
Fix $b$ and a bounded function $f$, and set
\[
v_s = \widehat{\overrightarrow K}_{s,b}f.
\]
Conditioning on the first jump gives the backward equation
\[
v_s = f+\int_s^b\mathcal A_tv_t\,\mathrm dt.
\]
Suppose
\[
v_a = \widehat{\overrightarrow K}_{a,b}f = 0.
\]
Subtracting the backward equation at level $a$ gives
\[
v_s = -\int_a^s\mathcal A_tv_t\,\mathrm dt.
\]
Hence
\[
\|v_s\|_\infty \leq 2q_{\max} \int_a^s \|v_t\|_\infty\,\mathrm dt.
\]
Because $\mathcal S$ is countable, the supremum is measurable in $t$.
Gr\"onwall's inequality yields $v_s=0$ for every $s\in[a,b]$, and therefore
\[
f=v_b=0.
\]
This proves injectivity.
Now assume \eqref{eq:app-one-start}. By the semigroup property,
\[
\widehat{\overrightarrow K}_{a_*,a} \left( \widehat{\overrightarrow K}_{a,b}u_b-u_a \right) = 0.
\]
Injectivity of
$\widehat{\overrightarrow K}_{a_*,a}$
implies
\[
\widehat{\overrightarrow K}_{a,b}u_b=u_a.
\]
Thus the consistency relation holds between all intermediate levels.
Taking
\[
u_s = \widehat{\overrightarrow K}_{s,b_*}u_{b_*}
\]
and applying the backward equation gives the second identity in
\eqref{eq:app-integrated-backward}.
\end{proof}
The count-model classes in Assumption~\ref{ass:app-direct} satisfy the
required uniform rate bound. The corresponding $q_{\max}$ can be taken to
be the value $J$ computed in
Appendix~\ref{app:modulus-estimates}.
\begin{corollary}
\label{cor:app-count-training}
Consider Poisson, binomial, or negative-binomial observations and fix
$a_*>0$. Assume the following.
\begin{enumerate}
\item On every finite continuation interval, the candidate class satisfies
Assumption~\ref{ass:app-direct}, contains the true predictions, and is
continuous in the information level at every state and prediction index.
\item The noisy regression targets are unbiased and satisfy the
integrability conditions of Lemma~\ref{lem:app-n2n}, and the regression
sampling law has coverage at $a_*$ as
described in Appendix~\ref{app:population-objective}.
\item Every simulated path starts at $a_*$. The consistency sampling law
of the starting state, endpoint level, and prediction index has full support
on
\[
\mathcal S\times(a_*,a_{\max})\times\Xi.
\]
For negative-binomial observations, $\Xi$ contains every coordinate and
every moment index $k\geq1$.
\end{enumerate}
Then every global minimizer of
\eqref{eq:app-training-objective} equals the true posterior predictions on
the entire continuation range
\[
[a_*,a_{\max}).
\]
The process generated by that minimizer and initialized with
$p_{a_*}$ has the true path law on every finite interval.
\end{corollary}
\begin{proof}
Fix an arbitrary $b_*<a_{\max}$. Proposition~\ref{prop:app-population-objective} gives two facts:
the candidate equals the truth at $a_*$, and its consistency residual is
zero almost surely under the consistency sampling law.
We next extend the zero residual from almost every sampled endpoint to
every endpoint in $[a_*,b_*]$. Bounded rates imply that the transition law
depends continuously on the endpoint in total variation. The predictions
are pointwise continuous in the information level. Therefore the residual
is continuous in the endpoint. Since the consistency sampling law has full support,
\[
\widehat{\overrightarrow K}_{a_*,b} \widehat h_{\xi,b} = \widehat h_{\xi,a_*}
\]
holds for every $b\leq b_*$, every reachable state, and every prediction
index.
Apply Lemma~\ref{lem:app-count-injective} separately to each output.
It gives consistency between every pair of intermediate levels and hence
the integrated version of
\eqref{eq:app-direct-equation}.
For the binomial model, the lemma first gives the integrated equation for
the posterior-count prediction. Equation
\eqref{eq:app-binomial-consistency} converts it into the corresponding
equation for the birth rates.
Corollary~\ref{cor:app-direct-continuation} now identifies the candidate
predictions with the true ones on $[a_*,b_*]$ and identifies the process
law. Since $b_*$ was arbitrary, the conclusion holds throughout
$[a_*,a_{\max})$.
\end{proof}
\subsection{One starting level for Gamma observations}
\label{app:gamma-training}
We extend this argument to Gamma observations. The Gamma process has infinitely many small jumps, so its generator
is not bounded on the entire space of bounded functions.
We instead work with bounded Lipschitz functions on
$(0,\infty)^d$. Denote this class by $\mathrm{BL}$, and measure
Lipschitz constants using the norm $\|\cdot\|_1$. For such a function, a
jump of size $w$ changes its value by at most a constant times $w$, which
cancels the singular factor $1/w$ in the Gamma jump measure near zero.
\begin{lemma}
\label{lem:app-gamma-injective}
Suppose the Gamma jump kernels satisfy the assumptions of
Lemma~\ref{lem:app-gamma-process} with constants $M_H$ and $L_H$, and let
$I=[a_*,b_*]$. Then, for $a\leq b$ in $I$, the following hold.
\begin{enumerate}
\item The transitions preserve bounded Lipschitz functions, and
\[
\|\widehat{\overrightarrow K}_{a,b}g\|_\infty \leq \|g\|_\infty,
\]
while
\[
\operatorname{Lip} ( \widehat{\overrightarrow K}_{a,b}g ) \leq e^{L_H(b-a)} \operatorname{Lip}(g).
\]
\item If $g\in\mathrm{BL}$ and
\[
v_s = \widehat{\overrightarrow K}_{s,b}g,
\]
then
\[
v_s = g+\int_s^b\mathcal A_tv_t\,\mathrm dt,
\]
and
\[
\|\mathcal A_tv_t\|_\infty \leq d\operatorname{Lip}(v_t) + 2M_H\|v_t\|_\infty.
\]
\item
$\widehat{\overrightarrow K}_{a,b}$ is injective on
$\mathrm{BL}$.
\item If a family $u_a\in\mathrm{BL}$ satisfies
\eqref{eq:app-one-start} for every $b\in I$, then
\eqref{eq:app-integrated-backward} holds.
\end{enumerate}
\end{lemma}
\begin{proof}
We first prove the Lipschitz estimate.
Start two copies of the Gamma process from states $s$ and $s'$ and drive
them with the same Poisson random measures. The integrated Lipschitz bound
on the jump kernels gives
\[
\mathbb E\|S_b^s-S_b^{s'}\|_1 \leq \|s-s'\|_1 + L_H \int_a^b \mathbb E\|S_t^s-S_t^{s'}\|_1 \,\mathrm dt.
\]
Gr\"onwall's inequality yields
\[
\mathbb E\|S_b^s-S_b^{s'}\|_1 \leq e^{L_H(b-a)} \|s-s'\|_1.
\]
Applying a Lipschitz function $g$ gives the claimed Lipschitz bound for
$\widehat{\overrightarrow K}_{a,b}g$. The supremum bound holds
because a Markov transition averages values of $g$ and therefore cannot
increase their maximum absolute value.
For later use, define
\[
c_g = e^{L_H(b_*-a_*)} \operatorname{Lip}(g).
\]
Next consider the generator. For any $f\in\mathrm{BL}$, split the jump
integral at $w=1$. For $w<1$,
\[
|f(s+we_i)-f(s)| \leq \operatorname{Lip}(f)w.
\]
For $w\geq1$,
\[
|f(s+we_i)-f(s)| \leq 2\|f\|_\infty.
\]
Therefore
\begin{equation}
\|\mathcal A_tf\|_\infty
\leq
d\operatorname{Lip}(f)
+
2M_H\|f\|_\infty.
\label{eq:app-gamma-bl-bound}
\end{equation}
To derive the backward equation, first remove all jumps smaller
than $\varepsilon\in(0,1)$. The remaining process has finite total rate
bounded by
\[
q_\varepsilon = d\log(1/\varepsilon)+M_H,
\]
so the standard backward equation for finite-rate jump processes holds.
Let
\[
v_s^\varepsilon = \widehat{\overrightarrow K}^{\varepsilon}_{s,b}g.
\]
The coupling argument above gives
\[
\operatorname{Lip}(v_s^\varepsilon) \leq c_g
\]
uniformly in $\varepsilon$. The cutoff estimate from
Lemma~\ref{lem:app-gamma-process} also gives
\[
\|v_t^\varepsilon-v_t\|_\infty \leq d(b_*-a_*)c_g\varepsilon.
\]
We now pass the cutoff backward equation to the limit. Write
\[
\mathcal A_t^\varepsilon v_t^\varepsilon-\mathcal A_tv_t = \mathcal A_t(v_t^\varepsilon-v_t) - (\mathcal A_t-\mathcal A_t^\varepsilon)v_t^\varepsilon.
\]
The second term is at most $d\varepsilon c_g$. For
\[
\chi=v_t^\varepsilon-v_t,
\]
we have both a supremum bound and a Lipschitz bound, so
\[
|\chi(s+we_i)-\chi(s)| \leq \min\{ 2c_gw, 2\|\chi\|_\infty \}.
\]
Splitting the jump integral at the point where these two bounds meet gives
\[
\|\mathcal A_t\chi\|_\infty \leq 2d\|\chi\|_\infty \left[ 1+\log^+ \left( \frac{c_g}{\|\chi\|_\infty} \right) \right] + 2M_H\|\chi\|_\infty.
\]
This tends to zero as
$\|\chi\|_\infty\to0$. If $c_g=0$, then $g$ is constant and the claim is
immediate. Bound~\eqref{eq:app-gamma-bl-bound} provides an integrable
dominating function in time. Hence the finite-rate backward equation
converges to the desired backward equation for the full Gamma process.
We next prove injectivity. Suppose
\[
\widehat{\overrightarrow K}_{a,b}g=0.
\]
Let
\[
v_s = \widehat{\overrightarrow K}_{s,b}g, \qquad \phi(s)=\|v_s\|_\infty.
\]
Subtracting the backward equations at $a$ and $s$ gives
\[
v_s = -\int_a^s\mathcal A_tv_t\,\mathrm dt.
\]
This time, split each Gamma generator at a small threshold
$\varepsilon$. Jumps below $\varepsilon$ contribute at most
\[
dc_g\varepsilon,
\]
while jumps above $\varepsilon$ have finite total rate
$q_\varepsilon$ and contribute at most
$2q_\varepsilon\phi(t)$. Therefore
\[
\phi(s) \leq dc_g\varepsilon(s-a) + 2q_\varepsilon \int_a^s\phi(t)\,\mathrm dt.
\]
Gr\"onwall's inequality gives
\[
\phi(s) \leq dc_g(s-a) e^{2M_H(s-a)} \varepsilon^{1-2d(s-a)}.
\]
If
\[
s-a\leq\theta<\frac1{2d},
\]
the exponent of $\varepsilon$ is positive. Letting
$\varepsilon\downarrow0$ gives $v_s=0$.
The same argument can then be restarted from the end of this short
interval. Repeating it finitely many times reaches $b$, and therefore
\[
g=v_b=0.
\]
This proves injectivity.
Finally, suppose \eqref{eq:app-one-start} holds. Preservation of
$\mathrm{BL}$ and the semigroup property give
\[
\widehat{\overrightarrow K}_{a_*,a} \left( \widehat{\overrightarrow K}_{a,b}u_b-u_a \right) = 0.
\]
Injectivity implies
\[
\widehat{\overrightarrow K}_{a,b}u_b=u_a.
\]
Thus the consistency relation holds between every two intermediate levels. Writing
\[
u_s = \widehat{\overrightarrow K}_{s,b_*}u_{b_*}
\]
and using the backward equation gives
\eqref{eq:app-integrated-backward}.
\end{proof}
\begin{corollary}\label{cor:app-gamma-training}
Suppose
\[
X\in[x_{\min},x_{\max}]^d \qquad\text{almost surely},
\]
and fix $a_*>0$. Assume the following.
\begin{enumerate} \item On every finite interval $I=[a_*,b_*]$, the candidate class satisfies Assumption~\ref{ass:app-direct} with $\alpha_I>1$ and also satisfies \begin{equation}
\sum_i\int_0^\infty
\bigl|\widehat H_{i,a}(s,w)-\widehat H_{i,a}(s',w)\bigr|\,\mathrm dw
\leq L_I\|s-s'\|_1.
\label{eq:app-gamma-training-lipschitz}
\end{equation}
The class contains the true predictions. Candidates are continuous in the
information level and jointly continuous in $(a,s,w)$.
\item Regression uses squared loss with a fixed target shape larger than
$1/2$, binary cross-entropy with a fixed target shape at least one, or
I-divergence with any fixed positive target shape. The targets satisfy
the integrability conditions of Lemma~\ref{lem:app-n2n}, and the regression
sampling law has coverage at $a_*$.
\item Every simulated path starts at $a_*$. The consistency sampling
law of the starting state, endpoint level, coordinate, and jump size
has full support on
\[
(0,\infty)^d \times (a_*,\infty) \times \{1,\ldots,d\} \times (0,\infty).
\]
\end{enumerate}
Then every global minimizer of
\eqref{eq:app-training-objective} equals the true Gamma posterior
predictions for every $a\geq a_*$, every positive state, every coordinate,
and every jump size $w>0$. The process generated from that minimizer and
initialized with $p_{a_*}$ has the true path law on every finite interval.
\end{corollary}
\begin{proof}
Fix
\[
I=[a_*,b_*].
\]
Since $\alpha_I>1$, the tail bound in
\eqref{eq:app-gamma-class} is integrable with respect to $\mathrm dw$.
Together with
\eqref{eq:app-gamma-training-lipschitz}, this verifies the assumptions of
Lemma~\ref{lem:app-gamma-process} for every candidate in the class.
The true Gamma kernel generates the true observation process.
Proposition~\ref{prop:app-population-objective} therefore gives exact
agreement with the true prediction at $a_*$ and zero consistency residual
almost surely under the candidate's consistency sampling law.
We next extend this almost-sure identity to every state, endpoint, and
queried jump size. The residual is jointly continuous in those variables:
the transition operators preserve Lipschitz bounds, the candidate is jointly
continuous and bounded, and the Gamma process is stochastically continuous
at every fixed level. Dominated convergence together with the coupling
estimate from Lemma~\ref{lem:app-gamma-injective} gives the required
continuity.
Because the consistency sampling law has full support, zero residual almost surely
implies
\[
\widehat{\overrightarrow K}_{a_*,b} \widehat H_{j,b}(\cdot,\zeta) = \widehat H_{j,a_*}(\cdot,\zeta)
\]
for every admissible tuple $(b,j,\zeta)$.
For fixed $(j,\zeta)$, the prediction as a function of the state belongs to
$\mathrm{BL}$. Lemma~\ref{lem:app-gamma-injective} then gives the
integrated consistency equation between all intermediate levels.
Corollary~\ref{cor:app-direct-continuation} identifies the predictions and
the process law on $I$.
Since $b_*$ was arbitrary, the conclusion holds for every
$a\geq a_*$.
\end{proof}
The true Gamma kernel satisfies these assumptions: its exponential
tail is stronger than any of the required polynomial tail bounds.
Moreover, integrating \eqref{eq:app-gamma-derivative} over jump size gives
\[
L_I = \frac{d\delta_X^2x_{\max}^2}{4}
\]
in \eqref{eq:app-gamma-training-lipschitz}.
Joint continuity is required so that an identity that holds almost surely
under the consistency sampling law extends to every tuple in its support.
\subsection{Generation from the empty state}
\label{app:zero-start}
The continuation theorems begin at a positive level $a_*$. Full generation,
however, begins from the empty or uninformative state.
The regression loss can identify the true predictions directly at the
levels covered by the noisy observations, namely $0<a\leq a_{\mathrm{in}}$. The continuation argument identifies them for $a\geq a_*$. If $a_*\leq a_{\mathrm{in}}$, these two ranges overlap and together cover every positive information
level.
\begin{corollary}
\label{cor:app-poisson-zero}
Suppose
\[
X\in[0,M]^d \qquad\text{almost surely}.
\]
Assume the count-model class conditions of
Corollary~\ref{cor:app-count-training}, with continuity at every positive
level.
Suppose the regression targets are unbiased and satisfy the integrability
conditions of Lemma~\ref{lem:app-n2n}, and
the regression sampling law has full support on
\[
(0,a_{\mathrm{in}}] \times \mathcal S \times \{1,\ldots,d\}.
\]
Assume also that the simulated paths satisfy the full-support condition of
Corollary~\ref{cor:app-count-training} for some
\[
0<a_*\leq a_{\mathrm{in}}.
\]
Then every global minimizer of
\eqref{eq:app-training-objective} equals the true Poisson jump rates on
\[
(0,\infty)\times \mathcal S.
\]
The process generated by these rates and started from zero is nonexplosive,
has the true Poisson measurement law, and satisfies
\[
\frac{N_\lambda}{\lambda} \longrightarrow X
\]
almost surely for a random image
$X\sim p_{\mathrm{clean}}$.
\end{corollary}
\begin{proof}
Proposition~\ref{prop:app-population-objective}, full support of the
regression sampling law, and continuity imply that the learned rates equal the true
rates throughout
\[
(0,a_{\mathrm{in}}].
\]
Corollary~\ref{cor:app-count-training} gives equality throughout
\[
[a_*,\infty).
\]
Since $a_*\leq a_{\mathrm{in}}$, these intervals overlap and cover all
positive levels.
The true total Poisson rate is bounded by
\[
dM.
\]
Uniformization therefore gives a unique nonexplosive process starting from
the zero state. The value assigned to the rate at the single level $a=0$
has no effect on the law of the process.
The clean limit then follows from
Proposition~\ref{prop:app-clean-limits}.
\end{proof}
The same argument applies to binomial and negative-binomial observations
under their respective class assumptions. For the negative-binomial model,
every moment index $k\geq1$ must be covered. For binomial observations,
the argument is applied on finite intervals below $p=1$.
For Gamma observations, starting from zero additionally requires the
entrance condition of Assumption~\ref{ass:app-gamma-entrance}.

\section{Sampling the learned process}
\label{app:sampling}
The continuation theorems above concern the exact stochastic processes:
the information level is continuous, all jump sizes are retained, and the
process is followed to its theoretical clean limit.
The numerical samplers used in the experiments use finitely many
information levels, may truncate small or large jumps, and stop at a finite
information level. The theorems do not cover these approximations, which we
describe here.
\subsection{Finite steps}
\label{app:sampler-steps}
For Gaussian observations, we discretize the stochastic differential
equation with Euler--Maruyama:
\[
z_{j+1} = z_j + (a_{j+1}-a_j)\widehat m_{a_j}(z_j) + \sqrt{a_{j+1}-a_j}\,\eta_j, \qquad \eta_j\sim\mathcal N(0,\Id).
\]
For Poisson observations, a tau-leap freezes the current posterior rate over
one step:
\[
n_{j+1,i} = n_{j,i}+P_{j,i}, \qquad P_{j,i} \sim \operatorname{Pois} \bigl( (\lambda_{j+1}-\lambda_j) \widehat h_{i,\lambda_j}(n_j) \bigr),
\]
independently across coordinates.
This frozen-rate approximation does not reproduce the exact conditional
variance of the increment. Under the true process,
\[
\begin{aligned} & \operatorname{Var} ( N_{\lambda_{j+1},i} - N_{\lambda_j,i} \mid N_{\lambda_j}=n_j ) \\ &\qquad= (\lambda_{j+1}-\lambda_j) h_{i,\lambda_j}(n_j) + (\lambda_{j+1}-\lambda_j)^2 \operatorname{Var} ( X_{i}\mid N_{\lambda_j}=n_j ). \end{aligned}
\]
The Poisson tau-leap reproduces the first term but omits the second, which
comes from uncertainty in the latent clean intensity during the step.
For binomial observations, the frozen birth rate is
\[
\frac{ \widehat D_{i,p_j}(n_j)-n_{j,i} }{ 1-p_j }.
\]
A learned prediction can fall below $n_{j,i}$, which makes this rate negative.
A Poisson approximation to the number of births can also leave the reachable
state space, although the exact binomial process cannot. The experiments
instead clip the predicted number of remaining births at zero, round it, and
draw the increment from the binomial law,
\[
n_{j+1,i}=n_{j,i}+\operatorname{Binomial}\Bigl(\operatorname{round}\bigl((\widehat D_{i,p_j}(n_j)-n_{j,i})_+\bigr),\,
\frac{p_{j+1}-p_j}{1-p_j}\Bigr),
\]
which cannot overshoot the predicted count. Clipping and rounding change the
transition law, and, like the Poisson tau-leap, this increment ignores the
uncertainty in $X_{i}$ during the step.
We do not prove convergence of these finite-step schemes for arbitrary
learned prediction functions.
\subsection{Sampling jump sizes}
\label{app:sampler-jumps}
For Gamma observations, keeping only jumps of size $w\geq\varepsilon$ turns the infinite-activity process into a finite-rate process. The total
rate is finite under either the true integrability bound
\eqref{eq:app-gamma-integrability} or the learned tail bound
\eqref{eq:app-gamma-class}.
Suppose, for example, that
\[
X_{i}\leq x_{\max}.
\]
Then the true posterior transform satisfies
\[
H_{i,L}(s,w)\leq e^{-w/x_{\max}}.
\]
If a learned kernel satisfies the same bound, one can simulate candidate
jumps in coordinate $i$ from the dominating measure
\[
\mathbf 1_{\{w\geq\varepsilon\}}\,e^{-w/x_{\max}}\frac{\mathrm dw}{w}
\]
and accept a proposed jump of size $w$ with probability
\[
e^{w/x_{\max}} \widehat H_{i,L}(s,w),
\]
evaluated at the current information level and state. Updating the state
after every accepted event gives an exact simulation of the
$\varepsilon$-cutoff process.
The effect of discarding small jumps can be bounded locally. If
$0\leq H_i\leq1$, then jumps smaller than $\varepsilon$ contribute at
most
\[
dT\varepsilon
\]
to the expected total added state over an interval of length $T$.
For a Lipschitz test function $f$, their direct contribution to the generator
is at most
\[
d\varepsilon\operatorname{Lip}(f).
\]
If the integrated state-Lipschitz assumption of
Lemma~\ref{lem:app-gamma-process} also holds, the coupling argument in that
lemma gives the stronger pathwise estimate
\[
\mathbb E \sup_{0\leq t\leq T} \|S_{a+t}-S_{a+t}^{\varepsilon}\|_1 \leq d\varepsilon T e^{L_HT}.
\]
Without such a stability assumption, controlling the jumps omitted in the
current step is not enough to control their effect on all future
state-dependent rates.
For negative-binomial observations, suppose
\[
q_i\leq\bar q<1.
\]
Then
\[
M_{i,k,r}\leq\bar q^k.
\]
An exact event sampler for any learned kernel obeying the same bound can
therefore use the following rejection scheme.
Propose events at total rate
\[
-d\log(1-\bar q).
\]
Choose the coordinate uniformly. Then choose the jump size $k$ with
probability
\[
\frac{ \bar q^k }{ k[-\log(1-\bar q)] }.
\]
Accept the proposed jump with probability
\[
\frac{ \widehat M_{i,k,r}(n) }{ \bar q^k },
\]
evaluated at the current level and state. This construction uses all
$k\geq1$ and therefore does not require a moment cutoff.
The image experiments instead freeze rates on a finite grid and predict
only finitely many moments. For the true process, omitting all jumps with
$k>k_{\max}$ removes at most
\[
d\sum_{k>k_{\max}} \frac{\bar q^k}{k}
\]
of the total jump intensity and at most
\[
\frac{ d\bar q^{k_{\max}+1} }{ 1-\bar q }
\]
of the expected count added per unit shape.
For
\[
\bar q=0.9, \qquad k_{\max}=30,
\]
these quantities are approximately $0.4\%$ of the total jump intensity
and $4\%$ of the expected count rate at a pixel with $q_i=\bar q$.
They are smaller for darker pixels.
These calculations quantify what is removed directly by truncating the
true jump kernel. They do not, by themselves, give a bound on the distance
between the final generated image distributions, because errors can
propagate through later state-dependent rates.
\subsection{Starting and stopping}
\label{app:sampler-endpoints}
Generation begins from the least informative state and passes through the
range in which noisy regression directly supervises the model.
For Gaussian, Poisson, binomial, and negative-binomial processes, zero belongs
to the state space and can serve as the initial state. For Gamma observations, zero lies on the boundary
of the state space used in the interior jump formulation, so starting from
zero requires the entrance assumption in
Appendix~\ref{app:model-gamma}.
In practical Gaussian implementations one often begins at a large but
finite noise level rather than at the exact infinite-noise limit, which is
another numerical approximation.
At the other end of the trajectory, stopping at a finite information level
does not produce an exactly clean observation. For example, conditional on
$X=x$,
\[
\operatorname{Var} \left( \frac{N_\lambda}{\lambda} \middle| X=x \right) = \frac{x}{\lambda},
\]
and
\[
\operatorname{Var} \left( \frac{S_L}{L} \middle| X=x \right) = \frac{x^2}{L}.
\]
Thus the normalized state still contains measurement noise at every finite
$\lambda$ or $L$.
An alternative is to output the posterior mean of the clean image at the
stopping state. This removes conditional measurement noise by averaging,
but it has a different distribution from the clean data at finite
information level. Indeed, whenever $X_{i}$ has finite variance,
\[
\operatorname{Var} ( \mathbb E[X_{i}\mid Z_a] ) = \operatorname{Var}(X_{i}) - \mathbb E \operatorname{Var} ( X_{i}\mid Z_a ).
\]
The posterior mean therefore has lower variance than the clean image unless
the remaining posterior uncertainty is zero.
We state explicitly when a reported metric is computed from the posterior
mean rather than from the normalized stochastic state.

\section{Experimental details and additional results}
\label{app:experiments}
\subsection{Networks and the estimate of the consistency loss}
\label{app:networks-loss}
The CIFAR-10 experiments use a U-Net with approximately 60 million
parameters. The Poisson and binomial networks predict posterior means.
The negative-binomial network additionally predicts the posterior moments
needed for its jump rates. The Gamma network predicts, for each color channel, the parameters of the mixture \eqref{eq:app-radar-mixture} used for radar, and a posterior mean, used only for the final image. The radar experiments use a U-Net with approximately 9 million parameters.
At every pixel it predicts the parameters of the mixture model described in
Appendix~\ref{app:radar}. Our implementation builds on the Poisson-jump
code of~\cite{chen2023learning}.\footnote{\url{https://github.com/tqch/poisson-jump}}

The networks predict the required posterior quantities directly. They do
not enforce the common-density-ratio representation studied in
Appendix~\ref{app:common}. Most experiments replace the squared loss in \eqref{eq:pf-loss} by another Bregman divergence with the same minimizer (Lemma~\ref{lem:app-n2n}, Table~\ref{tab:app-settings}).
For Poisson and binomial observations, the outputs are clipped to a fixed
nonnegative range. This gives a uniform output bound. The binomial
implementation, however, does not explicitly represent the reachable set
$\mathcal S$, so it does not guarantee that rates on forbidden edges vanish, as
required by Assumption~\ref{ass:app-direct}.
For negative-binomial observations, the output for the $k$th moment is a
sigmoid multiplied by $0.9^k$, where $0.9$ is the largest value of $q(X)$. Thus the network satisfies a geometric
upper bound in $k$. The implementation represents only finitely many moments, whereas
the theorem uses all $k\geq1$.

For the Gamma models, the mixture parameterization makes each pixelwise prediction a valid
Laplace transform. It does not by itself ensure the uniform tail and
state-Lipschitz conditions of Corollary~\ref{cor:app-gamma-training}, which we have
not verified over the full trained network class.
The experiments therefore test the method under approximations that the exact
continuation results do not cover: finite parameterizations, finite coverage of
levels, states and jump sizes in training, estimated losses and finite-step simulation.

\paragraph{Monte Carlo estimate.}
To estimate the consistency loss \eqref{eq:main-consistency-loss}, we start from a state $z$
at level $a$ and simulate two conditionally independent paths with the current model to
level $b$. Along path $j$, the residual
\[
\mathcal R^{(j)} = \widehat h_{\xi,a}(z) - \widehat h_{\xi,b}(Z_b^{(j)}),
\qquad j=1,2,
\]
is the difference between the prediction at the start and the prediction at the end.
Conditional on $z$, $a$, $b$ and $\xi$, its mean is the consistency residual
$\mathcal R_\xi(a,b,z)$ of \eqref{eq:app-consistency-residual}, and
\begin{equation}
\mathbb E\,\langle \mathcal R^{(1)},\mathcal R^{(2)}\rangle = \|\mathcal R_\xi(a,b,z)\|^2,
\label{eq:app-consistency-estimate}
\end{equation}
which is the squared norm in \eqref{eq:main-consistency-loss}. The inner product of the
two residuals is therefore an unbiased estimate of the loss, assuming finite second moments
and exact simulation of the paths.
For the jump laws it is enough to start every path at the training level, $a=a_{\rm in}$,
as we do in all experiments. Consistency from $a_{\rm in}$ to every later level then implies
consistency between any two later levels (Lemmas~\ref{lem:app-count-injective}
and~\ref{lem:app-gamma-injective}), and an exact minimizer has the true predictions at every
level in $[a_{\rm in},b_*]$ (Corollaries~\ref{cor:app-count-training}
and~\ref{cor:app-gamma-training}).

\begin{table}[htbp]
\centering
\scriptsize
\setlength{\tabcolsep}{2.5pt}
\renewcommand{\arraystretch}{1.15}
\begin{tabularx}{\linewidth}{@{}>{\raggedright\arraybackslash}p{2.1cm}*{6}{>{\raggedright\arraybackslash}X}@{}}
\toprule
 & Poisson & Binomial & Neg.\ binomial & Gamma & Conv-PC & SAR \\
\midrule
N2N loss & I-divergence, $\delta_\tau$-weighted & I-divergence & squared, variance-weighted & BCE $+$ squared mean & I-divergence, $\delta_\tau$-weighted & I-divergence kernel, Bregman mean with $\phi''(v)=v^{-1/2}$ \\
N2N levels & uniform $\tau$, atom $\tau_{\rm obs}/T$ & uniform $\tau$, atom $\tau_{\rm obs}/T$ & power law, atom $0.5$ & $L_{\rm obs}u^3$ & uniform $\tau\ge \tau_{\min}$; refinement adds atom $\tau_{\min}/T$ & $L_{\rm obs}u^{3/2}$ \\
Path steps & 32 & 4 & 32 & 16 & 4 & 24 \\
Path endpoints & uniform $\tau<\tau_{\rm obs}$ & uniform $\tau<\tau_{\rm obs}$ & log-uniform $(r_{\rm obs},10r_{\rm obs}]$ & log-uniform $(L_{\rm obs},50L_{\rm obs}]$ & uniform $\tau<\tau_{\min}$ & log-uniform $(L_{\rm obs},50L_{\rm obs}]$ \\
Path starts per device batch & $\le 1$ & $\le 8$ & $\le 1$ & 4 & $\le 8$ & 4, every third step \\
$\lambda_{\rm cons}$ & 8 & 1 & 8 & 0.3 & 0.003 & 1 \\
Gradient at $a_{\rm in}$ & yes & yes & yes & detached & yes & detached \\
Refinement & 200 epochs from 1800 & 200 from 1800 & 200 from 1800 & 200 from 1800 & 20 epochs from 100 & 40 epochs ($2\times20$) \\
Sampler steps & 1000 & 1000 & 2000 (power grid), then 500 to $10r_{\rm obs}$ & 250 & 1000 & 400 to $2L_{\rm obs}$ \\
Trajectory output & state $N_\lambda/\lambda$ & predicted image & predicted mean at $10r_{\rm obs}$ & predicted mean at $50L_{\rm obs}$ & predicted image & -- \\
Extra term & -- & -- & -- & -- & -- & rollout regression \\
\bottomrule
\end{tabularx}
\caption{Training and sampling settings. The CIFAR-10 models share the U-Net, batch size 128,
learning rate $2\cdot10^{-4}$ and EMA decay $0.9999$. Paths start at $a_{\rm in}$ in every experiment. In the N2N rows, $\delta_\tau$ is the exposure added by the sampling step at timestep $\tau$, $T=1000$ is the number of timesteps, $\tau_{\rm obs}$ is the timestep of $a_{\rm in}$, and $u$ is uniform on $(0,1)$.
The radar model simulates four paths and averages the residuals in pairs before taking their inner product. The \textsc{Conv-PC} model is a 35M-parameter
U-Net trained for 100 epochs of 500 steps on 128-pixel crops of single frames, with batch size 32,
learning rate $10^{-4}$ and EMA decay $0.9999$. For this model, $a_{\rm in}$ lies at timestep $\tau_{\min}=634$ of 1000, where the input
holds 98\% of a frame.}
\label{tab:app-settings}
\end{table}

\subsection{CIFAR-10 with Poisson, binomial, negative-binomial and Gamma observations}
\label{app:cifar-counts}
The CIFAR-10 experiments let us vary the measurement law and the information level of the training data
while keeping the underlying image distribution fixed. They test both the
importance of matching the process to the measurements and the effect of training with the consistency loss beyond the range supervised by noisy views.

\paragraph{Networks, views and training.}
All CIFAR-10 models use the same U-Net of approximately 60 million parameters and
the same training budget. For each
$32\times32$ image, we draw two conditionally independent noisy views, use one
as the network input and the other as the regression target, and redraw both at
every epoch. Base training runs for 1800 epochs and is followed by 200 epochs of
refinement with the consistency loss. Each training seed is refined from its own base
checkpoint. We compute FID from 50,000 samples generated with an exponential
moving average of the weights, using the clean CIFAR-10 training set as the
reference. Poisson and binomial trajectories use 1000 sampling steps, negative-binomial trajectories use 2000 up to $r_{\rm obs}$ and 500 above it, and Gamma trajectories 250.

Because both views are redrawn, these experiments assume that the acquisition can be repeated. In the microscopy and radar experiments the recorded measurement is fixed.

\paragraph{Information levels.}
The implementation indexes the process by a timestep $\tau$ that decreases as the observation
becomes more informative. After inversion and rescaling, $\tau$ corresponds to the information level $a$ used in the theory.
Table~\ref{tab:app-cifar-levels} gives the Poisson and binomial levels of
Table~\ref{tab:exp-cifar-generation} as timesteps and law parameters.

In the Poisson process the level at timestep $\tau$ is $\lambda_{\max}\alpha_\tau$ with $\lambda_{\max}=4.65\cdot10^{4}$, so the training data have level $\lambda_{\rm obs}=\lambda_{\max}\alpha_{\tau_{\rm obs}}$, which is the expected count at a white pixel quoted in the main text. In the binomial process, each pixel holds as many particles as its 8-bit clean value, so a white pixel holds $X_i=255$ particles and its expected count is $255\,p_{\tau_{\rm obs}}$.

Negative-binomial observations have no comparable count axis, because the factor $(1-q)^2$
in the delta-method variance makes bright pixels much quieter and dark pixels much noisier
than a Poisson observation of the same nominal budget. We place them instead by the root
mean squared error of the best unbiased single-view estimator of $X$ under the CIFAR-10
pixel law. On that scale $r_{\rm obs}=5$, $16$ and $50$ are as informative as $15.5$,
$55.2$ and $178.3$ Poisson counts, so the outer two negative-binomial rows of
Table~\ref{tab:exp-cifar-generation} are matched to its Poisson rows at $16.1$ and $178.4$
counts.

Gamma observations are indexed by the number of looks. The best unbiased single-view
estimator is $S_{L_{\rm obs}}/L_{\rm obs}$, whose relative error $1/\sqrt{L_{\rm obs}}$ is $12.5\%$ at
$L_{\rm obs}=64$ and, unlike the Poisson case, is the same at every pixel. A Poisson
observation with $64$ counts at a white pixel has the same relative error there, so the Gamma
row lies between the $16.1$- and $178.4$-count rows. This equivalence holds only at a white pixel.

\begin{table}[htbp]
\centering
\small
\begin{tabular}{@{}lllr@{}}
\toprule
Process & Timestep & Law parameter & Counts at a white pixel \\
\midrule
Poisson  & $\tau=440$ & $\alpha=1.3\cdot10^{-2}$ & $610$ \\
Poisson  & $\tau=500$ & $\alpha=3.8\cdot10^{-3}$ & $180$ \\
Poisson  & $\tau=600$ & $\alpha=3.5\cdot10^{-4}$ & $16$ \\
\midrule
Binomial & $\tau=132$ & $p=0.71$   & $180$ \\
Binomial & $\tau=226$ & $p=0.39$   & $100$ \\
Binomial & $\tau=364$ & $p=0.098$  & $25$ \\
Binomial & $\tau=448$ & $p=0.031$  & $8$ \\
\bottomrule
\end{tabular}
\caption{The training levels $a_{\rm in}$ of Table~\ref{tab:exp-cifar-generation}, as parameters of
the observation law: the counts are $\lambda_{\max}\alpha$ with $\lambda_{\max}=4.65\cdot10^{4}$ for the
Poisson process and $255\,p$ for the binomial one. Table~\ref{tab:exp-cifar-generation}
quotes the exact values to one decimal. Here they are rounded. The
negative-binomial and Gamma settings have no count axis and are placed on an equal-accuracy
scale in the text.}
\label{tab:app-cifar-levels}
\end{table}

The four families continue to different levels. The Poisson and binomial trajectories run to the end of the grid, for Poisson roughly $76\lambda_{\rm obs}$ at $613$ counts and $2900\lambda_{\rm obs}$ at $16$ counts. The negative-binomial trajectories stop at $10r_{\rm obs}$, ten times their observed shape, and the Gamma trajectories at $50L_{\rm obs}$. In each family, this is the largest level that the simulated paths reach during refinement.

\paragraph{Sampling the consistency term.}
Each training image draws one timestep $\tau$. If $\tau\ge \tau_{\rm obs}$, the image contributes an N2N
term at $\tau$, with input $\operatorname{Binomial}(Y_{\rm in},\alpha_\tau/\alpha_{\tau_{\rm obs}})$. Otherwise it
contributes an N2N term at $\tau_{\rm obs}$, with input $Y_{\rm in}$, and a simulated path from $Y_{\rm in}$ to
timestep $\tau$. For the Poisson and binomial models the
N2N levels carry an atom of mass $\tau_{\rm obs}/T$ at $\tau_{\rm obs}$, and path endpoints are uniform
over the timesteps no pair of views covers. No gradient flows through the simulated paths.
Table~\ref{tab:app-settings} lists for each experiment the number of path steps, the path starts per
batch, the consistency weight and whether the prediction at $a_{\rm in}$ is detached.
The Gamma and radar refinements degrade when this prediction is not detached.

\paragraph{Misspecified measurement laws.}
The two Gaussian baselines of Table~\ref{tab:exp-cifar-misspec} share the architecture and
budget of the Poisson model and differ only in the law they assume. The direct baseline uses
one global standard deviation $\sigma_{\rm obs}=\sqrt{\bar x/\lambda_{\rm obs}}$, the Poisson
standard deviation evaluated at the mean CIFAR-10 pixel value $\bar x=0.4734$ instead of at
each pixel. The Anscombe baseline works on $A(y)=2\sqrt{\lambda_{\rm obs}y+3/8}$, rescaled so
that a white pixel maps to one. Since $A(y)$ has approximately unit variance, the rescaled noise level is
$\sigma_{\rm obs}=1/(2\sqrt{\lambda_{\rm obs}+3/8})$. This baseline returns to intensity space
through the algebraic inverse of $A$, clipped to $[0,1]$, which is biased.
Both baselines use the true photon rate $\lambda_{\rm obs}$.
Each stops at the noise level of its training measurements, the smallest timestep $\tau$ whose noise variance
$(1-\bar\alpha_\tau)/\bar\alpha_\tau$ reaches $\sigma_{\rm obs}^2$, and returns the predicted posterior mean there.
The Poisson model is better by $11.0$--$17.5$ FID at the two higher counts, and at the lowest
count the baselines' FID is more than six times larger.

The most extreme misspecification treats each single view as a clean image. With the same
architecture and 1800 epochs, a diffusion model trained this way gives FID $148.09$ at 25 binomial counts, against $17.38$ with
two views. Under Gaussian diffusion at 178 Poisson counts it gives $69.42$, against $8.62$.
Both gaps, $130.7$ and $60.8$ FID, arise at $a_{\rm in}$, before any continuation.

\begin{table}[htbp]
    \centering
    \small
    \setlength{\tabcolsep}{8pt}
    \begin{tabular}{lrrr}
        \toprule
        & \multicolumn{3}{c}{Counts at a white pixel} \\
        \cmidrule(lr){2-4}
        Law assumed for the measurements & 613.1 & 178.4 & 16.1 \\
        \midrule
        Poisson (ours)          & $6.50 \pm 0.42$ & $8.62 \pm 0.03$ & $16.74 \pm 0.18$ \\
        Gaussian, one variance  & $17.49 \pm 0.77$ & $25.58 \pm 2.73$ & $110.02 \pm 1.91$ \\
        Gaussian after Anscombe & $21.13 \pm 0.74$ & $26.16 \pm 0.83$ & $111.03 \pm 13.47$ \\
        \bottomrule
    \end{tabular}
    \caption{A misspecified measurement law. All models stop at the noise level of the training measurements and return the predicted posterior mean. FID from 50,000
    samples, mean $\pm$ standard deviation over two training seeds. All three models are 1800-epoch checkpoints.}
    \label{tab:exp-cifar-misspec}
\end{table}

\paragraph{Calibration of the continued prediction.}
We measure calibration of the posterior-mean prediction by the slope
\[
\frac{ \operatorname{Cov}(\widehat X,X) }{ \operatorname{Var}(\widehat X) },
\]
which equals one for a calibrated predictor, $\E[X\mid \widehat X] = \widehat X$. States are drawn from the
measurement law $Q_a(x,\cdot)$ of 2048 held-out CIFAR-10 test images $x$, so levels
above $a_{\rm in}$ can be evaluated even though no pair of views supervises them.
We compare the refined model with a zero-weight control, trained for the same 200 epochs with
$\lambda_{\rm cons}=0$. The control and the refined model are the No Consistency and Consistency columns
of Table~\ref{tab:exp-cifar-generation}. Above $a_{\rm in}$, the slope of the control is off
by $(0.50$--$1.9)\cdot10^{-2}$ at 178 Poisson counts (two training seeds) and by
$(0.40$--$1.3)\cdot10^{-2}$ at 100 binomial counts (one seed). The refined slopes are off by only
$(1.3$--$9.7)\cdot10^{-4}$ and $(1.2$--$1.8)\cdot10^{-3}$. Since the control differs from the
refined model only in the consistency weight, the improvement comes from the consistency term.

\paragraph{Jump rates for the negative binomial.}
The negative-binomial models use the linear encoding $q(X)=0.9X$ and draw two
conditionally independent views from $\operatorname{NB}(r_{\rm obs},q(X))$. The network
predicts the posterior moments
\[
M_{i,k,r}(n) = \E[\,q_i(X)^k \mid N_r=n\,],\qquad k=1,\ldots,66,
\]
which the sampler needs for its jump rates. They are trained on the unbiased targets in
\eqref{eq:app-nb-target}. A separate output predicts the final image. It is trained
on $T_{1,r_{\rm obs}}(Y_{\rm tar})$ and divided by $0.9$ at sampling time, so the rescaled output
estimates $\E[X\mid N_r]$, as does the rescaled $k=1$ moment. The two
agree to within $0.01$ dB in reconstruction quality. Sampling uses tau-leaping on a power-law
grid of 2000 shape steps, which captures 63 of the 64 modes of a finite-mixture oracle on 64
images, against 38 for a uniform grid of the same length. Above $r_{\rm obs}$ the trajectory
continues with 500 geometrically spaced steps to $10r_{\rm obs}$.

A geometric weighting of the moment loss in $k$ leaves large $k$ weakly supervised. The models
reported in Table~\ref{tab:exp-cifar-generation} instead weight each moment target by its
estimated variance, which supervises every $k$. They are sampled with all 66 predicted moments and no further cutoff. On the refined checkpoint
at $r_{\rm obs}=50$, FID from 10,000 samples is $10.32$, $9.91$ and $9.97$ for cutoffs
$k_{\max}=10$, $30$ and $66$, so it barely depends on the cutoff. The geometrically weighted model needs a cutoff at
$k_{\max}=30$, because its weakly supervised large-$k$ outputs produce isolated bright pixels.
With this cutoff it reaches a lower FID at $r_{\rm obs}$ than the
variance-weighted model. We report the variance-weighted model because a cutoff changes the
process that the theory specifies.

If the rate for each jump size is instead computed from the predicted first moment $\widehat q=\widehat M_{i,1,r}$ as
$\widehat q^{\,k}/k$, these rates carry no information beyond the first moment. On the same checkpoint, with
sampling stopped at $r_{\rm obs}$, FID rises from $9.61$ with the predicted moments to $18.50$ with
rates computed from the mean. With the
network and the sampling seed fixed, the same comparison on the geometrically weighted model at
1200 epochs and $k_{\max}=30$ gives $7.88$ with the predicted moments and $13.97$ with rates
computed from the mean. The predicted higher moments thus have a much larger effect than the cutoff.

\paragraph{Laplace transforms for the Gamma process.}
The mixture outputs are trained on the target $T_{w,L_{\rm obs}}(Y_{\rm tar})$ of Proposition~\ref{prop:app-gamma-target}, with one $w$ per image drawn log-uniformly. For $L_{\rm obs}\geq 1$ this target lies in $[0,1]$, so we use the binary cross-entropy, which, like the squared error, is minimized by the posterior Laplace transform $H_{i,L}(S_L,w)= \E[e^{-w/X_{i}} \mid S_L]$, where $S_L$ is the network input obtained by degrading $Y_{\rm in}$ to a level $L \leq L_{\rm obs}$. The radar target has shape $1/2$ and infinite variance, which is why radar uses the I-divergence instead (Appendix~\ref{app:radar}). The mean output is trained with the squared error on $Y_{\rm tar}/L_{\rm obs}$. The sampler uses only the mixture. 

\paragraph{Entrance level of the Gamma process.}
The Gamma process reaches $S_0=0$ only as $L\downarrow0$: for every $\varepsilon>0$, the state
$S_\varepsilon$ is almost surely positive. Unlike for the count models
(Appendix~\ref{app:zero-start}), the sampler must therefore start at a positive level,
and we choose this level below the training range. The network is trained on
$[\varepsilon_0,L_{\rm obs}]$ with $\varepsilon_0=10^{-4}L_{\rm obs}$, where the zero state is
no longer typical. We start from $S=0$ at $10^{-5}L_{\rm obs}$, where it still is, and advance
to $\varepsilon_0$ with the exact kernel for a clean-image law that has independent pixels.
Every step above $\varepsilon_0$ uses the learned kernel. This falls short of
Assumption~\ref{ass:app-gamma-entrance}, which requires the law $p_{\mathrm{clean}}Q_\varepsilon$.
The entrance laws we construct are consistent as $\varepsilon\downarrow0$ and have the
per-pixel marginals of $p_{\mathrm{clean}}$, but not its spatial dependence.

Two measurements indicate that this substitution has little effect. First, the entrance state
carries a fraction $\varepsilon_0/L_{\rm obs}=10^{-4}$ of the mass of the state at $L_{\rm obs}$, and
replacing its spatial content by uniform, shuffled or smooth-noise scale fields changes the
result only within noise. Second, the independent-pixel law has an $8\times8$ block
correlation of $0.125$, the value for white noise, while the generated samples reach $0.807$
against $0.799$ for real data, so this structure comes from the learned
kernel. The independent-pixel law is used only for sampling, and training is Noise2Noise on
two noisy Gamma views throughout.

The Gamma trajectories return the predicted posterior mean at $50L_{\rm obs}$ because the endpoints of the simulated paths are
drawn log-uniformly on $[L_{\rm obs},50L_{\rm obs}]$, so $82\%$ of them lie beyond
$2L_{\rm obs}$. Stopping earlier, at $2L_{\rm obs}$, gives $7.11\pm0.01$ against
$7.01\pm0.27$ at $50L_{\rm obs}$. The difference is smaller than the $0.143$ change from
redrawing the 50,000 samples.

\paragraph{Seeds and FID resampling.}
Every configuration in Table~\ref{tab:exp-cifar-generation} is trained with two seeds, each
refined from its own 1800-epoch checkpoint, and the table reports their mean and standard
deviation. The seed spread does not include the variation of FID
across sample sets. Scoring one checkpoint on a fresh set of 50,000 samples changes FID by
$0.143$. A standard deviation smaller than this reflects only the agreement between the two
seeds.

\subsection{Photon-counting microscopy}
\label{app:microscopy}
\paragraph{Generation.}
The \textsc{Conv-PC} dataset contains repeated microscopy frames consisting of photon
counts. Each observed frame is randomly split into a thinned input and its
complement. This matches the subset-training construction used by
Generative Accumulation of Photons~\citep{krull2024image}.
As in GAP, training includes the frames of the test field of view, which supplies reference crops and the frame shown in Figure~\ref{fig:exp-convpc-main}.
For the main generation experiment, we draw 10,000 samples from each model.
All methods are rendered using the same fixed intensity range $[0,4]$ and
are evaluated with the same clean-FID pipeline.
Starting from one base checkpoint, we run 20 epochs of refinement with the
consistency loss for three random seeds. We use
consistency weight $0.003$, two simulated paths per starting state and four
path steps beyond $a_{\rm in}$. Generation starts from zero
counts. Sampling either stops at the exposure of one frame or continues to the reference exposure of 512 frames. The weight was chosen on a pilot run with 256 samples.

At $64\times64$ resolution, refinement with the consistency loss improves over our
reproduction of GAP when either a fixed rendering range or a percentile
computed from the reference distribution is used. If each image is clipped
using its own percentiles, the ordering reverses.
The published GAP FIDs, $134.20$ and $86.82$, are not directly
comparable because the public archive and the rendering differ from the
original.

All scores above use the predicted image at the final state, whereas the reference crops are
count images. With Poisson noise at the reference exposure added to every predicted image,
stopping at one frame scores $42.4$ and the refined continuation $35.6$ at $256\times256$. Part of
the FID of the predicted images therefore reflects the noise texture of the reference. With the
added noise, the continuation without the consistency loss still fails (FID $227$), and GAP scores
$73.6$.
\paragraph{Conditional continuation.}
The $512$-frame reference in Figure~\ref{fig:exp-convpc-continuation} is not
a clean image: it averages $39.5$ counts per pixel in the displayed crop, so
it carries $16\%$ shot noise at its mean and more in the dark lumens. With
every image divided by its mean, the images predicted by N2N and GAP from the frame, panels (b) and (c), differ
from the reference by $0.49$ and $0.51$ RMSE, a single continuation from
either method by $0.63$ and $0.64$, and the N2N-only continuation by $1.19$.
Since one frame of $5{,}020$ photons does not determine the scene, a sample should differ from
the reference by about the posterior spread. The errors are indeed largest where the spread
over $500$ continuations is large: $1.06$ error and $0.44$ spread on the bright walls, against
$0.34$ and $0.14$ in the lumens.

Panel~(h) of Figure~\ref{fig:exp-convpc-continuation} shows the spread of the predicted image at the endpoint, relative to its pixelwise
mean. It concentrates on the wall boundaries and on the dim cluster at the upper left. Count
images at the reference exposure spread more, because they add Poisson shot noise, which
accounts for $23\%$ of their variance on the walls and $58\%$ in the lumens. Without the shot
noise, their relative spread is $0.22$, against $0.19$ for the predicted image.

\paragraph{Restoration and data access.}
The restoration experiment follows the transductive protocol used by GAP.
The released GAP training split contains frames from the field of view used
for testing. Under the matched Noise2Noise protocol, all frames from this
field of view may therefore also be used during training.
We evaluate all 512 frames using GAP's scale-normalized PSNR.
With each frame randomly split for Noise2Noise training, three training seeds give $25.081$,
$24.993$ and $25.038$ dB, against $24.917$ dB for our reproduction of GAP, a mean improvement
of $0.12\pm0.04$ dB. When both models instead use 511-frame leave-one-out targets, the
Noise2Noise advantage is $1.031$ dB (Table~\ref{tab:app-microscopy-psnr}).
If Noise2Noise is trained only on the other three fields of view, excluding the test field of
view entirely, the PSNR falls to $23.168\pm0.005$ dB.
\begin{table}[htbp]
\centering
\small
\begin{tabular}{@{}lr@{}}
\toprule
Method & PSNR (dB) \\
\midrule
GAP, split & $24.917\pm0.005$ \\
N2N, split, three seeds & $25.037\pm0.044$ \\
N2N, 511-frame target, 256-pixel crops & $26.836\pm0.010$ \\
GAP, 511-frame target & $25.805\pm0.009$ \\
\bottomrule
\end{tabular}
\caption{Single-frame \textsc{Conv-PC} restoration under transductive training.
The row with three training seeds reports the standard deviation across seeds.
The other rows report the standard error across the 512 evaluated frames.
The published GAP result for the split setting is $24.84$ dB.}
\label{tab:app-microscopy-psnr}
\end{table}

\subsection{Sentinel-1 radar}
\label{app:radar}
\paragraph{Views and preprocessing.}
We model the sum of $L$ radar looks as
\[
S_L\mid X \sim \operatorname{Gamma}(L,1/X).
\]
The normalized intensity $S_L/L$ has conditional mean $X$ and relative conditional
variance $1/L$. The data consist of three Sentinel-1 interferometric-wide
single-look-complex acquisitions from 2024. After TOPS deramping, the
squared real and imaginary components provide two half-looks, which serve
as the input and target views $Y_{\rm in}$ and $Y_{\rm tar}$. Following
\cite{dalsasso2021if}, we model each as
$\operatorname{Gamma}(1/2,1/X)$ given $X$, so $L_{\mathrm{obs}}=1/2$.
Under this model, the ratio $Y_{\rm in}/(Y_{\rm in}+Y_{\rm tar})$ has a beta distribution with
variance $0.125$. The empirical variance is $0.1247$. After correcting for image texture,
the estimated conditional correlation between the two views is $+0.003$.

Raw neighboring radar samples still have appreciable speckle correlation:
approximately $0.31$ in range and $0.44$ in azimuth. To reduce this
dependence, we separate each image into the four interlaced
$2\times2$ sublattices. This keeps all pixels but increases the spacing
between neighboring samples in each resulting image. The empirical
adjacent-sample correlation then falls to approximately $0.04$.
The split is needed because the generative model assumes conditionally
independent Gamma noise across pixels. Without it, the network can learn spatially
correlated speckle from the data, which the sampling model does not reproduce. In our
experiments this mismatch prevents the sampled images from reproducing the
scene structure reliably.
The resulting dataset contains 25,920 patches of size $128\times128$, of which
ten percent are held out. After the sublattice split, the ground spacing is
approximately $4.7\times27.8$\,m.

\paragraph{Pixelwise posterior predictions.}
At every pixel $i$, the network represents the Gamma posterior transform as a four-component mixture:
\begin{equation}
\widehat H_{i,L}(s,w)
= \sum_{j=1}^4
\omega_{i,j}(L,s)
\left(1+\frac{w}{\kappa_{i,j}(L,s)}\right)^{-\alpha_{i,j}(L,s)}.
\label{eq:app-radar-mixture}
\end{equation}
The weights $\omega_{i,j}$ are nonnegative and sum to one, and
$\alpha_{i,j}$ and $\kappa_{i,j}$ are positive.
For fixed network outputs, this function is the Laplace transform of a
Gamma mixture for $1/X_{i}$, which corresponds to an inverse-Gamma mixture
for $X_{i}$. Although the output is pixelwise, all mixture parameters can
depend on the complete input patch.
Uniform bounds $\alpha_{i,j}\geq\alpha_0>0$ and $\kappa_{i,j}\leq\kappa_{\max}<\infty$ would
give the polynomial large-$w$ decay that \eqref{eq:app-gamma-class} requires. We do not
enforce them. The remaining assumptions of the Gamma continuation theorem, uniform
Lipschitz continuity in the observed state and the integrated bound
\eqref{eq:app-gamma-training-lipschitz}, do not follow from the mixture representation
either.

Training uses the I-divergence with the Gamma target constructed from the
second half-look, with target shape $1/2$. By Proposition~\ref{prop:app-gamma-target},
this target has infinite second moment for every $w>0$, which rules out squared-error
regression. The optimal I-divergence risk is finite by
Proposition~\ref{prop:app-gamma-idiv}. Since the one-dimensional Gamma family with known
shape is complete, this target is the only integrable unbiased estimator of
$e^{-w/X_{i}}$ based on one observation, so no unbiased target from the same measurement
has a finite second moment.

To reduce the effect of extreme target values, we average the target over a $2\times2$
window. The average still has infinite second moment, and it is biased where the
reflectivity $X$ changes within the window: a plain $2\times2$ average reduces the mean
intensity of the brightest $0.1\%$ of pixels from $0.77$ to $0.62$ relative to the
reference. We therefore average the target only for jump sizes $w$ below a local,
state-dependent scale and leave larger jump sizes unsmoothed.

On held-out patches, the probability integral transform (PIT) of the second half-look under
the predicted law has Kolmogorov--Smirnov (KS) distance $0.005$, against $0.25$ for a simple
Gamma plug-in predictor centered on the noisy observation. Without the target averaging, the
pooled distance increases to $0.016$. Pixel-scale distances of this size should be read as
comparisons between predictors. About $1\%$ of the observed pixel values are clipped at their
lower bound, which puts a small atom in the empirical distribution that no continuous
predictive law has, and the resulting spike in the PIT is of the same order as the distances
themselves. A clipped value moves a $16\times16$ block sum by about $10^{-7}$ in relative
terms, so block-sum distances are unaffected. Comparisons that score two predictors against
the same target are unaffected at every scale. Nominal $50\%$, $80\%$, $90\%$ and $95\%$
predictive intervals have empirical coverage $49.9\%$, $80.2\%$, $90.2\%$ and $95.2\%$.

\paragraph{Refinement from an observed half-look.}
All results below except the longer chains use two refinement seeds of a single base model. We study conditional
trajectories that start from the observed input half-look $Y_{\rm in}$. Refinement simulates
additional looks and lasts $40$ epochs, run as two consecutive $20$-epoch stages, the second
of which widens the simulated range.
Refinement adds two terms to the Noise2Noise loss. In the consistency term, the prediction at the
starting state is detached from differentiation. The rollout regression uses half of each
training batch. Starting from the first half-look, we simulate the current model to a level
$L\in(L_{\mathrm{obs}},50L_{\mathrm{obs}}]$ drawn log-uniformly, and regress the prediction at
the simulated state on the target constructed from the second half-look. This target remains
unbiased at the simulated state, because the two views are conditionally independent given
the reflectivity.

The Noise2Noise loss supervises levels up to
$L_{\mathrm{obs}}$, and the rollout regression places the network on its own states up to
$50L_{\mathrm{obs}}$. The held-out tests are at $2L_{\mathrm{obs}}$, against the second view, and at about $4.5L_{\mathrm{obs}}$, against a reference band reserved from training. 

The rollout regression does not supervise the law of the trajectory. The rollout draws its
randomness independently of the second view given the first, so the optimum of the regression
at a simulated state is the posterior prediction given the \emph{starting} view, which is
broader than the posterior given the simulated state. We keep the term because it stabilizes
the prediction at $L_{\rm obs}$, and we evaluate the simulated increments separately.

We normalize increments by $L-L_{\mathrm{obs}}$. For the true Gamma process,
\[
\mathbb E\!\left[ \frac{ S_L-Y_{\rm in} }{ L-L_{\mathrm{obs}} } \middle| Y_{\rm in} \right] = \mathbb E[ X\mid Y_{\rm in} ].
\]
After refinement, the mean normalized increment of the learned trajectory agrees with the
network's predicted posterior mean to within $4\%$ through $16L_{\mathrm{obs}}$. At
$L=2L_{\mathrm{obs}}$, the true increment $S_{2L_{\mathrm{obs}}}-Y_{\rm in}$ has, conditional on
$X$, exactly the same distribution as $Y_{\rm tar}$. For $16\times16$ block sums, the
second-view PIT of the learned trajectory has KS distance $0.048$ and $0.030$ on the two seeds
(Table~\ref{tab:app-radar-blocks}). Because of the clipped pixel values, we do not read the
pixel column of that table as a property of the trajectory.

We compare second moments of the simulated increments with the Gamma scaling, using a
reference texture variance estimated from the cross-covariance of the two half-looks. The
posterior mean alone would underestimate this variance by roughly a factor of four. On this
heavy-tailed field, the cross-covariance estimate is itself biased downward, most strongly at
small block sizes. At the $4$-, $8$- and $16$-pixel block scales, the second moments follow
the Gamma scaling to within approximately $10\%$ for $L\leq8L_{\mathrm{obs}}$. We do not report the pixel scale over this range. At $L=16L_{\mathrm{obs}}$ the
fine-scale variance falls below the reference. Since the reference is biased downward, a ratio against it would understate this gap, and we do not quantify it.

\paragraph{Joint and independent samplers.}
The pixelwise posterior mixture in
\eqref{eq:app-radar-mixture} specifies one-pixel conditional
distributions, but it does not by itself specify how uncertainty is shared
across pixels. We therefore evaluate spatial dependence by predicting sums
over blocks of the held-out second half-look.
We compare two samplers.
The first draws one reflectivity value independently at every pixel from
the predicted pixelwise mixture and then draws an independent Gamma
half-look conditional on those reflectivities.
The second is the joint sampler, which follows the learned conditional trajectory and can propagate dependence across pixels through the network's state-dependent rates. For each of the two refinement seeds, both methods use 32 Monte Carlo draws
for each of 256 held-out patches. Table~\ref{tab:app-radar-blocks} reports
the KS distance and the coverage of nominal $90\%$ intervals.

At the single-pixel level, the two methods have similar coverage. At
$16\times16$ block scale, however, the independent sampler gives only
$82.5$--$83.2\%$ coverage, whereas the joint sampler gives
$88.6$--$89.2\%$, close to the nominal $90\%$.
\begin{table}[htbp]
\centering
\small
\setlength{\tabcolsep}{4pt}
\begin{tabular}{@{}clcccc@{}}
\toprule
Seed & Sampler & Pixel & $4\times4$ & $8\times8$ & $16\times16$ \\
\midrule
1 & Joint sampler & $0.004/89.9$ & $0.013/89.9$ & $0.026/89.6$ & $0.048/88.6$ \\
1 & Independent sampler & $0.002/90.3$ & $0.029/88.3$ & $0.056/85.9$ & $0.100/82.5$ \\
0 & Joint sampler & $0.004/89.8$ & $0.013/89.5$ & $0.019/89.4$ & $0.030/89.2$ \\
0 & Independent sampler & $0.002/90.3$ & $0.031/88.3$ & $0.057/86.1$ & $0.088/83.2$ \\
\bottomrule
\end{tabular}
\caption{Prediction of held-out Sentinel-1 second-view block sums.
Each entry reports KS distance / empirical coverage of the nominal
$90\%$ interval, in percent. For each refinement seed, both samplers use
32 draws on 256 patches drawn from the training run's own held-out split,
disjoint from every patch the model was trained on.}
\label{tab:app-radar-blocks}
\end{table}

At the $16\times16$ scale, the PIT of the joint sampler is skewed to one side while
its interval widths stay accurate. Its KS distance grows with block size, but its
coverage falls by at most 1.3 points. The lowest of ten PIT bins holds $13.3\%$ and $11.9\%$ of the mass on
the two seeds, against $10\%$ expected, so the real block sums tend to lie below the generated
draws. Only the part of this excess below the $5\%$ point reduces the coverage of the $90\%$
interval, which is consistent with the $88$--$89\%$ in Table~\ref{tab:app-radar-blocks}. The
independent sampler has a larger excess in the same bin, $17.4\%$ and $18.6\%$.
Block-sum calibration also varies between adjacent checkpoints of a single run, and the
training loss does not identify the best checkpoint. We select the checkpoint on a held-out
slice disjoint from the one we report.

\paragraph{Ablation of the refinement terms.}
To separate the consistency term from the rollout regression, we branch four models from the
same unrefined checkpoint and refine each for $40$ epochs in a single stage, with each term
switched on or off. The budget matches that of the reported model, but none of the four is the
reported model. For each, we measure the fraction of the network's own per-pixel posterior
spread that the trajectory realizes at the reserved reference band.

We use this ratio because absolute coverage at the band is hard to interpret. The band is
narrower than the training views, so it resolves a somewhat smoother field, and its effective
number of looks, about $1.8$, is known only approximately. These effects and the clipped pixel
values are shared by all four models and cancel in the ratio. For the same reason, only changes
in coverage at the band are informative. Coverage is $91.4\%$ and $91.1\%$ on the two seeds
before refinement and $90.8\%$ on both after, against a nominal $90\%$. Refinement thus
changes coverage by less than a point, while the realized spread increases from $0.947$ to
about $0.985$.
\begin{table}[htbp]
\centering
\small
\begin{tabular}{@{}lcc@{}}
\toprule
Refinement & Seed 1 & Seed 0 \\
\midrule
None, unrefined checkpoint      & $0.947$ & $0.947$ \\
Neither term, same budget       & $0.969$ & $0.964$ \\
Rollout regression only         & $0.916$ & $0.934$ \\
Consistency loss only           & $0.990$ & $0.992$ \\
Both                            & $0.986$ & $0.984$ \\
\bottomrule
\end{tabular}
\caption{Fraction of the network's own per-pixel posterior spread that the sampled trajectory
realizes at the reserved reference band, as a ratio of log-variances over $512$
held-out patches. A value of $1$ means the trajectory expresses exactly the spread the network
predicts. All refined rows branch from the same unrefined checkpoint and use the reported
model's budget of $40$ epochs, run in a single stage. The second row is the zero-weight control,
whose seed-0 run stopped early at 30 of 40 epochs on the training criterion.}
\label{tab:app-radar-ablation}
\end{table}

The consistency loss alone raises the ratio from $0.947$ to $0.990$ and $0.992$, whereas rollout
regression alone lowers it to $0.916$ and $0.934$, below the unrefined checkpoint. With both
terms the ratio is $0.986$ and $0.984$. The zero-weight control, trained for the same budget
without either term, reaches $0.969$ and $0.964$, so extra training alone gives roughly
$40$ to $50\%$ of the gain from the consistency loss.
The model refined with the consistency loss alone is, however, less stable at $L_{\rm obs}$. Across two seeds, its mean residual there varies more than that of the other refined models. We therefore report the model with both terms.

\paragraph{Despeckling.}
We score an estimator $\widehat X$ by the residual ratio
$Y_{\rm tar}/(L_{\mathrm{obs}}\widehat X)$ on the same $256$ held-out patches. Its mean is $1$
when the estimate is calibrated, in the sense $\mathbb E[X\mid \widehat X]=\widehat X$.
The usual unbiasedness condition $\mathbb E[\widehat X\mid X]=X$ is a different
requirement. By Jensen's inequality, an unbiased estimate has a residual mean of at least one. The $5\%$ null floor of the residual ratio's KS distance is $0.0007$.
Table~\ref{tab:app-radar-baselines} compares our posterior mean with a $7\times7$ Lee filter.

\begin{table}[htbp]
\centering
\small
\begin{tabular}{@{}lcc@{}}
\toprule
Estimator of $X$ & residual mean & residual KS \\
\midrule
Ours, posterior mean & $1.003$ & $0.014$ \\
Lee $7\times7$       & $0.707$ & $0.141$ \\
\bottomrule
\end{tabular}
\caption{Despeckling on held-out Sentinel-1 patches. The residual ratio
$Y_{\rm tar}/(L_{\mathrm{obs}}\widehat X)$ has mean $1$ for a calibrated estimate. Its $5\%$
null KS floor is $0.0007$.}
\label{tab:app-radar-baselines}
\end{table}

\paragraph{Longer chains.}
In one additional experiment with a single training seed, the sampler runs out to $50L_{\mathrm{obs}}$. At $8\times8$ block scale, coverage of the held-out second view is $89.0\%$, compared with $85.0\%$ for independent pixelwise draws. The mean simulated increment stays within $5\%$ of the network's posterior mean through $32L_{\mathrm{obs}}$, and within $8\%$ at $50L_{\mathrm{obs}}$. Variance at the $8$- and $16$-pixel block scales remains close
to the reference through $L=25$, that is $50L_{\mathrm{obs}}$. At the pixel scale, the variance
falls below the reference beyond $16L_{\mathrm{obs}}$. As discussed above, the
cross-covariance estimate of the reference is unreliable on this field, so we do not quantify
the deficit. Even at $L=25$, the observation model is not noise-free: its conditional speckle variance is $4\%$ of the squared reflectivity. Calibration against the held-out second view also does not show that the chain state at 25 looks follows the true conditional law of a 25-look measurement given the observed half-look. 

\end{document}

%% file: preambule.tex
\usepackage{amsmath}

\usepackage{amsthm}
\usepackage{amssymb}
\usepackage{amsfonts}
\IfFileExists{dsfont.sty}{\usepackage{dsfont}}{}
\IfFileExists{accents.sty}{\usepackage{accents}}{}
\usepackage{bm}
\usepackage{hyperref}
\IfFileExists{bbm.sty}{\usepackage{bbm}}{}
\usepackage{mathtools}

\usepackage{tabularx} 
\usepackage{caption}
\usepackage{multirow}
\usepackage{booktabs}
\usepackage{graphicx}
\graphicspath{{Figures/}}
 \usepackage{subcaption}
 \usepackage{tcolorbox}
 \usepackage{wasysym}
  \usepackage{longtable}

\usepackage{tikz,pgfplots}
\pgfplotsset{compat=newest}
\usepgfplotslibrary{groupplots}
\usetikzlibrary{intersections,calc,arrows,matrix,spy}
\usepackage{color}
\definecolor{darkred}{rgb}{0.6,0,0}
\definecolor{darkgreen}{rgb}{0.19,0.44,0.09}
\definecolor{darkblue}{rgb}{0,0,0.5}
\definecolor{SkyBlue}{rgb}{0.53, 0.81, 0.92}

\IfFileExists{algorithm.sty}{\IfFileExists{algpseudocode.sty}{%
\usepackage{algorithm}
\usepackage{algpseudocode}}{}}{}
\IfFileExists{SIunits.sty}{\usepackage[squaren,Gray]{SIunits}}{}

\usepackage[normalem]{ulem}
\usepackage{enumitem}

\usepackage{pgfplotstable}
\usepackage{pgfplots}
\pgfplotsset{compat=1.18}
\usepgfplotslibrary{groupplots}
\pgfplotstableset{col sep=comma}

\usepackage{datatool}

%% file: commands.tex
\def\E{\mathbb{E}}

\newcommand{\xb}{\mathrm{\mathbf{x}}}

\newcommand{\dd}{\mathrm{d}}

\newcommand{\Nc}{\mathcal{N}}

\def\Id{\mathrm{Id}}

\usepackage{tcolorbox}
\usepackage{mdframed}
\usepackage{lipsum}

\newtheorem{remark}{Remark}[section]

\newtheorem{lemma}{Lemma}[section]
\newtheorem{theorem}[lemma]{Theorem}
\newtheorem{proposition}[lemma]{Proposition}
\newtheorem{corollary}[lemma]{Corollary}

\theoremstyle{definition}
\newtheorem{assumption}{Assumption}[section]

\newcommand{\eqdef}{\ensuremath{\stackrel{\mbox{\upshape\tiny def.}}{=}}}

\def\1{1\!\!1}

\def\Lso3{L^2(SO(3))}
\def\so3{SO(3)}

\definecolor{cOv1}{HTML}{9F8FFA}
\definecolor{cOv2}{HTML}{7963F8}
\definecolor{cOv3}{HTML}{5437F6}
\definecolor{cOv5}{HTML}{2E0BF4}
\definecolor{cOv8}{HTML}{2209B5}
\definecolor{cOv10}{HTML}{1E079C}
\definecolor{cOv18}{HTML}{0A021F}
\definecolor{cOurs}{HTML}{FFB066}

\definecolor{cOv20}{HTML}{9ECAE1}
\definecolor{cOv30}{HTML}{C6DBEF}
\definecolor{cOurs10}{HTML}{FFB066}
\definecolor{cOurs100}{HTML}{CC7A00}
\definecolor{cGD}{HTML}{3A7F6F}

%% file: process_overview.tex
\begin{figure}[t]
\centering
\begin{tikzpicture}[x=1cm,y=1cm,>=stealth,font=\footnotesize]
\definecolor{pathblue}{RGB}{53,94,127}
\definecolor{pathgreen}{RGB}{47,133,82}
\definecolor{n2nfill}{RGB}{234,242,250}
\definecolor{consfill}{RGB}{237,245,239}
\fill[n2nfill,rounded corners=5pt] (0.05,-2.05) rectangle (7.98,1.9);
\fill[consfill,rounded corners=5pt] (8.04,-2.05) rectangle (13.9,1.9);
\node[anchor=north,align=center,inner sep=2pt] at (4.015,1.87)
  {\textbf{N2N loss} ($a\le a_{\rm in}$)\\[-1pt]
   {\scriptsize\color{black!65}inputs by degrading the data, targets from a second measurement}};
\node[anchor=north,align=center,inner sep=2pt] at (10.97,1.87)
  {\textbf{Consistency loss} ($a>a_{\rm in}$)\\[-1pt]
   {\scriptsize\color{black!65}prediction at $a_{\rm in}$ matches its average along simulated paths}};
\node[inner sep=0pt] at (1.05,0) {\includegraphics[width=1.4cm]{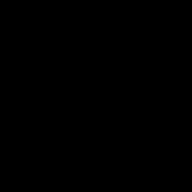}};
\node[inner sep=0pt] at (3.025,0) {\includegraphics[width=1.4cm]{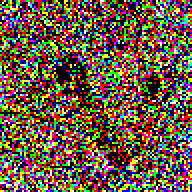}};
\node[inner sep=0pt] at (5.0,0) {\includegraphics[width=1.4cm]{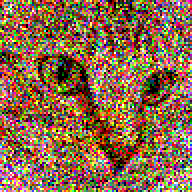}};
\node[inner sep=0pt] at (6.975,0) {\includegraphics[width=1.4cm]{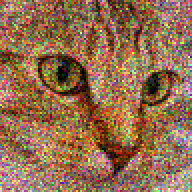}};
\node[inner sep=0pt] at (10.925,0) {\includegraphics[width=1.4cm]{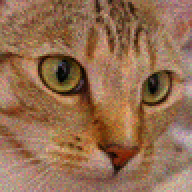}};
\node[inner sep=0pt] at (12.9,0) {\includegraphics[width=1.4cm]{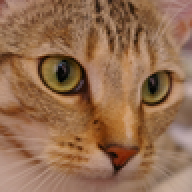}};
\node[inner sep=0pt] at (8.95,0.48) {\includegraphics[width=0.44cm]{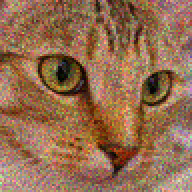}};
\node[inner sep=0pt] at (8.95,0) {\includegraphics[width=0.44cm]{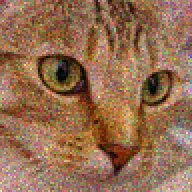}};
\node[inner sep=0pt] at (8.95,-0.48) {\includegraphics[width=0.44cm]{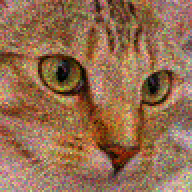}};
\foreach \x in {1.05, 3.025, 5.0}
  \draw[pathblue,line width=0.5pt] (\x-0.7,-0.7) rectangle (\x+0.7,0.7);
\draw[pathblue,line width=1.3pt] (6.975-0.7,-0.7) rectangle (6.975+0.7,0.7);
\foreach \x in {10.925, 12.9}
  \draw[pathgreen,dashed,line width=0.5pt] (\x-0.7,-0.7) rectangle (\x+0.7,0.7);
\foreach \y in {0.48, 0, -0.48}
  \draw[pathgreen,dashed,line width=0.4pt] (8.95-0.22,\y-0.22) rectangle (8.95+0.22,\y+0.22);
\node[anchor=south west,font=\scriptsize,text=black!60,inner sep=1pt] at (0.35,0.76) {empty state};
\node[anchor=south,font=\scriptsize,text=pathblue,inner sep=1pt] at (6.975,0.76) {training data};
\node[anchor=south east,font=\scriptsize,text=black!60,inner sep=1pt] at (13.6,0.76) {clean limit};
\node at (1.05,-0.98) {$a=0$};
\node at (3.025,-0.98) {$a=1$};
\node at (5.0,-0.98) {$a=4$};
\node[pathblue] at (6.975,-0.98) {$Y_{\rm in}$, $a_{\rm in}=16$};
\node at (8.95,-0.98) {$a=64$};
\node at (10.925,-0.98) {$a=256$};
\node at (12.9,-0.98) {$a\to\infty$};
\foreach \xa/\xb in {1.05/3.025, 3.025/5.0, 5.0/6.975}
  \draw[->,pathgreen,line width=0.6pt] (\xa+0.74,0.24) to[bend left=50] (\xb-0.74,0.24);
\foreach \y in {0.48, 0, -0.48}
  \draw[->,pathgreen,line width=0.6pt] (7.725,0) .. controls (8.25,0) and (8.4,\y) .. (8.68,\y);
\draw[->,pathgreen,line width=0.6pt] (9.2,0) to[bend left=30] (10.19,0);
\draw[->,pathgreen,line width=0.6pt] (11.665,0) to[bend left=50] (12.16,0);
\foreach \xa/\xb in {3.025/1.05, 5.0/3.025, 6.975/5.0}
  \draw[->,pathblue,line width=0.6pt] (\xa-0.74,-0.24) to[bend left=50] (\xb+0.74,-0.24);
\draw[->,pathblue,line width=0.6pt] (2.45,-1.42) to[bend left=50] (1.95,-1.42);
\node[anchor=west,text=pathblue,inner sep=1pt] at (2.6,-1.45) {degradation with the known $\overleftarrow{K}_{b,a}$};
\draw[->,pathgreen,line width=0.6pt] (8.55,-1.42) to[bend left=50] (9.05,-1.42);
\node[anchor=west,text=pathgreen,inner sep=1pt] at (9.2,-1.45) {generation with the learned $\widehat h_{\xi,a}$};
\draw[->,black!70,line width=0.7pt] (0.3,-1.78) -- (13.65,-1.78);
\node[anchor=north east,font=\scriptsize,text=black!70,inner sep=1pt] at (13.65,-1.8) {information level $a$};
\end{tikzpicture}
\caption{The Poisson observation process of one image. The training data are counts $Y_{\rm in}$ at one
level $a_{\rm in}$. Below $a_{\rm in}$, the known kernels degrade them into N2N inputs (blue). Generation runs with the learned predictions from the empty state at $a=0$ (green arrows), and only the learned process reaches levels above $a_{\rm in}$. There, the consistency loss matches the prediction at $a_{\rm in}$ with its average along simulated paths (green, Section~\ref{sec:formulation}). The tiles above $a_{\rm in}$ show the true continuation of this image, and the three at $a=64$ are independent continuations of the same measurement.}
\label{fig:diff-overview}
\end{figure}